\documentclass[11pt]{article}

\usepackage[preprint]{acl}

\usepackage{times}
\usepackage{latexsym}

\usepackage[T1]{fontenc}

\usepackage[utf8]{inputenc}
\usepackage{CJKutf8}

\usepackage{microtype}

\usepackage{inconsolata}

\usepackage{graphicx}
\usepackage{svg}

\usepackage{CJK}
\usepackage[english]{babel}
\usepackage{hyperref}       
\usepackage{url}            
\usepackage{booktabs}       
\usepackage{amsfonts}       
\usepackage{nicefrac}       
\usepackage{microtype}      
\usepackage{xcolor}         

\title{HKCanto-Eval: A Benchmark for Evaluating Cantonese Language Understanding and Cultural Comprehension in LLMs}

\author{
 \textbf{Tsz Chung Cheng\textsuperscript{1}},
 \textbf{Chung Shing Cheng\textsuperscript{2}},
 \textbf{Chaak Ming Lau\textsuperscript{3}},
\\
 \textbf{Eugene Tin-Ho Lam\textsuperscript{5}},
 \textbf{Chun Yat Wong\textsuperscript{4}},
 \textbf{Hoi On Yu\textsuperscript{5}},
\textbf{Cheuk Hei Chong\textsuperscript{6,7}}
\\
\\
 \textsuperscript{1} Kyushu University,
 \textsuperscript{2} hon9kon9ize,
 \textsuperscript{3} The Education University of Hong Kong,
\\
 \textsuperscript{4} The University of Hong Kong,
 \textsuperscript{5} Independent Researcher,
 \textsuperscript{6} Votee AI,
 \textsuperscript{7} Beever AI
\\
 \small{
   \textbf{Correspondence:} Tsz Chung Cheng: \href{mailto:jed.cheng@mag.ed.kyushu-u.ac.jp}{jed.cheng@mag.ed.kyushu-u.ac.jp}, }
   \\
\small{Chung Shing Cheng: \href{mailto:joseph.cheng@hon9kon9ize.com}{joseph.cheng@hon9kon9ize.com}
Chaak Ming Lau: \href{mailto:lchaakming@eduhk.hk}{lchaakming@eduhk.hk}
}
}

\begin{document}

\maketitle
\begin{abstract}
The ability of language models to comprehend and interact in diverse linguistic and cultural landscapes is crucial. The Cantonese language used in Hong Kong presents unique challenges for natural language processing due to its rich cultural nuances and lack of dedicated evaluation datasets. The HKCanto-Eval benchmark addresses this gap by evaluating the performance of large language models (LLMs) on Cantonese language understanding tasks, extending to English and Written Chinese for cross-lingual evaluation. HKCanto-Eval integrates cultural and linguistic nuances intrinsic to Hong Kong, providing a robust framework for assessing language models in realistic scenarios. Additionally, the benchmark includes questions designed to tap into the underlying linguistic metaknowledge of the models. Our findings indicate that while proprietary models generally outperform open-weight models, significant limitations remain in handling Cantonese-specific linguistic and cultural knowledge, highlighting the need for more targeted training data and evaluation methods. The code can be accessed at \href{https://github.com/hon9kon9ize/hkeval2025}{https://github.com/hon9kon9ize/hkeval2025}
\end{abstract}

\section{Introduction}

Recent advancements in large language models (LLMs) such as GPT-4, Gemini, and various open-weight models have demonstrated remarkable capabilities in natural language understanding across multiple languages \citep{xu2024survey}. However, the performance of most models significantly declines when applied to languages other than English, yielding particularly poor outcomes for low-resource languages (LRLs). These languages are under-represented lingua francas that play a crucial role in certain communities, and it is imperative to improve multilingual support for LRLs by creating benchmarks to guide the future development of multilingual LLMs. Since they are poorly supported due to the lack of training data, if there is a close language with more resources, this problem can potentially be mitigated through few-shot learning. A notable example of this strategy is the use of Bahasa Indonesian to handle regional languages in Indonesia \citep{aji-etal-2022-one, winata-etal-2022-cross}. This strategy aligns with the spirit of language sustainability and AI support for marginalised communities \citep{du2020marginalized}, which is also applicable to Cantonese.

This paper investigates the status of LLM support for Cantonese (ISO 639-3: \textit{yue}), a member of the \textit{Sinitic} (``Chinese'') branch of the Sino-Tibetan language family, and a distinct variety unintelligible to users of Mandarin, the standard variety of Chinese used in Mainland China (Pǔtōnghuà) and Taiwan (Guóyǔ). Cantonese, spoken by over 85 million people according to \textit{Ethnologue} \citep{ethnologue2024}, serves as the most common and \textit{de facto} official language of Hong Kong and Macau, and is also widely used in parts of Guangdong, Guangxi, Malaysia, and Singapore. Additionally, it is used as a diasporic language in countries such as Canada \citep{sachdevl1987language}, the United States \citep{leung2012relationships}, Australia \citep{zhang2023home}, and the United Kingdom \citep{bauer2016hong, tsapali2023future}. Despite its widespread use, Cantonese is still considered a low-resource language \citep{xiang2024cantonese} due to the lack of quality written resources. This scarcity results from a ``diglossia" that requires Written Chinese (which resembles Mandarin) to be used in formal settings\footnote{Even in Mandarin-like Written Chinese, there are persistent lexical differences with other regions due to vastly different governmental, legal and education systems. For instance, the word ``taxi'' is rendered as ``\raisebox{-0.15em}{\includegraphics[height=1.0em]{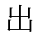}}\hskip0pt{}\raisebox{-0.15em}{\includegraphics[height=1.0em]{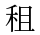}}\hskip0pt{}\raisebox{-0.15em}{\includegraphics[height=1.0em]{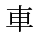}}\hskip0pt{}'' in mainland China, ``\raisebox{-0.15em}{\includegraphics[height=1.0em]{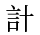}}\hskip0pt{}\raisebox{-0.15em}{\includegraphics[height=1.0em]{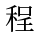}}\hskip0pt{}\raisebox{-0.15em}{\includegraphics[height=1.0em]{images/U+8ECA.png}}\hskip0pt{}'' in Taiwan, and ``\raisebox{-0.15em}{\includegraphics[height=1.0em]{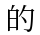}}\hskip0pt{}\raisebox{-0.15em}{\includegraphics[height=1.0em]{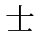}}\hskip0pt{}'' in Hong Kong and Macau.}, and a longstanding, ideologically-driven stigmatisation of Cantonese as an informal/vulgar language \citep{lau2024ideologically}, further confines written Cantonese to informal contexts like social media and texting.

Cantonese is partially supported by certain LLMs, with models like GPT-4 and Gemini capable of comprehending and responding in Cantonese \citep{fu2024efficacy, hong2024cantonmt, jiang2024how}. There are models dedicated to better supporting Chinese languages and dialects: The Hong Kong government is developing an internal tool based on locally developed LLMs for administrative use \citep{hkgpt}; SenseTime released SenseChat (Cantonese), a model trained on 6 billion tokens of Hong Kong-specific data \citep{sensetime}. However, the current support level is mostly contributed to by small pockets of Cantonese presented in the sheer volume of Written Chinese training data. 
There have been comparisons between Chinese and Western models on how well languages spoken in China are handled \citep{wen2025chinese}, showing that Chinese models outperformed Western ones on Mandarin, but the same cannot be said for Cantonese or other languages in China. The following section outlines how current benchmarking studies have yet to provide a comprehensive evaluation for Cantonese and Hong Kong-related tasks that tap into the in-depth representation of underlying aspects of the language, which we believe is the prerequisite for accurate comprehension in uncommon scenarios.

\begin{figure*}[t]
  \includegraphics[width=0.99\linewidth]{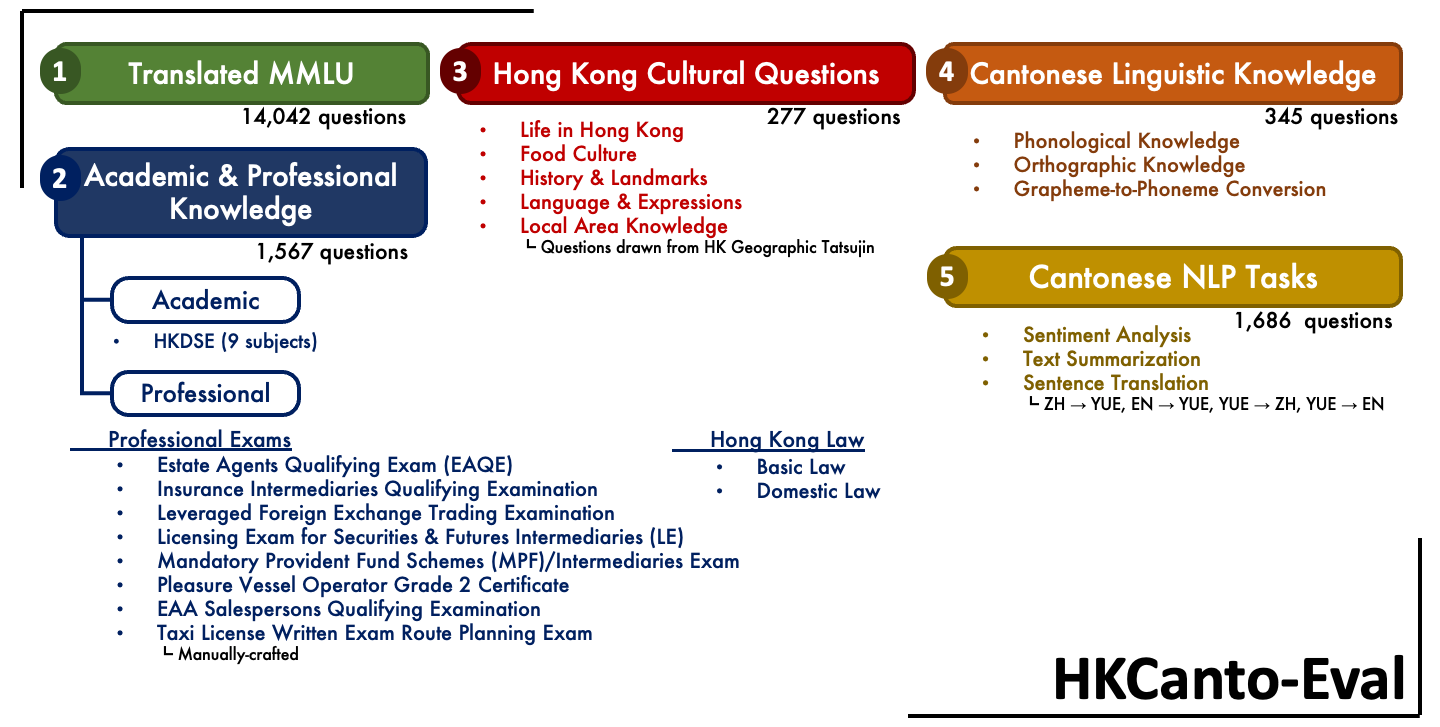} 
  \caption {Diagram showing the tasks of the HKCanto-Eval Benchmark}
\end{figure*}

\section{Related Benchmarks}

The development of LLMs has spurred significant research into evaluating their performance and comparing their capabilities to human reasoning across general and domain-specific tasks. A prominent benchmark in this area is the MMLU dataset \citep{hendrycks2020measuring}, which comprises 57 tasks ranging from elementary to university-level multiple-choice questions. Despite its widespread use, MMLU has been criticised for containing flawed questions and answers \citep{gema2024we, gupta2024changing}. To address these shortcomings, alternative benchmarks such as BIG-Bench \citep{srivastava2022beyond}, MMLU-Pro \citep{taghanaki2024mmlu}, and MMLU-Pro+ \citep{wang2024mmlu} have been introduced, aiming to improve accuracy while presenting more diverse and challenging questions.

In addition to comprehensive benchmarks, researchers have developed domain-specific, expert-curated datasets to evaluate the reasoning capabilities of LLMs in specialised fields such as programming (HumanEval \citep{chen2021evaluating}; NL2Code \citep{zan2022large}) and mathematical reasoning (GSM8K \citep{cobbe2021training}; MATH \citep{hendrycks2021measuring}; MATH 401 \citep{yuan2023well}; Omni-MATH \citep{gao2024omni}).

Although most existing LLM benchmarks focus on English-language tasks, culturally-aware datasets integrating machine-translated questions, native datasets, and exam questions have been developed in other languages, including Arabic \citep{koto2024arabicmmlu}, Basque \citep{etxaniz2024bertaqa, etxaniz2024latxa}, Spanish \citep{plaza2024spanish}, Indic languages \citep{verma2024milu}, and Korean \citep{son2024kmmlu}. Similar benchmarks have been published for Chinese, such as CMMLU \citep{li2023cmmlu} and C-Eval \citep{huang2024c} that gathered questions from various academic and professional exams in mainland China, and TMLU \citep{chen2024measuring} and TMMLU+ \citep{tam2024improved} that evaluate knowledge in Traditional Chinese in the context of Taiwan.

These benchmarks are not applicable to the Hong Kong context due to the aforementioned diglossia and regional lexical differences. Recently, \citet{jiang2024how} introduced a Cantonese evaluation benchmark that combines four datasets translated from other languages (ARC, GSM8K, CMMLU, and Truthful-QA)\footnote{It also contains a translation evaluation component for English-Cantonese and Simplified-to-Traditional Chinese translations, but its data sources and evaluation methods are not fully transparent.}, resulting in a dataset that is heavily biased towards American culture (16.9\% entries in the Truthful-QA dataset reference the United States) or mainland Chinese exams (CMMLU) (see Appendix A). 

\section{Methodology}

HKCanto-Eval introduces a specialised benchmark to address the lack of systematic tests for evaluating the Cantonese capabilities and Hong Kong knowledge of an LLM in these aspects: (1) \textbf{Language Proficiency}, the capability in an accurate and nuanced understanding of Cantonese and local-flavoured Written Chinese, as well as generating fluent, idiomatic, genre-appropriate Cantonese text in question and answering, translation, and summarisation tasks; (2) \textbf{Cultural Knowledge}, in-depth knowledge about not only general historical and geographical facts related to Hong Kong, but also everyday practices, local customs, beliefs and values, and cultural references from movies, music, literature, and internet culture; (3) \textbf{Reasoning and Problem-Solving}, reasoning and problem-solving skills within a Cantonese and/or Hong Kong-based context, including reasoning about the sound and written forms of the language.

These aspects are incorporated into the five datasets outlined below.

\subsection{Translated MMLU Dataset}
The first dataset comprises 14,042 questions from the original MMLU dataset in English \citep{hendrycks2020measuring} and their Cantonese translation\footnote{The translation was done by the Google Gemini 1.5 Flash API, which offers a balance between top performance and cost as one would find in the later section. To address concerns regarding the accuracy of LLM translation, we have selected 4 questions from each category for human checking. 202 out of 228 sentences were judged to be good by the raters.}. This allows us to compare how LLMs perform when handling knowledge in a wide range of subjects in Cantonese rather than in English (See Appendix B). 

\subsection{Academic and Professional Dataset}
The Academic and Professional Dataset is a set of multiple-choice questions curated to measure LLMs' reasoning and problem-solving abilities in domain-specific knowledge. The dataset contains multiple-choice questions from 3 sub-categories: (1) \textbf{Academic}: Questions sourced from Hong Kong Diploma of Secondary Education (HKDSE), a territory-wide high-school graduate-level exam; extracted and manually corrected from scanned PDFs and are believed to have never appeared online in a plain-text form; (2) \textbf{Professional}: Questions from seven professional qualification exams, extracted from text PDF files found on the corresponding official sites (in which the model answers were not on the same page as the questions, avoiding data contamination concerns), and an additional set of Taxi Licensing Exam Styled Route Planning questions on Hong Kong roads and geographical features; (3) \textbf{Law}: Questions about law in Hong Kong across 15 categories sourced from the Internet, and an additional subset of the Basic Law edited by the authors.

All questions are in Written Chinese (in the Traditional script). We also included an English version if it is available. The details of this dataset can be found in Appendix C. 

\subsection{Hong Kong Cultural Questions Dataset}
This dataset contains 277 manually crafted questions divided into five categories that capture cultural knowledge common to people who have lived or grown up in Hong Kong, that are often not learned in schools. The categories are \textbf{Food Culture}, \textbf{History and Landmarks}, \textbf{Language and Expressions}, \textbf{Life in Hong Kong} and \textbf{Local Area Knowledge}. The questions were collected in a way to capture knowledge from all walks of life. 244 questions were developed by the authors and volunteers for the first four categories, and the last category comes from an online quiz. Questions were created so that they were non-trivial and at the same time not too obscure, and have been verified by all the authors. Details can be found in Appendix D.

\subsection{Linguistic Knowledge Dataset}

This is an assessment of the linguistic knowledge represented in the models, inspired by the approach of PhonologyBench \citep{suvarna-etal-2024-phonologybench} for English. To our knowledge, this innovative approach has never been incorporated into existing Cantonese or Chinese benchmarks in general.

\subsubsection{Phonological Knowledge}

The dataset contains 100 questions that evaluate phonological knowledge about characters and words of an LLM, including the judgment of homophones and rhyming and other non-trivial reasoning tasks based on word pronunciation. These are particularly important in the Cantonese context, as the writing system does not provide reliable cues about the pronunciation of words, and Cantonese materials are not accompanied by sound transcription. This knowledge needs to be present in the training data for tasks that require sound-related operations or reasoning (See Appendix E.1).

\subsubsection{Orthographic Knowledge}

The Orthographic Knowledge Dataset evaluates the character meta-knowledge of an LLM. Cantonese users from Hong Kong need to know around 4,000 characters by the age of 12 and will have built sound knowledge about the representation of the characters. This subset contains 100 questions about the strokes, structure, arrangement, and radical and constituent components of common characters. Cantonese uses the Traditional Chinese script (ISO 15924: Hant) in Hong Kong and Macau, and the script is also used in Taiwan. There could be influence from Mandarin data or Taiwan usage not shared by Cantonese. It is also expected that certain models may produce incorrect answers due to the over-reliance on simplified Chinese data (See Appendix E.2).

\subsubsection{Grapheme-to-Phoneme (G2P) Conversion} 

This dataset addresses the task of converting a string of written text represented in Traditional Chinese characters into Jyutping, a widely adopted romanisation standard of Cantonese\footnote{\href{https://lshk.org/jyutping-scheme}{https://lshk.org/jyutping-scheme}}. This is similar to typical G2P tasks except that Jyutping is used instead of the International Phonetic Alphabet (IPA) as the output. G2P functionalities have been implemented by PyCantonese \citep{lee-etal-2022-pycantonese}, a Cantonese NLP package, Hambaanglaang Converter\footnote{\href{https://test.hambaanglaang.hk/}{\href{https://test.hambaanglaang.hk}{https://test.hambaanglaang.hk}}} and Visual Fonts\footnote{\href{https://visual-fonts.com}{https://visual-fonts.com}}. As the task is non-deterministic, rule-based conversions are bound to be unreliable (although Visual Fonts have achieved very high accuracy now). There is also no reliable non-rule-based G2P system to our best knowledge. This part of the dataset contains 150 pairs of Character-Jyutping sentences from both Standard Written Chinese and Cantonese and in a range of formality levels, manually checked by professional linguists from the Linguistic Society of Hong Kong, the organisation that established and maintains the Jyutping system. The score calculation method is discussed in Appendix E.3.

\subsection{NLP Tasks Dataset}
\label{sec:method_nlp}
Multiple-choice questions offer a structured approach to assess LLM factual knowledge and reasoning, but they are insufficient for evaluating real-world language understanding and generation. Open-ended tasks, including translation and summarisation, were incorporated.

A translation dataset comprising 20 Cantonese sentences with complex linguistic nuances was created, with each sentence manually translated into English and written Chinese (resulting in 4 translation pairs per sentence) (See Appendix F). For summarisation, 10 Cantonese articles and 10 TED talk subtitles were used. The importance of transcription-based summarisation, reflecting Cantonese's prevalence in oral communication, is emphasised by the inclusion of TED talks (See Appendix G).

Performance on traditional NLP tasks like sentiment analysis was also evaluated.  Leveraging the OpenRice dataset \citep{openrice} (restaurant reviews categorised as positive, neutral, or negative), 1200 reviews (avg. 309 characters) with a balanced sentiment distribution were included.  Additionally, a new dataset of 399 Facebook comments (avg. 24 characters), labelled by paid interns, was created (See Appendix H).

\subsection{Evaluation Method}
The evaluation process of multiple-choice questions follows the standard 5-shot evaluation procedures in MMLU formulation. However, for the Hong Kong Cultural Questions Dataset, a zero-shot evaluation was also conducted to emulate actual usage. The translated MMLU dataset used the same system prompt as the original MMLU dataset. For other multiple-choice questions, a short sentence with the name of the exam or question subcategory is added.

For the G2P dataset, character error rates (CER) and Levenshtein distance were both used to calculate the discrepancy between the model output and the gold standard in a five-shot evaluation. The summarisation tasks were evaluated without any example to avoid exceeding the context length of any model, while zero and three-shot evaluations were carried out for the translation task.

The outputs of both translation and summarisation evaluation were evaluated and graded by paid undergraduate students and teaching assistants. The rubric can be found in Appendix F and G. As technology improves, future LLMs can perform the task to offer scalability. Nonetheless, the results from this human evaluation will be useful for verifying the validity and consistency of LLM-as-a-judge in the future.

\subsection{Model Selection}
13 model families were selected for evaluation. Proprietary models including OpenAI GPT4o \citep{openai2024gpt4ocard} and GPT4-mini \citep{openaigpt4omini}, Google Gemini 1.5 Flash and Gemini 1.5 Pro \citep{team2024gemini} and Anthropic Claude 3.5 Sonnet \citep{anthropicclaude3} were selected for their reported superior performance across different languages. 

Three proprietary models from Chinese companies, including Doubao Pro from ByteDance \citep{doubao}, Erne 4.0 from Baidu \citep{BaiduErine} and SenseChat (Cantonese) from SenseTime \citep{sensetime}, were also incorporated. All proprietary models were accessed through their API, except SenseChat, which was accessed via the web interface due to a failure to get verified to use their API.

Popular multilingual open-weight models including Aya 23 8B \citep{aryabumi2024aya23openweight}, Gemma 2 2B, 9B and 27B \citep{team2024gemma}, Llama 3.1 8B, 70B and 405B \citep{dubey2024llama}, and Mistral Nemo Instruct 2407 12B \citep{mistralnemo} were included to assess their cross-lingual ability. The collection also included two open-weight multilingual models from Chinese companies, Yi 1.5 6B, 9B and 34B \citep{young2024yi} and Qwen2 7B and 72B \citep{yang2024qwen2}. In addition, CantoneseLLM (CLLM) v0.5 6B and 34B\footnote{\href{https://huggingface.co/hon9kon9ize/CantoneseLLMChat-v0.5}{https://huggingface.co/hon9kon9ize/CantoneseLLMChat-v0.5}} are two of the few open-weight models trained specifically on Cantonese data. They were trained by fine-tuning Yi 1.5 6B and 34B models with around 400 million tokens of Hong Kong-related content. Open-weight instructions fine-tuned models smaller than 70B parameters were evaluated using Nvidia H100 GPUs. The 70B and 405B models were evaluated using the API of SiliconFlow\footnote{\href{https://siliconflow.cn}{https://siliconflow.cn}}.

\section{Results}

\begin{table*}[t]
\centering
\begin{tabular}{l|cc|cc|cc|cc}
                  & \multicolumn{2}{c|}{MMLU}         & \multicolumn{2}{c|}{\begin{tabular}[c]{@{}c@{}}Academic \& \\ Professional\end{tabular}} & \multicolumn{2}{c|}{Cultural}     & \multicolumn{2}{c}{Average}       \\
Model             & EN              & YUE             & EN                                          & ZH                                         & 0-shot          & 5-shot          & EN              & ZH/YUE          \\ \hline
Claude 3.5 Sonnet & \textbf{85.0\%} & \textbf{81.5\%} & 75.1\%                                      & 75.2\%                                     & 71.7\%          & 75.0\%          & 80.1\%          & 75.8\%          \\
Doubao Pro        & 79.8\%          & 74.2\%          & 60.8\%                                      & 70.5\%                                     & 70.7\%          & 75.0\%          & 70.3\%          & 72.6\%          \\
Ernie 4.0         & 81.0\%          & 75.2\%          & 70.4\%                                      & 72.4\%                                     & 68.2\%          & 75.2\%          & 75.7\%          & 72.8\%          \\
Gemini 1.5 Flash  & 79.0\%          & 73.1\%          & 67.4\%                                      & 68.3\%                                     & 61.0\%          & 64.0\%          & 73.2\%          & 66.6\%          \\
Gemini 1.5 Pro    & 83.2\%          & 77.6\%          & 71.0\%                                      & 71.7\%                                     & 74.0\%          & 73.8\%          & 77.1\%          & 74.3\%          \\
GPT4o             & 84.8\%          & 80.3\%          & \textbf{77.6\%}                             & 75.3\%                                     & \textbf{77.5\%} & 77.2\%          & \textbf{81.2\%} & \textbf{77.6\%} \\
GPT4o-mini        & 76.7\%          & 69.4\%          & 62.0\%                                      & 65.6\%                                     & 55.6\%          & 60.6\%          & 69.4\%          & 62.8\%          \\
SenseChat         & 78.7\%          & 70.1\%          & 73.6\%                                      & \textbf{75.6\%}                            & 67.4\%          & \textbf{77.4\%} & 76.1\%          & 68.8\%          \\ \hline
Aya 23 8B         & 56.6\%          & 47.1\%          & 44.8\%                                      & 49.0\%                                     & 39.5\%          & 37.7\%          & 50.7\%          & 43.3\%          \\
CLLM v0.5 6B           & 58.6\%          & 51.7\%          & 50.9\%                                      & 53.5\%                                     & 52.0\%          & 56.1\%          & 54.7\%          & 53.3\%          \\
CLLM v0.5 34B          & 75.9\%          & 69.9\%          & 66.8\%                                      & 69.9\%                                     & 72.5\%          & 76.7\%          & 71.3\%          & 72.3\%          \\
Yi 1.5 6B         & 64.1\%          & 54.0\%          & 53.7\%                                      & 58.3\%                                     & 47.7\%          & 50.7\%          & 58.9\%          & 52.7\%          \\
Yi 1.5 9B         & 70.9\%          & 60.8\%          & 59.2\%                                      & 63.3\%                                     & 48.7\%          & 57.3\%          & 65.0\%          & 57.5\%          \\
Yi 1.5 34B        & 76.1\%          & 68.5\%          & 63.7\%                                      & 68.7\%                                     & 67.7\%          & 72.9\%          & 69.9\%          & 69.5\%          \\
Gemma 2 2B        & 58.5\%          & 46.5\%          & 45.3\%                                      & 48.5\%                                     & 33.3\%          & 35.2\%          & 51.9\%          & 40.9\%          \\
Gemma 2 9B        & 73.4\%          & 64.3\%          & 63.6\%                                      & 64.0\%                                     & 49.1\%          & 51.6\%          & 68.5\%          & 57.3\%          \\
Gemma 2 27B       & 76.4\%          & 68.4\%          & 65.1\%                                      & 68.1\%                                     & 57.1\%          & 60.9\%          & 70.7\%          & 63.6\%          \\
Llama 3.1 8B      & 69.0\%          & 56.4\%          & 51.4\%                                      & 57.1\%                                     & 45.6\%          & 52.7\%          & 60.2\%          & 52.9\%          \\
Llama 3.1 70B     & 80.3\%          & 74.9\%          & 68.2\%                                      & 70.0\%                                     & 63.0\%          & 64.4\%          & 74.2\%          & 68.1\%          \\
Llama 3.1 405B    & \textbf{84.5\%} & \textbf{78.4\%} & 70.9\%                                      & 74.2\%                                     & 67.9\%          & 69.9\%          & 77.7\%          & 72.6\%          \\
Mistral Nemo 12B  & 68.8\%          & 58.4\%          & 54.6\%                                      & 58.0\%                                     & 40.1\%          & 42.7\%          & 61.7\%          & 49.8\%          \\
Qwen2 7B          & 71.2\%          & 64.8\%          & 60.7\%                                      & 65.4\%                                     & 53.6\%          & 54.8\%          & 66.0\%          & 59.6\%          \\
Qwen2 72B         & 82.9\%          & 78.3\%          & \textbf{74.7\%}                             & \textbf{76.3\%}                            & \textbf{72.9\%} & \textbf{77.7\%} & \textbf{78.8\%} & \textbf{76.3\%} \\ \hline
Random            & 25.0\%          & 25.5\%          & 22.9\%                                      & 24.6\%                                     & 29.8\%          & 28.1\%          & 23.9\%          & 27.0\%         
\end{tabular}
\caption{\label{tab:mc-perf-table} Model performance on MMLU, Academic and Professional, and Cultural questions. Note that SenseChat refused to answer one subset of questions in Cultural Question 5-shot evaluation.}
\end{table*}

\subsection{MMLU}
Table \ref{tab:mc-perf-table} shows the results of the multiple-choice questions. Proprietary models and open-weight models like Llama 3.1 70B, 405B, and Qwen 2 72B performed well in MMLU, but experienced an average of 7.46 percentage point drop when questions were in Cantonese. Considering potential errors from machine translations, this is evidence of \textit{Cantonese reasoning and problem-solving ability}.

\subsection{Academic and Professional Questions}

The results of this dataset showed expected problem-solving abilities across models in different subject areas, in particular, general weaknesses in handling secondary school-level mathematics and strong performance in legal questions. Proprietary models generally performed better than open-weight models. The sub-scores in the individual tasks show that most models struggled with academic questions that were never posted online. It is worth noting that some open-weight models  (e.g. CLLM v0.5 34B and Qwen2 72B) outperformed most models, and we can conduct further investigation on what additional training data was used to achieve this performance. Written Chinese yielded better overall results, and this is attributed to the Law dataset, which only came in Chinese. Discounting this set, Written Chinese caused a slight drop in performance. This indicates that \textit{multi-lingual open-weight LLMs showed cross-lingual capabilities}, maintaining similar performance across both languages.

\subsection{Hong Kong Cultural Questions}

Proprietary models and Qwen 2 72B showed a good understanding of Hong Kong cultural knowledge, yet none of the models performed well across the subcategories. Looking into the sub-scores, models occasionally matched humans in most sub-tests (e.g. Food Culture and Life in HK ). 
However, when inspecting the results, good performance by percentage \textit{only reflects the size of existing Hong Kong knowledge represented in Wikipedia entries.}
For example, only two models (Yi 1.5 6B and Qwen2 72B) correctly answered the origin of Demae Itcho noodles sold in Hong Kong, while 94\% of humans did. 
The results for Language \& Expressions also show that \textit{most models did not have a nuanced understanding of Cantonese}. Compared to human performance at 85.8\%, SenseChat scored the highest point out of all models in 5-shot (79.6\%), but its performance dropped significantly in zero-shot (61.4\%). In zero-shot evaluation, CLLM v0.5 34B delivered the best performance at 77.3\%.
Furthermore, model size affects the performance of geospatial tasks, with open-source models in the 6-9B parameter range achieving only about 50\% of larger models' performance on Local Area Knowledge (e.g. Yi 1.5 34B 67.9\%, 9B 35.7\%). The overall results of this dataset suggest that Hong Kong cultural knowledge is underrepresented in LLM training. See Appendix C for details.

\begin{table*}[t]
\centering
\begin{tabular}{l|ccc|ccc|c}
                  & \multicolumn{3}{c|}{Phonological Knowledge}                                               & \multicolumn{3}{c|}{Orthographic Knowledge}                                                                                       & NLP        \\
Model             & \begin{tabular}[c]{@{}c@{}}Homo-\\ phone\end{tabular} & Rhyme           & Misc.           & \begin{tabular}[c]{@{}c@{}}Visual\\ Sim.\end{tabular} & \begin{tabular}[c]{@{}c@{}}Canton.\\ Char.\end{tabular} & Misc.           & Avg.            \\ \hline
Claude 3.5 Sonnet & 28.0\%                                                & 64.0\%          & 16.0\%          & 50.0\%                                                & 76.9\%                                                  & 59.3\%          & 89.2\%          \\
Doubao Pro        & 16.0\%                                                & 44.0\%          & 16.0\%          & 70.0\%                                                & 80.8\%                                                  & 48.1\%          & 87.0\%          \\
Ernie 4.0         & 28.0\%                                                & 60.0\%          & 18.0\%          & 70.0\%                                                & 80.8\%                                                  & 53.7\%          & 82.7\%          \\
Gemini 1.5 Flash  & 12.0\%                                                & 20.0\%          & 24.0\%          & 40.0\%                                                & 73.1\%                                                  & 31.5\%          & 83.2\%          \\
Gemini 1.5 Pro    & 16.0\%                                                & 40.0\%          & 24.0\%          & 50.0\%                                                & \textbf{88.5\%}                                         & 46.3\%          & 87.9\%          \\
GPT4o             & \textbf{56.0\%}                                       & \textbf{96.0\%} & \textbf{28.0\%} & 50.0\%                                                & 65.4\%                                                  & \textbf{63.0\%} & \textbf{89.6\%} \\
GPT4o-mini        & 20.0\%                                                & 60.0\%          & 20.0\%          & 30.0\%                                                & 57.7\%                                                  & 40.7\%          & 86.1\%          \\
SenseChat         & 16.0\%                                                & 36.0\%          & 22.0\%          & \textbf{75.0\%}                                       & 76.9\%                                                  & 42.6\%          & 78.8\%          \\ \hline
Aya 23 8B         & 12.0\%                                                & 40.0\%          & 14.0\%          & 15.0\%                                                & 19.2\%                                                  & 31.5\%          & 70.1\%          \\
CLLM v0.5 6B           & 24.0\%                                                & 8.0\%           & 18.0\%          & 20.0\%                                                & 50.0\%                                                  & 27.8\%          & 71.9\%          \\
CLLM v0.5 34B          & 28.0\%                                                & 28.0\%          & 14.0\%          & 35.0\%                                                & 76.9\%                                                  & 37.0\%          & 73.3\%          \\
Yi 1.5 6B         & 28.0\%                                                & 12.0\%          & 12.0\%          & 10.0\%                                                & 50.0\%                                                  & 20.4\%          & 56.6\%          \\
Yi 1.5 9B         & \textbf{36.0\%}                                       & 40.0\%          & 24.0\%          & 30.0\%                                                & 57.7\%                                                  & 18.5\%          & 72.2\%          \\
Yi 1.5 34B        & 16.0\%                                                & 32.0\%          & \textbf{26.0\%} & 30.0\%                                                & 61.5\%                                                  & 33.3\%          & 82.9\%          \\
Gemma 2 2B        & 8.0\%                                                 & 24.0\%          & 18.0\%          & 25.0\%                                                & 53.8\%                                                  & 22.2\%          & 73.4\%          \\
Gemma 2 9B        & 20.0\%                                                & 28.0\%          & 24.0\%          & 25.0\%                                                & 50.0\%                                                  & 33.3\%          & 85.0\%          \\
Gemma 2 27B       & 20.0\%                                                & 12.0\%          & 16.0\%          & 25.0\%                                                & 65.4\%                                                  & 24.1\%          & 83.2\%          \\
Llama 3.1 8B      & 12.0\%                                                & 16.0\%          & 18.0\%          & 25.0\%                                                & 42.3\%                                                  & 38.9\%          & 60.3\%          \\
Llama 3.1 70B     & 28.0\%                                                & 40.0\%          & 12.0\%          & 30.0\%                                                & 61.5\%                                                  & 35.2\%          & \textbf{84.5\%} \\
Llama 3.1 405B    & 20.0\%                                                & \textbf{44.0\%} & 18.0\%          & 35.0\%                                                & 65.4\%                                                  & \textbf{50.0\%} & 64.4\%          \\
Mistral Nemo 12B  & 12.0\%                                                & 28.0\%          & 10.0\%          & 25.0\%                                                & 23.1\%                                                  & 37.0\%          & 68.8\%          \\
Qwen2 7B          & 8.0\%                                                 & 40.0\%          & 12.0\%          & 35.0\%                                                & 46.2\%                                                  & 33.3\%          & 66.8\%          \\
Qwen2 72B         & 12.0\%                                                & 28.0\%          & 16.0\%          & \textbf{50.0\%}                                       & \textbf{76.9\%}                                         & 48.1\%          & 83.5\%          \\ \hline
Random/Control    & 16.0\%                                                & 28.0\%          & 24.0\%          & 30.0\%                                                & 11.5\%                                                  & 27.8\%          & 76.8\%         
\end{tabular}
\caption{\label{tab:linguistic-mc-perf-table}Model performance on Linguistic Knowledge Dataset multiple-choice questions and NLP tasks. The bottom row indicates the expected correctness from random selection for the Phonological and Orthographic Knowledge tasks. For NLP, the reported figure is the average evaluation of professionally prepared translations for translation tasks serving as a control.}
\end{table*}

\begin{table}[t]
\centering
\begin{tabular}{l|cc}
Model             & CER    & Levenshtein \\ \hline
Claude 3.5 Sonnet & 7.9\% & 0.018       \\
Doubao Pro        & 20.9\% & 0.044       \\
Ernie 4.0         & 34.4\% & 0.094       \\
Gemini 1.5 Flash  & 34.7\% & 0.083       \\
Gemini 1.5 Pro    & 15.3\% & 0.030       \\
GPT4o             & \textbf{5.4\%} & \textbf{0.009}       \\
GPT4o-mini        & 12.0\% & 0.023       \\
SenseChat         & 54.4\% & 0.163       \\ \hline
Aya 23 8B         & 96.6\% & 0.724       \\
CLLM v0.5 6B      & 94.1\% & 0.859       \\
CLLM v0.5 34B     & \textbf{23.4\%} & \textbf{0.058}       \\
Yi 1.5 6B         & 99.0\% & 0.577       \\
Yi 1.5 9B         & 97.2\% & 0.528       \\
Yi 1.5 34B        & 79.6\% & 0.837       \\
Gemma 2 2B        & 97.5\% & 0.524       \\
Gemma 2 9B        & 73.0\% & 0.259       \\
Gemma 2 27B       & 62.5\% & 0.201       \\
Llama 3.1 8B      & 69.9\% & 0.270       \\
Llama 3.1 70B     & 31.3\% & 0.086       \\
Llama 3.1 405B    & 26.3\% & 0.074       \\
Mistral Nemo 12B  & 59.8\% & 0.201       \\
Qwen2 7B          & 97.3\% & 0.466       \\
Qwen2 72B         & 74.0\% & 0.268       \\ \hline
Rule Based        & 0.8\% & 0.001      
\end{tabular}
\caption{\label{tab:g2p-perf-table}Model performance in the Grapheme-to-Phoneme (G2P) dataset. Scores calculated based on character error rates (CER) and Levenshtein distance. (Lower is better)}
\end{table}

\subsection{Linguistic and NLP Tasks}

These two groups of tasks reveal the representation of Cantonese phonological, orthographic, lexical and grammatical knowledge in existing models. The overall results (Table \ref{tab:linguistic-mc-perf-table}) show a consistent trend where proprietary models outperformed open-weight models (but more pronounced in linguistic tasks). GPT-4o led with 76.7\% and 89.6\% in both \textit{linguistic} and \textit{NLP} tasks. Lower scores are often due to chance-level performance when knowledge is absent, or below chance-level due to influence from Mandarin. Here are the key findings and observations:

\textit{Most LLMs understand Cantonese fine.} Most models performed well in Sentiment Analysis (GPT4o 79.7\%, Llama 3.1 405B 78.8\%), Translation (3-shot: GPT4o 98.3\%, Qwen2 72B 96.6\%), and Summarisation (Claude 3.5 Sonnet 92.7\%, Gemma 2 9B 91.3\%). Models that obtained lower scores are often due to task completion problems, e.g. failure to handle long input and problems with low-frequency/mixed-language tokens.

\textit{Proprietary and large open-weight models have good Cantonese lexical knowledge.} 
The performance in translation and sentiment analysis is closely tied to the ability to determine the meaning of Cantonese-specific words that are not found or used differently in Mandarin.
Most models also performed well in the Cantoense Character Selection sub-task (Canton. Char. in Table \ref{tab:linguistic-mc-perf-table}) under Orthographic Knowledge. It is noteworthy that despite good performance with proprietary models (73.1\% - 88.5\%) and some open-weight models (CLLM v0.5 34B and Qwen2 72B, both 76.9\%), GPT4o struggled with Cantonese orthography (65.4\%). 

\textit{LLMs in general lack knowledge about Cantonese pronunciation.} In the Grapheme-to-Phoneme (G2P) conversion task, all models performed far worse than the rule-based control (Visual Fonts v3.3, CER 0.8\%), with the closest being GPT-4o (5.4\%) and Claude 3.5 Sonnet (7.9\%) as shown in Table \ref{tab:g2p-perf-table}. The appalling results from all tested language models reveal how linguistic knowledge is seriously under-represented. While it is expected that the G2P tasks will be significantly improved in newer/future models, actual linguistic tasks that involve sounds require more advanced knowledge about the language's sound system. Most models struggled with tasks like judging homophone or rhyme pairs in Table \ref{tab:linguistic-mc-perf-table}, with GPT-4o being a notable exception (Homophone: 56.0\%; Rhyming: 96.0\%).  Poor (close to chance level) performance in other models is not only due to the lack of G2P ability, a prerequisite for phonological reasoning, but also due to how Mandarin homophones partially influence this task. This will continue to be challenging for Cantonese due to limited specialised data.

\textit{LLMs in general do not have meta-linguistic knowledge represented in Cantonese.} Although certain models, especially the Chinese proprietary models, performed well in the visual similarity task (SenseChat 70\%, Doubao 70\%, Ernie 75\%) or orthographic reasoning (GPT4o 63.0\%), the knowledge seems to have come from Simplified Chinese, thus their good performance is not transferred to Cantonese-specific items. This seems to be caused by insufficient descriptive knowledge about the structure and properties associated with the individual glyphs.

\section{Conclusion}

This paper presents HKCanto-Eval, the first comprehensive evaluation benchmark focusing on Hong Kong Cantonese, by comparing the Cantonese language support of 6 proprietary and 7 open-weight model families. Our findings indicate that while these models can understand Cantonese in various contexts, retrieve knowledge about Hong Kong, and address problems written in or about Cantonese to some extent, there are notable limitations. Most models, especially open-weight models in the 6–9B range, lack sufficient linguistic, cultural and professional knowledge in Cantonese and Hong Kong. Performance was particularly poor for questions requiring knowledge not commonly found in major online sources. 

One area that we paid close attention to is the presence of metalinguistic knowledge in these models. There is concern that models showed Cantonese proficiency in linguistic and NLP tasks primarily through Mandarin. If their linguistic understanding is based solely on Mandarin, they may perform well on simpler tasks but struggle significantly with ``false friends" between languages, as Mandarin knowledge becomes a hindrance. This benchmark introduces a novel perspective, focusing on Cantonese processing abilities beyond superficial slang and expressions. By requiring reasoning about sounds and characters specific to Cantonese, our benchmark provides a fairer judgement that credits models accurately capturing Cantonese phonology and orthography, while exposing those that appear competent in Cantonese but are heavily reliant on Mandarin.

This challenge in processing Cantonese is shared by other low-resource languages. As training data increases, models tend to favour high-resource languages like Mandarin Chinese. The apparent similarity between Cantonese and Written Chinese further affects the ability of even proprietary models to distinguish between these linguistic contexts accurately. Addressing the segregation of regional and linguistic knowledge is crucial for developing culturally and linguistically adaptive LLMs. This issue extends beyond Cantonese to other under-represented language communities.

\section{Limitations \& Future Directions}
The current benchmark exhibits several limitations. 

\textbf{Inaccuracies in machine-translated materials:} First, the use of machine translation introduces potential inaccuracies. While Gemini 1.5 Flash balances cost and quality, human-translated questions could provide a more reliable benchmark, albeit at a higher resource cost. The reliance on multiple-choice and text-based questions does not fully capture the capabilities required for practical LLM applications such as code generation and mathematical problem-solving, which demand coherent and contextual text generation. The dataset also lacks multi-modal data like image and audio, which is now supported by proprietary models and should be evaluated.

\textbf{Biases in topic selection:} The newly and manually created questions might contain biases and a lack of scalability and comprehensiveness. The cultural questions, predominantly created by colleagues and relatives of the authors, may introduce bias in cultural references and wordings, leading to an over-representation of certain perspectives while under-representing others, such as traditional practices. Political topics were also specifically excluded, due to political complications, limiting cultural representation. This can also be considered a reasonable compromise since many models (e.g. those from Chinese companies) are configured to censor these topics, and there is a risk that our accounts or IP addresses will be banned before we complete all the benchmarking tasks for this paper. 

\textbf{Lack of Crosslingual Evaluation:} English translations for cross-lingual ability evaluation were also not included due to resource limitations. An additional comparison should be added to compare whether the same set of questions will be answered less satisfactorily when presented in English or Standard Written Chinese instead of Cantonese, in line with the evaluation done for Basque \citep{etxaniz2024bertaqa} and Mongolian and Tibetan \citep{zhang2025cross}. We will leave this for future research.

\textbf{Reliance on human evaluation:} Human evaluation, while insightful, is not scalable. Automated and objective evaluation methods, such as LLM-as-a-judge or rule-based approaches, are necessary for efficient evaluation, but this is challenging due to the low-resource nature of Cantonese.

\textbf{Future directions} include developing benchmarks incorporating audio, images, and tables, and addressing the aforementioned limitations to create more comprehensive and representative evaluations.

\section*{Acknowledgments}
Open-source models evaluated with computer resources offered under the categories of Trial Use and General Projects by the Research Institute for Information Technology, Kyushu University. Votee AI gratefully sponsored the usage fee. T.C.C is supported by the MEXT Initiative to Establish Next-Generation Novel Integrated Circuit Centers (X-NICS). The project is partially supported by funding from the Centre for Research on Linguistics and Language Studies (CRLLS), the Education University of Hong Kong.

\bibliography{custom}

\newpage

\section*{Appendix}
\appendix

\begin{CJK*}{UTF8}{bsmi}
\section{Questionable Practices in an Existing Work}
\label{sec:appendix1}
\citet{jiang2024how} recently proposed a Cantonese evaluation dataset consisting of 5 datasets. The authors cited a HuggingFace organisation homepage in a footnote for their translation dataset but offered no further specifics. Yet, the dataset's entries were not identifiable within the linked account. Nonetheless, the translation pairs between Cantonese and English are dubious. During the machine translation process, the less advanced models before GPT-3.5 tend to break English sentences into phrases by punctuation marks or connecting words such as "and" and "but". Then the phrases were translated individually and finally put together into one sentence in Cantonese. As a result, the translated texts were full of wrong wordings such as the following translation pair:

English:
\begin{quote}
Once upon a time, there was a three year old girl named Gwen. One day, she was walking with her mom when she saw something unusual.  She wondered out loud, â€œWhat is that?â€ Her mom explained, â€œThatâ€™s an old license. It shows that the person is allowed to drive.â€  Gwen then asked, â€œCan I get one?â€ Her mom smiled, shaking her head. â€œNo, not now. 
\end{quote}
Cantonese translation:
\begin{quote}
    \raisebox{-0.15em}{\includegraphics[height=1.0em]{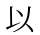}}\hskip0pt{}\raisebox{-0.15em}{\includegraphics[height=1.0em]{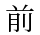}}\hskip0pt{}\raisebox{-0.15em}{\includegraphics[height=1.0em]{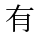}}\hskip0pt{}\raisebox{-0.15em}{\includegraphics[height=1.0em]{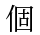}}\hskip0pt{}\raisebox{-0.15em}{\includegraphics[height=1.0em]{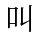}}\hskip0pt{} Gwen \raisebox{-0.15em}{\includegraphics[height=1.0em]{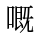}}\hskip0pt{}\raisebox{-0.15em}{\includegraphics[height=1.0em]{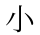}}\hskip0pt{}\raisebox{-0.15em}{\includegraphics[height=1.0em]{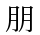}}\hskip0pt{}\raisebox{-0.15em}{\includegraphics[height=1.0em]{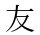}}\hskip0pt{}，\raisebox{-0.15em}{\includegraphics[height=1.0em]{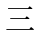}}\hskip0pt{}\raisebox{-0.15em}{\includegraphics[height=1.0em]{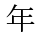}}\hskip0pt{}\raisebox{-0.15em}{\includegraphics[height=1.0em]{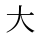}}\hskip0pt{}。\raisebox{-0.15em}{\includegraphics[height=1.0em]{images/U+6709.png}}\hskip0pt{}\raisebox{-0.15em}{\includegraphics[height=1.0em]{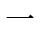}}\hskip0pt{}\raisebox{-0.15em}{\includegraphics[height=1.0em]{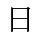}}\hskip0pt{}，\raisebox{-0.15em}{\includegraphics[height=1.0em]{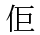}}\hskip0pt{}\raisebox{-0.15em}{\includegraphics[height=1.0em]{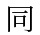}}\hskip0pt{}\raisebox{-0.15em}{\includegraphics[height=1.0em]{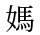}}\hskip0pt{}\raisebox{-0.15em}{\includegraphics[height=1.0em]{images/U+5ABD.png}}\hskip0pt{}\raisebox{-0.15em}{\includegraphics[height=1.0em]{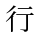}}\hskip0pt{}\raisebox{-0.15em}{\includegraphics[height=1.0em]{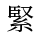}}\hskip0pt{}\raisebox{-0.15em}{\includegraphics[height=1.0em]{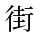}}\hskip0pt{}，\raisebox{-0.15em}{\includegraphics[height=1.0em]{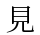}}\hskip0pt{}\raisebox{-0.15em}{\includegraphics[height=1.0em]{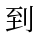}}\hskip0pt{}\raisebox{-0.15em}{\includegraphics[height=1.0em]{images/U+4E00.png}}\hskip0pt{}\raisebox{-0.15em}{\includegraphics[height=1.0em]{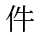}}\hskip0pt{}\raisebox{-0.15em}{\includegraphics[height=1.0em]{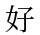}}\hskip0pt{}\raisebox{-0.15em}{\includegraphics[height=1.0em]{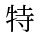}}\hskip0pt{}\raisebox{-0.15em}{\includegraphics[height=1.0em]{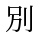}}\hskip0pt{}\raisebox{-0.15em}{\includegraphics[height=1.0em]{images/U+5605.png}}\hskip0pt{}\raisebox{-0.15em}{\includegraphics[height=1.0em]{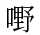}}\hskip0pt{}。\raisebox{-0.15em}{\includegraphics[height=1.0em]{images/U+4F62.png}}\hskip0pt{}\raisebox{-0.15em}{\includegraphics[height=1.0em]{images/U+597D.png}}\hskip0pt{}\raisebox{-0.15em}{\includegraphics[height=1.0em]{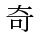}}\hskip0pt{}\raisebox{-0.15em}{\includegraphics[height=1.0em]{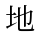}}\hskip0pt{}\raisebox{-0.15em}{\includegraphics[height=1.0em]{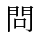}}\hskip0pt{}：「\raisebox{-0.15em}{\includegraphics[height=1.0em]{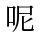}}\hskip0pt{}\raisebox{-0.15em}{\includegraphics[height=1.0em]{images/U+500B.png}}\hskip0pt{}\raisebox{-0.15em}{\includegraphics[height=1.0em]{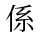}}\hskip0pt{}\raisebox{-0.15em}{\includegraphics[height=1.0em]{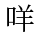}}\hskip0pt{}\raisebox{-0.15em}{\includegraphics[height=1.0em]{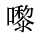}}\hskip0pt{}\raisebox{-0.15em}{\includegraphics[height=1.0em]{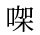}}\hskip0pt{}？」\raisebox{-0.15em}{\includegraphics[height=1.0em]{images/U+5ABD.png}}\hskip0pt{}\raisebox{-0.15em}{\includegraphics[height=1.0em]{images/U+5ABD.png}}\hskip0pt{}\raisebox{-0.15em}{\includegraphics[height=1.0em]{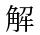}}\hskip0pt{}\raisebox{-0.15em}{\includegraphics[height=1.0em]{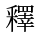}}\hskip0pt{}：「\raisebox{-0.15em}{\includegraphics[height=1.0em]{images/U+5462.png}}\hskip0pt{}\raisebox{-0.15em}{\includegraphics[height=1.0em]{images/U+500B.png}}\hskip0pt{}\raisebox{-0.15em}{\includegraphics[height=1.0em]{images/U+4FC2.png}}\hskip0pt{}\raisebox{-0.15em}{\includegraphics[height=1.0em]{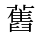}}\hskip0pt{}\raisebox{-0.15em}{\includegraphics[height=1.0em]{images/U+5605.png}}\hskip0pt{}\raisebox{-0.15em}{\includegraphics[height=1.0em]{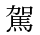}}\hskip0pt{}\raisebox{-0.15em}{\includegraphics[height=1.0em]{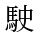}}\hskip0pt{}\raisebox{-0.15em}{\includegraphics[height=1.0em]{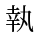}}\hskip0pt{}\raisebox{-0.15em}{\includegraphics[height=1.0em]{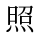}}\hskip0pt{}，\raisebox{-0.15em}{\includegraphics[height=1.0em]{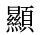}}\hskip0pt{}\raisebox{-0.15em}{\includegraphics[height=1.0em]{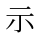}}\hskip0pt{}\raisebox{-0.15em}{\includegraphics[height=1.0em]{images/U+5462.png}}\hskip0pt{}\raisebox{-0.15em}{\includegraphics[height=1.0em]{images/U+500B.png}}\hskip0pt{}\raisebox{-0.15em}{\includegraphics[height=1.0em]{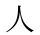}}\hskip0pt{}\raisebox{-0.15em}{\includegraphics[height=1.0em]{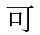}}\hskip0pt{}\raisebox{-0.15em}{\includegraphics[height=1.0em]{images/U+4EE5.png}}\hskip0pt{}\raisebox{-0.15em}{\includegraphics[height=1.0em]{images/U+99D5.png}}\hskip0pt{}\raisebox{-0.15em}{\includegraphics[height=1.0em]{images/U+99DB.png}}\hskip0pt{}。」 Gwen \raisebox{-0.15em}{\includegraphics[height=1.0em]{images/U+597D.png}}\hskip0pt{}\raisebox{-0.15em}{\includegraphics[height=1.0em]{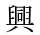}}\hskip0pt{}\raisebox{-0.15em}{\includegraphics[height=1.0em]{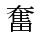}}\hskip0pt{}\raisebox{-0.15em}{\includegraphics[height=1.0em]{images/U+5730.png}}\hskip0pt{}\raisebox{-0.15em}{\includegraphics[height=1.0em]{images/U+554F.png}}\hskip0pt{}：「\raisebox{-0.15em}{\includegraphics[height=1.0em]{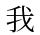}}\hskip0pt{}\raisebox{-0.15em}{\includegraphics[height=1.0em]{images/U+53EF.png}}\hskip0pt{}\raisebox{-0.15em}{\includegraphics[height=1.0em]{images/U+4EE5.png}}\hskip0pt{}\raisebox{-0.15em}{\includegraphics[height=1.0em]{images/U+6709.png}}\hskip0pt{}\raisebox{-0.15em}{\includegraphics[height=1.0em]{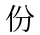}}\hskip0pt{}\raisebox{-0.15em}{\includegraphics[height=1.0em]{images/U+35CE.png}}\hskip0pt{}？！」\raisebox{-0.15em}{\includegraphics[height=1.0em]{images/U+5ABD.png}}\hskip0pt{}\raisebox{-0.15em}{\includegraphics[height=1.0em]{images/U+5ABD.png}}\hskip0pt{}\raisebox{-0.15em}{\includegraphics[height=1.0em]{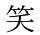}}\hskip0pt{}\raisebox{-0.15em}{\includegraphics[height=1.0em]{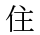}}\hskip0pt{}\raisebox{-0.15em}{\includegraphics[height=1.0em]{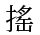}}\hskip0pt{}\raisebox{-0.15em}{\includegraphics[height=1.0em]{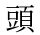}}\hskip0pt{}\raisebox{-0.15em}{\includegraphics[height=1.0em]{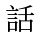}}\hskip0pt{}：「\raisebox{-0.15em}{\includegraphics[height=1.0em]{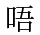}}\hskip0pt{}\raisebox{-0.15em}{\includegraphics[height=1.0em]{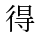}}\hskip0pt{}，\raisebox{-0.15em}{\includegraphics[height=1.0em]{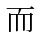}}\hskip0pt{}\raisebox{-0.15em}{\includegraphics[height=1.0em]{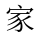}}\hskip0pt{}\raisebox{-0.15em}{\includegraphics[height=1.0em]{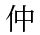}}\hskip0pt{}\raisebox{-0.15em}{\includegraphics[height=1.0em]{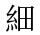}}\hskip0pt{}\raisebox{-0.15em}{\includegraphics[height=1.0em]{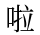}}\hskip0pt{}。
\end{quote}

The age of a person was counted using \raisebox{-0.15em}{\includegraphics[height=1.0em]{images/U+5E74.png}}\hskip0pt{} (year) instead of the correct word \raisebox{-0.15em}{\includegraphics[height=1.0em]{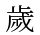}}\hskip0pt{} (years of age).

Another example is incomprehensible:

English:
\begin{quote}
    Once upon a time, there was a big dinosaur. He was very fast and could run really quickly. One day, the dinosaur went for a walk and saw a little boy. The boy was sad because he lost his toy car. The dinosaur felt sorry for the boy and decided to help him. He ran very fast to search for the toy car. After a while, the dinosaur found the toy car and returned it to the little boy. The boy was happy and thanked the dinosaur for being so kind. From that day on, the boy and the dinosaur became good friends and went on many adventures together.
\end{quote}
Cantonese translation:
\begin{quote}
    \raisebox{-0.15em}{\includegraphics[height=1.0em]{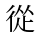}}\hskip0pt{}\raisebox{-0.15em}{\includegraphics[height=1.0em]{images/U+524D.png}}\hskip0pt{}\raisebox{-0.15em}{\includegraphics[height=1.0em]{images/U+6709.png}}\hskip0pt{}\raisebox{-0.15em}{\includegraphics[height=1.0em]{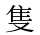}}\hskip0pt{}\raisebox{-0.15em}{\includegraphics[height=1.0em]{images/U+597D.png}}\hskip0pt{}\raisebox{-0.15em}{\includegraphics[height=1.0em]{images/U+5927.png}}\hskip0pt{}\raisebox{-0.15em}{\includegraphics[height=1.0em]{images/U+5605.png}}\hskip0pt{}\raisebox{-0.15em}{\includegraphics[height=1.0em]{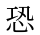}}\hskip0pt{}\raisebox{-0.15em}{\includegraphics[height=1.0em]{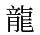}}\hskip0pt{}，\raisebox{-0.15em}{\includegraphics[height=1.0em]{images/U+4F62.png}}\hskip0pt{}\raisebox{-0.15em}{\includegraphics[height=1.0em]{images/U+597D.png}}\hskip0pt{}\raisebox{-0.15em}{\includegraphics[height=1.0em]{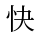}}\hskip0pt{}，\raisebox{-0.15em}{\includegraphics[height=1.0em]{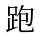}}\hskip0pt{}\raisebox{-0.15em}{\includegraphics[height=1.0em]{images/U+5F97.png}}\hskip0pt{}\raisebox{-0.15em}{\includegraphics[height=1.0em]{images/U+597D.png}}\hskip0pt{}\raisebox{-0.15em}{\includegraphics[height=1.0em]{images/U+5FEB.png}}\hskip0pt{}。\raisebox{-0.15em}{\includegraphics[height=1.0em]{images/U+6709.png}}\hskip0pt{}\raisebox{-0.15em}{\includegraphics[height=1.0em]{images/U+4E00.png}}\hskip0pt{}\raisebox{-0.15em}{\includegraphics[height=1.0em]{images/U+65E5.png}}\hskip0pt{}，\raisebox{-0.15em}{\includegraphics[height=1.0em]{images/U+6050.png}}\hskip0pt{}\raisebox{-0.15em}{\includegraphics[height=1.0em]{images/U+9F8D.png}}\hskip0pt{}\raisebox{-0.15em}{\includegraphics[height=1.0em]{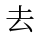}}\hskip0pt{}\raisebox{-0.15em}{\includegraphics[height=1.0em]{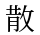}}\hskip0pt{}\raisebox{-0.15em}{\includegraphics[height=1.0em]{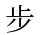}}\hskip0pt{}，\raisebox{-0.15em}{\includegraphics[height=1.0em]{images/U+898B.png}}\hskip0pt{}\raisebox{-0.15em}{\includegraphics[height=1.0em]{images/U+5230.png}}\hskip0pt{}\raisebox{-0.15em}{\includegraphics[height=1.0em]{images/U+4E00.png}}\hskip0pt{}\raisebox{-0.15em}{\includegraphics[height=1.0em]{images/U+500B.png}}\hskip0pt{}\raisebox{-0.15em}{\includegraphics[height=1.0em]{images/U+5C0F.png}}\hskip0pt{}\raisebox{-0.15em}{\includegraphics[height=1.0em]{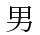}}\hskip0pt{}\raisebox{-0.15em}{\includegraphics[height=1.0em]{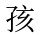}}\hskip0pt{}。\raisebox{-0.15em}{\includegraphics[height=1.0em]{images/U+500B.png}}\hskip0pt{}\raisebox{-0.15em}{\includegraphics[height=1.0em]{images/U+7537.png}}\hskip0pt{}\raisebox{-0.15em}{\includegraphics[height=1.0em]{images/U+5B69.png}}\hskip0pt{}\raisebox{-0.15em}{\includegraphics[height=1.0em]{images/U+597D.png}}\hskip0pt{}\raisebox{-0.15em}{\includegraphics[height=1.0em]{images/U+5514.png}}\hskip0pt{}\raisebox{-0.15em}{\includegraphics[height=1.0em]{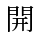}}\hskip0pt{}\raisebox{-0.15em}{\includegraphics[height=1.0em]{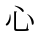}}\hskip0pt{}，\raisebox{-0.15em}{\includegraphics[height=1.0em]{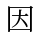}}\hskip0pt{}\raisebox{-0.15em}{\includegraphics[height=1.0em]{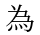}}\hskip0pt{}\raisebox{-0.15em}{\includegraphics[height=1.0em]{images/U+4F62.png}}\hskip0pt{}\raisebox{-0.15em}{\includegraphics[height=1.0em]{images/U+5514.png}}\hskip0pt{}\raisebox{-0.15em}{\includegraphics[height=1.0em]{images/U+898B.png}}\hskip0pt{}\raisebox{-0.15em}{\includegraphics[height=1.0em]{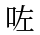}}\hskip0pt{}\raisebox{-0.15em}{\includegraphics[height=1.0em]{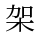}}\hskip0pt{}\raisebox{-0.15em}{\includegraphics[height=1.0em]{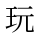}}\hskip0pt{}\raisebox{-0.15em}{\includegraphics[height=1.0em]{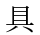}}\hskip0pt{}\raisebox{-0.15em}{\includegraphics[height=1.0em]{images/U+8ECA.png}}\hskip0pt{}。\raisebox{-0.15em}{\includegraphics[height=1.0em]{images/U+6050.png}}\hskip0pt{}\raisebox{-0.15em}{\includegraphics[height=1.0em]{images/U+9F8D.png}}\hskip0pt{}\raisebox{-0.15em}{\includegraphics[height=1.0em]{images/U+597D.png}}\hskip0pt{}\raisebox{-0.15em}{\includegraphics[height=1.0em]{images/U+540C.png}}\hskip0pt{}\raisebox{-0.15em}{\includegraphics[height=1.0em]{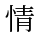}}\hskip0pt{}\raisebox{-0.15em}{\includegraphics[height=1.0em]{images/U+500B.png}}\hskip0pt{}\raisebox{-0.15em}{\includegraphics[height=1.0em]{images/U+7537.png}}\hskip0pt{}\raisebox{-0.15em}{\includegraphics[height=1.0em]{images/U+5B69.png}}\hskip0pt{}，\raisebox{-0.15em}{\includegraphics[height=1.0em]{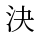}}\hskip0pt{}\raisebox{-0.15em}{\includegraphics[height=1.0em]{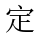}}\hskip0pt{}\raisebox{-0.15em}{\includegraphics[height=1.0em]{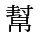}}\hskip0pt{}\raisebox{-0.15em}{\includegraphics[height=1.0em]{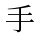}}\hskip0pt{}。\raisebox{-0.15em}{\includegraphics[height=1.0em]{images/U+4F62.png}}\hskip0pt{}\raisebox{-0.15em}{\includegraphics[height=1.0em]{images/U+597D.png}}\hskip0pt{}\raisebox{-0.15em}{\includegraphics[height=1.0em]{images/U+5FEB.png}}\hskip0pt{}\raisebox{-0.15em}{\includegraphics[height=1.0em]{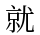}}\hskip0pt{}\raisebox{-0.15em}{\includegraphics[height=1.0em]{images/U+8DD1.png}}\hskip0pt{}\raisebox{-0.15em}{\includegraphics[height=1.0em]{images/U+53BB.png}}\hskip0pt{}\raisebox{-0.15em}{\includegraphics[height=1.0em]{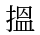}}\hskip0pt{}\raisebox{-0.15em}{\includegraphics[height=1.0em]{images/U+73A9.png}}\hskip0pt{}\raisebox{-0.15em}{\includegraphics[height=1.0em]{images/U+5177.png}}\hskip0pt{}\raisebox{-0.15em}{\includegraphics[height=1.0em]{images/U+8ECA.png}}\hskip0pt{}。\raisebox{-0.15em}{\includegraphics[height=1.0em]{images/U+4E00.png}}\hskip0pt{}\raisebox{-0.15em}{\includegraphics[height=1.0em]{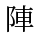}}\hskip0pt{}，\raisebox{-0.15em}{\includegraphics[height=1.0em]{images/U+6050.png}}\hskip0pt{}\raisebox{-0.15em}{\includegraphics[height=1.0em]{images/U+9F8D.png}}\hskip0pt{}\raisebox{-0.15em}{\includegraphics[height=1.0em]{images/U+6435.png}}\hskip0pt{}\raisebox{-0.15em}{\includegraphics[height=1.0em]{images/U+5230.png}}\hskip0pt{}\raisebox{-0.15em}{\includegraphics[height=1.0em]{images/U+73A9.png}}\hskip0pt{}\raisebox{-0.15em}{\includegraphics[height=1.0em]{images/U+5177.png}}\hskip0pt{}\raisebox{-0.15em}{\includegraphics[height=1.0em]{images/U+8ECA.png}}\hskip0pt{}，\raisebox{-0.15em}{\includegraphics[height=1.0em]{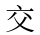}}\hskip0pt{}\raisebox{-0.15em}{\includegraphics[height=1.0em]{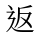}}\hskip0pt{}\raisebox{-0.15em}{\includegraphics[height=1.0em]{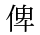}}\hskip0pt{}\raisebox{-0.15em}{\includegraphics[height=1.0em]{images/U+500B.png}}\hskip0pt{}\raisebox{-0.15em}{\includegraphics[height=1.0em]{images/U+5C0F.png}}\hskip0pt{}\raisebox{-0.15em}{\includegraphics[height=1.0em]{images/U+7537.png}}\hskip0pt{}\raisebox{-0.15em}{\includegraphics[height=1.0em]{images/U+5B69.png}}\hskip0pt{}。\raisebox{-0.15em}{\includegraphics[height=1.0em]{images/U+500B.png}}\hskip0pt{}\raisebox{-0.15em}{\includegraphics[height=1.0em]{images/U+7537.png}}\hskip0pt{}\raisebox{-0.15em}{\includegraphics[height=1.0em]{images/U+5B69.png}}\hskip0pt{}\raisebox{-0.15em}{\includegraphics[height=1.0em]{images/U+597D.png}}\hskip0pt{}\raisebox{-0.15em}{\includegraphics[height=1.0em]{images/U+958B.png}}\hskip0pt{}\raisebox{-0.15em}{\includegraphics[height=1.0em]{images/U+5FC3.png}}\hskip0pt{}，\raisebox{-0.15em}{\includegraphics[height=1.0em]{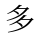}}\hskip0pt{}\raisebox{-0.15em}{\includegraphics[height=1.0em]{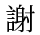}}\hskip0pt{}\raisebox{-0.15em}{\includegraphics[height=1.0em]{images/U+6050.png}}\hskip0pt{}\raisebox{-0.15em}{\includegraphics[height=1.0em]{images/U+9F8D.png}}\hskip0pt{}\raisebox{-0.15em}{\includegraphics[height=1.0em]{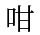}}\hskip0pt{}\raisebox{-0.15em}{\includegraphics[height=1.0em]{images/U+597D.png}}\hskip0pt{}\raisebox{-0.15em}{\includegraphics[height=1.0em]{images/U+5FC3.png}}\hskip0pt{}。\raisebox{-0.15em}{\includegraphics[height=1.0em]{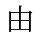}}\hskip0pt{}\raisebox{-0.15em}{\includegraphics[height=1.0em]{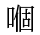}}\hskip0pt{}\raisebox{-0.15em}{\includegraphics[height=1.0em]{images/U+65E5.png}}\hskip0pt{}\raisebox{-0.15em}{\includegraphics[height=1.0em]{images/U+958B.png}}\hskip0pt{}\raisebox{-0.15em}{\includegraphics[height=1.0em]{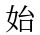}}\hskip0pt{}，\raisebox{-0.15em}{\includegraphics[height=1.0em]{images/U+500B.png}}\hskip0pt{}\raisebox{-0.15em}{\includegraphics[height=1.0em]{images/U+7537.png}}\hskip0pt{}\raisebox{-0.15em}{\includegraphics[height=1.0em]{images/U+5B69.png}}\hskip0pt{}\raisebox{-0.15em}{\includegraphics[height=1.0em]{images/U+540C.png}}\hskip0pt{}\raisebox{-0.15em}{\includegraphics[height=1.0em]{images/U+6050.png}}\hskip0pt{}\raisebox{-0.15em}{\includegraphics[height=1.0em]{images/U+9F8D.png}}\hskip0pt{}\raisebox{-0.15em}{\includegraphics[height=1.0em]{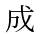}}\hskip0pt{}\raisebox{-0.15em}{\includegraphics[height=1.0em]{images/U+70BA.png}}\hskip0pt{}\raisebox{-0.15em}{\includegraphics[height=1.0em]{images/U+597D.png}}\hskip0pt{}\raisebox{-0.15em}{\includegraphics[height=1.0em]{images/U+670B.png}}\hskip0pt{}\raisebox{-0.15em}{\includegraphics[height=1.0em]{images/U+53CB.png}}\hskip0pt{}，\raisebox{-0.15em}{\includegraphics[height=1.0em]{images/U+4E00.png}}\hskip0pt{}\raisebox{-0.15em}{\includegraphics[height=1.0em]{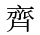}}\hskip0pt{}\raisebox{-0.15em}{\includegraphics[height=1.0em]{images/U+53BB.png}}\hskip0pt{}\raisebox{-0.15em}{\includegraphics[height=1.0em]{images/U+597D.png}}\hskip0pt{}\raisebox{-0.15em}{\includegraphics[height=1.0em]{images/U+591A.png}}\hskip0pt{}\raisebox{-0.15em}{\includegraphics[height=1.0em]{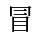}}\hskip0pt{}\raisebox{-0.15em}{\includegraphics[height=1.0em]{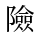}}\hskip0pt{}。
\end{quote}

The translation service broke down the original English sentences by punctuation and connectives (such as "and"). Then, it translated the broken-down phrases individually and joined them into a Cantonese/Chinese sentence, resulting in awkward and incorrect punctuation usage.
The heavy use of poor machine translation of existing work offered little in terms of novel contributions or insights into the language. 

The use of the BLEU score is inappropriate for Cantonese translation. Please see Appendix \ref{sec:appendix6} for our rationale.

\section{Translated MMLU Dataset}
\label{sec:appendix2}
The original MMLU dataset was translated into Cantonese by Gemini 1.5 Flash using the following prompt:

\begin{quote}
    \raisebox{-0.15em}{\includegraphics[height=1.0em]{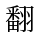}}\hskip0pt{}\raisebox{-0.15em}{\includegraphics[height=1.0em]{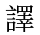}}\hskip0pt{}\raisebox{-0.15em}{\includegraphics[height=1.0em]{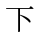}}\hskip0pt{}\raisebox{-0.15em}{\includegraphics[height=1.0em]{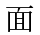}}\hskip0pt{}\raisebox{-0.15em}{\includegraphics[height=1.0em]{images/U+5605.png}}\hskip0pt{}\raisebox{-0.15em}{\includegraphics[height=1.0em]{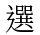}}\hskip0pt{}\raisebox{-0.15em}{\includegraphics[height=1.0em]{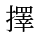}}\hskip0pt{}\raisebox{-0.15em}{\includegraphics[height=1.0em]{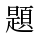}}\hskip0pt{}\raisebox{-0.15em}{\includegraphics[height=1.0em]{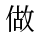}}\hskip0pt{}\raisebox{-0.15em}{\includegraphics[height=1.0em]{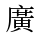}}\hskip0pt{}\raisebox{-0.15em}{\includegraphics[height=1.0em]{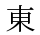}}\hskip0pt{}\raisebox{-0.15em}{\includegraphics[height=1.0em]{images/U+8A71.png}}\hskip0pt{}：\\

*** Input: (Example question 1) \\
*** A: (Example 1 option A)\\
*** B: (Example 1 option B)\\
*** C: (Example 1 option C)\\
*** D: (Example 1 option D)\\
*** Target: (Example 1 Answer)\\

\raisebox{-0.15em}{\includegraphics[height=1.0em]{images/U+5EE3.png}}\hskip0pt{}\raisebox{-0.15em}{\includegraphics[height=1.0em]{images/U+6771.png}}\hskip0pt{}\raisebox{-0.15em}{\includegraphics[height=1.0em]{images/U+8A71.png}}\hskip0pt{}\raisebox{-0.15em}{\includegraphics[height=1.0em]{images/U+7FFB.png}}\hskip0pt{}\raisebox{-0.15em}{\includegraphics[height=1.0em]{images/U+8B6F.png}}\hskip0pt{}：\\

*** \raisebox{-0.15em}{\includegraphics[height=1.0em]{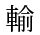}}\hskip0pt{}\raisebox{-0.15em}{\includegraphics[height=1.0em]{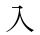}}\hskip0pt{}：(Manually translated Example Question 1)\\
*** \raisebox{-0.15em}{\includegraphics[height=1.0em]{images/U+9078.png}}\hskip0pt{}\raisebox{-0.15em}{\includegraphics[height=1.0em]{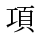}}\hskip0pt{}A：(Manually translated Example 1 option A)\\
*** \raisebox{-0.15em}{\includegraphics[height=1.0em]{images/U+9078.png}}\hskip0pt{}\raisebox{-0.15em}{\includegraphics[height=1.0em]{images/U+9805.png}}\hskip0pt{}B：(Manually translated Example 1 option B)\\
*** \raisebox{-0.15em}{\includegraphics[height=1.0em]{images/U+9078.png}}\hskip0pt{}\raisebox{-0.15em}{\includegraphics[height=1.0em]{images/U+9805.png}}\hskip0pt{}C：(Manually translated Example 1 option C)\\
*** \raisebox{-0.15em}{\includegraphics[height=1.0em]{images/U+9078.png}}\hskip0pt{}\raisebox{-0.15em}{\includegraphics[height=1.0em]{images/U+9805.png}}\hskip0pt{}D：(Manually translated Example 1 option D)\\
*** \raisebox{-0.15em}{\includegraphics[height=1.0em]{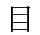}}\hskip0pt{}\raisebox{-0.15em}{\includegraphics[height=1.0em]{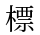}}\hskip0pt{}：(Example 1 Answer)\\

=================\\

(Another four examples)\\

=================\\

\raisebox{-0.15em}{\includegraphics[height=1.0em]{images/U+7FFB.png}}\hskip0pt{}\raisebox{-0.15em}{\includegraphics[height=1.0em]{images/U+8B6F.png}}\hskip0pt{}\raisebox{-0.15em}{\includegraphics[height=1.0em]{images/U+4E0B.png}}\hskip0pt{}\raisebox{-0.15em}{\includegraphics[height=1.0em]{images/U+9762.png}}\hskip0pt{}\raisebox{-0.15em}{\includegraphics[height=1.0em]{images/U+5605.png}}\hskip0pt{}\raisebox{-0.15em}{\includegraphics[height=1.0em]{images/U+9078.png}}\hskip0pt{}\raisebox{-0.15em}{\includegraphics[height=1.0em]{images/U+64C7.png}}\hskip0pt{}\raisebox{-0.15em}{\includegraphics[height=1.0em]{images/U+984C.png}}\hskip0pt{}\raisebox{-0.15em}{\includegraphics[height=1.0em]{images/U+505A.png}}\hskip0pt{}\raisebox{-0.15em}{\includegraphics[height=1.0em]{images/U+5EE3.png}}\hskip0pt{}\raisebox{-0.15em}{\includegraphics[height=1.0em]{images/U+6771.png}}\hskip0pt{}\raisebox{-0.15em}{\includegraphics[height=1.0em]{images/U+8A71.png}}\hskip0pt{}：\\
*** Input: (Input Question)\\
*** A: (Option A)\\
*** B: (Option B)\\
*** C: (Option C)\\
*** D: (Option D)\\
*** Target: (Answer)\\

\end{quote}

English translation of the prompt:

\begin{quote}
    Translate the following multiple choice question to Cantonese：\\

*** Input: (Example question 1) \\
*** A: (Example 1 option A)\\
*** B: (Example 1 option B)\\
*** C: (Example 1 option C)\\
*** D: (Example 1 option D)\\
*** Target: (Example 1 Answer)\\

Cantonese Translation：\\

*** Input：(Manually translated Example Question 1)\\
*** Option A：(Manually translated Example 1 option A)\\
*** Option B：(Manually translated Example 1 option B)\\
*** Option C：(Manually translated Example 1 option C)\\
*** Option D：(Manually translated Example 1 option D)\\
*** Target：(Example 1 Answer)\\

=================\\

(Another four examples)\\

=================\\

Translate the following multiple choice question to Cantonese：\\
*** Input: (Input Question)\\
*** A: (Option A)\\
*** B: (Option B)\\
*** C: (Option C)\\
*** D: (Option D)\\
*** Target: (Answer)\\

\end{quote}

\section{Academic and Professional Dataset}
\label{sec:appendix3}
The Professional Dataset subset consists of 7 professional qualification exams:
\begin{enumerate}
    \item \textbf{Estate Agents Qualifying Exam (EAQE)}: Exam from the Estate Agents Authority (EAA) which grants an estate agent's license for a person to become a director or a partner of an estate agency.
    \item \textbf{Insurance Intermediaries Qualifying Examination}: The Insurance Authority in Hong Kong regulates and licenses all insurance intermediaries in Hong Kong and offers this exam for any person who wishes to carry out insurance activities.
    \item \textbf{Leveraged Foreign Exchange Trading Examination}: An industry qualification exam offered by the Vocational Training Council (VTC) and approved by the Securities and Future Commission.
    \item \textbf{Licensing Exam for Securities \& Futures Intermediaries}: An exam offered by the Hong Kong Securities and Investment Institute and recognised by the Securities and Futures Commission (SFC). It is a professional qualification exam for people who would like to work in the securities and investment industry in Hong Kong.
    \item \textbf{Mandatory Provident Fund Schemes (MPF)/Intermediaries Exam}: The MPF is the pension scheme in Hong Kong and any intermediaries involved in providing the service are required to pass this exam.
    \item \textbf{Pleasure Vessel Operator Grade 2 Certificate}: The Marine Department of Hong Kong offers this exam for anyone who wishes to operate a boat less than 15 meters in length for pleasure purposes.
    \item \textbf{EAA Salespersons Qualifying Examination}: Exam from the EAA which grants a salesperson's license for a salesperson to carry out estate agency work.
    \item \textbf{Taxi License Written Exam Route Planning Exam}: Unique questions created in the form of the Taxi License Written Exam Route Planning Exam which are multiple-choice questions about selecting the shortest route between any two places or landmarks without considering toll fees. 

\end{enumerate}
All the exams suggested above were sourced from PDF files provided by the exam provider and processed by Google Gemini 1.5 Flash. The pages containing the model answers were separated from the questions pages, which could mitigate the data contamination issue. The following sentence is added to the front of the prompt during the evaluation:
\begin{quote}
    Follow the given examples and answer the question. The question is about professional knowledge in Hong Kong. You should only return the answer: A, B, C, or D.
\end{quote}

The next two law-related categories were sourced from the Internet:

\begin{enumerate}
    \item \textbf{Hong Kong Law}: Questions across 15 categories such as child abuse, domestic violence, the Employment Ordinance, Equal Opportunities, family law, etc. were gathered across the Internet.
    \item \textbf{Basic Law}: Constitution-type question sourced from a website. The questions were updated according to the current and actual practice of the Basic Law and options were also appended to 4 options per question by the authors.
\end{enumerate}

A sentence describing the task is added at the front of the prompt:
\begin{quote}
    Follow the given examples and answer the question. The question is about Hong Kong law. You should only return the answer: A, B, C, or D.
\end{quote}

The number of questions and the source of each professional exam can be found in Table \ref{tab:prof-dataset-summary}.

\begin{table}[t]
  \label{professional-dataset-table}
  \centering
  \renewcommand{\arraystretch}{1.2}
\begin{tabular}{llrl}
Name                  & Source & \multicolumn{1}{l}{No.} & Eng. \\ \hline
Estate Agent          & PDF    & 50                      & Y    \\
Insurance Interm.     & PDF    & 130                     & Y    \\
Leveraged FX          & PDF    & 20                      & Y    \\
Securities \& Futures & PDF    & 400                     & Y    \\
MPF Interm.           & PDF    & 26                      & Y    \\
Pleasure Vessel       & PDF    & 66                      & Y    \\
Salesperson           & PDF    & 50                      & Y    \\
Taxi                  & New    & 30                      & N    \\
HK Law                & Web    & 360                     & N    \\
Basic Law             & Web    & 64                      & N    \\ \hline
Total                 &        & 1,196                   &     
\end{tabular}
\caption{\label{tab:prof-dataset-summary}Professional Dataset Summary}
\end{table}

The Academic Dataset subset was sourced from scanned copies of the HKDSE exam from the last 2 to 4 years. The details can be found in Table \ref{tab:academic-dataset-summary}. The scanned PDFs were processed to extract text with the Google Gemini 1.5 Pro API. The extracted text was then corrected and filtered by the authors to remove questions that required information from other questions. Those requiring references to visual materials like pictures and maps were also removed but reserved for future evaluation of vision-enabled LLMs.

Although only multiple-choice questions were included in the dataset, more complicated questions that require complex reasoning and human judgment, such as Chinese listening and Liberal Studies exams were studied. Initial testing showed proprietary LLMs can easily achieve perfect scores in the Chinese listening exam when given the transcript, while the length of the transcript exceeded some open-weight LLMs' context length. 
Further experiments were conducted with Google Gemini 1.5 Flash and Pro to take the audio file and written questions as input. Although the two models again achieved a perfect score, no other model supported audio input at the time of the experiment, so the task was excluded. 
For the now-defunct Liberal Studies, examinees were required to write a passage taking into account the provided reference materials and then develop their own point of view supported by examples from the candidates' own knowledge. 
However, initial testing showed that none of the LLMs were able to include their own examples despite being explicitly requested in the prompt. These tasks were therefore not included in the final evaluation.

During the evaluation, a short sentence is added to the front to explain the nature of the dataset:

\begin{quote}
    Follow the given examples and answer the question. The question is about Hong Kong DSE. You should only return the answer: A, B, C, or D.
\end{quote}

\begin{table}[t]
  \label{academic-dataset-table}
  \centering
\begin{tabular}{lrl}
Subject                                                                              & No.       & Eng. \\ \hline
Biology                                                                              & 52        & Y    \\
\begin{tabular}[c]{@{}l@{}}Business, Accounting \\ \& Financial Studies\end{tabular} & 50        & Y    \\
Chemistry                                                                            & 42        & Y    \\
Economics                                                                            & 46        & Y    \\
Geography                                                                            & 54        & Y    \\
\begin{tabular}[c]{@{}l@{}}Information \&\\ Comm. Tech.\end{tabular}                 & 73        & Y    \\ 
Mathematics                                                                          & 63        & Y    \\
Physics                                                                              & 30        & Y    \\
\begin{tabular}[c]{@{}l@{}}Tourism \& \\ Hospitality Studies\end{tabular}            & 56        & N    \\ \hline
Total                                                                                & 466       &     
\end{tabular}
\caption{\label{tab:academic-dataset-summary}Academic Dataset Summary}
\end{table}

\section{Hong Kong Cultural Questions Dataset}
\label{sec:appendix4}
The Hong Kong Cultural Questions Dataset contains four categories of newly created questions and one existing question set, which the breakdown can be found in Table \ref{tab:cultural-dataset-summary}.

\begin{enumerate}
    \item \textbf{Food Culture}: Hong Kong has a unique culinary and food culture due to cultural influences from East and West. The questions were designed to highlight local dishes, street food, high-end dining, and ``\raisebox{-0.15em}{\includegraphics[height=1.0em]{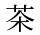}}\hskip0pt{}\raisebox{-0.15em}{\includegraphics[height=1.0em]{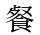}}\hskip0pt{}\raisebox{-0.15em}{\includegraphics[height=1.0em]{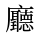}}\hskip0pt{}'' Cha chaan teng (Hong Kong-styled cafe) culture.
    
    \item \textbf{History and Landmarks}: Questions in this category assess the knowledge of significant historical events, figures, and iconic landmarks that have shaped Hong Kong's identity. 
    
    \item \textbf{Language and Expressions}: This category focuses on Cantonese-specific vocabulary, colloquialisms, slang, and common expressions used in daily life and social media networks.
    
    \item \textbf{Life in Hong Kong}: This category covers the customs, traditions, social norms, popular culture (TV and cinema), recent events and everyday life experiences unique to Hong Kong.
    
    \item \textbf{Local Area Knowledge}: Contains 33 questions on non-trivial local area knowledge selected from the Hong Kong Geography Tatsujin Challenge, an online questionnaire participated by over 100,000 participants. 
\end{enumerate}

A question bank of 244 questions, written in pure Cantonese, was created for the first four categories. To ensure the questions were based on common knowledge of the population, these questions were sent to 30 Hong Kong volunteers aged 20-50 years, and they were instructed not to use a search engine when attempting the questions. Consent was sought for using their data to inform the design of an LLM benchmark for Cantonese. Items with an accuracy under $40.0\%$  were excluded. For the last category, the questions were randomly selected from the Hong Kong Geography Tatsujin Challenge, which we have access to question-based correctness measures. The average correct rate of the selected questions is 53.0\% for this subset. The questions in this subcategory were in Written Chinese.

\begin{table}[t]
  \label{cultural-table}
  \centering
  \begin{tabular}{lll}
    Subject & No. of Q. & New Q.\\
    \midrule
    History \& Landmarks & 61 & Y\\
    Food Culture & 59 & Y\\
    Lang. \& Expressions & 49 & Y\\ 
    Life in HK & 75 & Y\\
    Local Area Knowledge & 33 & N\\\hline
    Total & 277
  \end{tabular} 
\caption{\label{tab:cultural-dataset-summary} Summary of the Hong Kong Cultural Questions Dataset}
\end{table}

The following sentence is added to the front of the prompt:
\begin{quote}
    Follow the given examples and answer the question. The question is about Hong Kong. Only return the answer: A, B, C, or D. DO NOT EXPLAIN.
\end{quote}

Given the unique and novel challenges presented by this dataset, a breakdown of model performance at five-shot evaluation is included in Table \ref{tab:cultural-perf-table}. The average score of the human evaluation was also included as a reference, where no model outperformed humans in the Food Culture and Language and Expressions category. SenseChat (Cantonese) refused to answer any questions in the History \& Landmark due to the appearance of the name of the Hong Kong Chief Executive John Lee in the 5-shot examples. 

\begin{table*}[p]
\centering
\begin{tabular}{l|c|c|c|c|c|c}
Model                                                                 & \begin{tabular}[c]{@{}c@{}}Food \\ Culture\end{tabular} & \begin{tabular}[c]{@{}c@{}}History \&\\ Landmarks\end{tabular} & \begin{tabular}[c]{@{}c@{}}Language \&\\ Expressions\end{tabular} & Life in HK & \begin{tabular}[c]{@{}c@{}}Local Area\\ Knowledge\end{tabular} & \multicolumn{1}{l}{Average} \\ \hline
Claude 3.5 Sonnet                                                     & 74.1\%                                                  & 80.4\%                                                         & 70.5\%                                                            & \textbf{85.7\%}     & 64.3\%                        & 75.0\%                      \\
Doubao Pro                                                            & 81.5\%                                                  & 82.1\%                                                         & 77.3\%                                                            & 77.1\%     & 57.1\%                        & 75.0\%                      \\
Ernie 4.0                                                             & 74.1\%                                                  & 75.0\%                                                         & 75.0\%                                                            & 84.3\%     & 67.9\%                        & 75.2\%                      \\
Gemini 1.5 Flash                                                      & 59.3\%                                                  & 69.6\%                                                         & 56.8\%                                                            & 77.1\%     & 57.1\%                        & 64.0\%                      \\
Gemini 1.5 Pro                                                        & 75.9\%                                                  & 82.1\%                                                         & 65.9\%                                                            & 84.3\%     & 60.7\%                        & 73.8\%                      \\
GPT4o                                                                 & 83.3\%                                                  & \textbf{83.9\%}                                                         & 68.2\%                                                            & 82.9\%     & \textbf{67.9\%}                        & 77.2\%                      \\
GPT4o-mini                                                            & 57.4\%                                                  & 67.9\%                                                         & 50.0\%                                                            & 74.3\%     & 53.6\%                        & 60.6\%                      \\
SenseChat                                                            & \textbf{87.0\%}                                                  & N.A.                                                         & \textbf{79.6\%}                                                            & 78.6\%     & 64.3\%                        & \textbf{77.4\%}                      \\ \hline
Aya 23 8B                                                             & 35.2\%                                                  & 46.4\%                                                         & 34.1\%                                                            & 44.3\%     & 28.6\%                        & 37.7\%                      \\
CLLM v0.5 6B                                                          & 51.9\%                                                  & 67.9\%                                                         & 63.6\%                                                            & 54.3\%     & 42.9\%                        & 56.1\%                      \\
CLLM v0.5 34B                                                         & \textbf{77.8\%}                                                  & \textbf{87.5\%}                                                         & \textbf{72.7\%}                                                            & 74.3\%     & 71.4\%                        & 76.7\%                      \\
Yi 1.5 6B                                                             & 40.7\%                                                  & 64.3\%                                                         & 50.0\%                                                            & 55.7\%     & 42.9\%                        & 50.7\%                      \\
Yi 1.5 9B                                                             & 53.7\%                                                  & 60.7\%                                                         & 52.3\%                                                            & 70.0\%     & 50.0\%                        & 57.3\%                      \\
Yi 1.5 34B                                                            & 74.1\%                                                  & 82.1\%                                                         & 61.4\%                                                            & 75.7\%     & 71.4\%                        & 72.9\%                      \\
Gemma 2 2B                                                            & 29.6\%                                                  & 37.5\%                                                         & 34.1\%                                                            & 42.9\%     & 32.1\%                        & 35.2\%                      \\
Gemma 2 9B                                                            & 48.2\%                                                  & 57.1\%                                                         & 50.0\%                                                            & 60.0\%     & 42.9\%                        & 51.6\%                      \\
Gemma 2 27B                                                           & 61.1\%                                                  & 69.6\%                                                         & 56.8\%                                                            & 67.1\%     & 50.0\%                        & 60.9\%                      \\
Llama 3.1 8B                                                          & 51.9\%                                                  & 57.1\%                                                         & 50.0\%                                                            & 54.3\%     & 50.0\%                        & 52.7\%                      \\
Llama 3.1 70B                                                         & 61.1\%                                                  & 78.6\%                                                         & 52.3\%                                                            & 72.9\%     & 57.1\%                        & 64.4\%                      \\
Llama 3.1 405B                                                        & 64.8\%                                                  & 75.0\%                                                         & 59.1\%                                                            & 75.7\%     & 75.0\%                        & 69.9\%                      \\
Mistral Nemo 12B & 31.5\%                                                  & 53.6\%                                                         & 45.5\%                                                            & 54.3\%     & 28.6\%                        & 42.7\%                      \\
Qwen2 7B                                                              & 55.6\%                                                  & 67.9\%                                                         & 50.0\%                                                            & 68.6\%     & 32.1\%                        & 54.8\%                      \\
Qwen2 72B                                                             & 75.9\%                                                  & 82.1\%                                                         & \textbf{72.7\%}                                                            & \textbf{82.9\%}     & \textbf{75.0\%}                        & \textbf{77.7\%}                      \\
Random                                                                & 35.2\%                                                  & 21.4\%                                                         & 27.3\%                                                            & 31.4\%     & 25.0\%                        & 28.1\%                      \\ \hline
Human                                                                 & 81.7\%                                                  & 73.4\%                                                         & 85.8\%                                                            & 83.9\%     & 53.0\%                        & 75.6\%                     
\end{tabular}
\caption{\label{tab:cultural-perf-table}Model performance on the Hong Kong Cultural Questions Dataset in 5-shots settings}
\end{table*}

\section{Linguistic Knowledge Dataset}

This dataset comprises three sub-datasets with questions carefully designed with expert input from researchers specialising in different areas of Cantonese linguistics. The structure and questions of each dataset are outlined below. 

\subsection{Phonological Knowledge Dataset}
\label{sec:appendix5_1}
The Phonological Knowledge Dataset include three groups of questions: Homophone Judgment (25 questions), Rhyme Judgment (25 questions), and a group of Phonological Reasoning Tasks (MultiPron Resolution, Tone Matching, Poetry Rhyme, Shared Feature Judgment, and Couplet Reasoning, 50 questions in total).

\begin{enumerate}
    \item \textbf{Homophone Judgment}: The task is to determine which character from a list shares the same pronunciation in Cantonese as the given character, if any. For example, if the given character is \raisebox{-0.15em}{\includegraphics[height=1.0em]{images/U+4E00.png}}\hskip0pt{} and the list of characters is A \raisebox{-0.15em}{\includegraphics[height=1.0em]{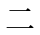}}\hskip0pt{}, B \raisebox{-0.15em}{\includegraphics[height=1.0em]{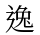}}\hskip0pt{}, C \raisebox{-0.15em}{\includegraphics[height=1.0em]{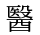}}\hskip0pt{}, D \raisebox{-0.15em}{\includegraphics[height=1.0em]{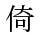}}\hskip0pt{}, the answer would be none of the above, as none of the characters (ji6, jat6, ji1, ji2) is a homophone of the given character (jat1). 
    \item \textbf{Rhyme Judgment}: The task is to determine which character from a list rhymes with a given character in Cantonese. For example, if the given character is \raisebox{-0.15em}{\includegraphics[height=1.0em]{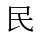}}\hskip0pt{} and the list is A \raisebox{-0.15em}{\includegraphics[height=1.0em]{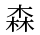}}\hskip0pt{}, B \raisebox{-0.15em}{\includegraphics[height=1.0em]{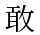}}\hskip0pt{}, C \raisebox{-0.15em}{\includegraphics[height=1.0em]{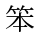}}\hskip0pt{}, D \raisebox{-0.15em}{\includegraphics[height=1.0em]{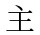}}\hskip0pt{}, the answer will be C, as it is pronounced b\textbf{an}6 in Cantonese, which rhymes with the given character m\textbf{an}4.
   
    \item[3a.] \textbf{MultiPron Resolution}: A word that contains a character that can be pronounced in multiple ways (like the English word read can be pronounced as red or reed), and the task is to decide how this character should be pronounced. For example, if the given word is \raisebox{-0.15em}{\includegraphics[height=1.0em]{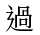}}\hskip0pt{}\raisebox{-0.15em}{\includegraphics[height=1.0em]{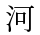}}\hskip0pt{} and the question asks for a homophone for the 2nd character, given the list A \raisebox{-0.15em}{\includegraphics[height=1.0em]{images/U+53EF.png}}\hskip0pt{}  B \raisebox{-0.15em}{\includegraphics[height=1.0em]{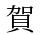}}\hskip0pt{} C \raisebox{-0.15em}{\includegraphics[height=1.0em]{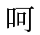}}\hskip0pt{} D \raisebox{-0.15em}{\includegraphics[height=1.0em]{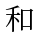}}\hskip0pt{}, the answer is none of the above. The given character should be pronounced as ho4 and none of the answer (ho2, ho6, ho1, wo4) matches this pronunciation.
    
    \item[3b.] \textbf{Tone Matching}: Given a polysyllabic word, choose a word from a list that matches the tonal (pitch) pattern based on tone-melodic requirement. For example, given the word \raisebox{-0.15em}{\includegraphics[height=1.0em]{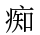}}\hskip0pt{}\raisebox{-0.15em}{\includegraphics[height=1.0em]{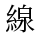}}\hskip0pt{}, and the list A \raisebox{-0.15em}{\includegraphics[height=1.0em]{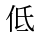}}\hskip0pt{}B, B \raisebox{-0.15em}{\includegraphics[height=1.0em]{images/U+591A.png}}\hskip0pt{}\raisebox{-0.15em}{\includegraphics[height=1.0em]{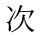}}\hskip0pt{}, C \raisebox{-0.15em}{\includegraphics[height=1.0em]{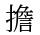}}\hskip0pt{}\raisebox{-0.15em}{\includegraphics[height=1.0em]{images/U+5FC3.png}}\hskip0pt{}, D \raisebox{-0.15em}{\includegraphics[height=1.0em]{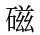}}\hskip0pt{}\raisebox{-0.15em}{\includegraphics[height=1.0em]{images/U+7DDA.png}}\hskip0pt{}, the answer is B. Its pronunciation is do\textbf{1}ci\textbf{3}, which is a high tone followed by a mid tone, same as the given word ci\textbf{1}sin\textbf{3}. 
    
    \item[3c.] \textbf{Poetry Rhyme}: Given a poem, choose the correct combination of rhyming characters, e.g. in the poem \raisebox{-0.15em}{\includegraphics[height=1.0em]{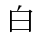}}\hskip0pt{}\raisebox{-0.15em}{\includegraphics[height=1.0em]{images/U+65E5.png}}\hskip0pt{}\raisebox{-0.15em}{\includegraphics[height=1.0em]{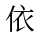}}\hskip0pt{}\raisebox{-0.15em}{\includegraphics[height=1.0em]{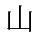}}\hskip0pt{}\raisebox{-0.15em}{\includegraphics[height=1.0em]{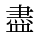}}\hskip0pt{}，\raisebox{-0.15em}{\includegraphics[height=1.0em]{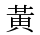}}\hskip0pt{}\raisebox{-0.15em}{\includegraphics[height=1.0em]{images/U+6CB3.png}}\hskip0pt{}\raisebox{-0.15em}{\includegraphics[height=1.0em]{images/U+5165.png}}\hskip0pt{}\raisebox{-0.15em}{\includegraphics[height=1.0em]{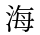}}\hskip0pt{}\raisebox{-0.15em}{\includegraphics[height=1.0em]{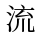}}\hskip0pt{}，\raisebox{-0.15em}{\includegraphics[height=1.0em]{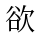}}\hskip0pt{}\raisebox{-0.15em}{\includegraphics[height=1.0em]{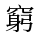}}\hskip0pt{}\raisebox{-0.15em}{\includegraphics[height=1.0em]{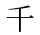}}\hskip0pt{}\raisebox{-0.15em}{\includegraphics[height=1.0em]{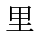}}\hskip0pt{}\raisebox{-0.15em}{\includegraphics[height=1.0em]{images/U+76EE.png}}\hskip0pt{}，\raisebox{-0.15em}{\includegraphics[height=1.0em]{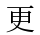}}\hskip0pt{}\raisebox{-0.15em}{\includegraphics[height=1.0em]{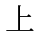}}\hskip0pt{}\raisebox{-0.15em}{\includegraphics[height=1.0em]{images/U+4E00.png}}\hskip0pt{}\raisebox{-0.15em}{\includegraphics[height=1.0em]{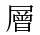}}\hskip0pt{}\raisebox{-0.15em}{\includegraphics[height=1.0em]{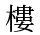}}\hskip0pt{}。 The last syllables are z\textbf{eon}6, l\textbf{au}4, m\textbf{uk}6, l\textbf{au}4 in Cantonese, \raisebox{-0.15em}{\includegraphics[height=1.0em]{images/U+6D41.png}}\hskip0pt{} and \raisebox{-0.15em}{\includegraphics[height=1.0em]{images/U+6A13.png}}\hskip0pt{} are the two characters that rhyme in the poem when recited in Cantonese.
    
    \item[3d.] \textbf{Shared Feature Judgment}: The task is to choose an odd character from a list that does not share a common phonological feature. For example, the list A \raisebox{-0.15em}{\includegraphics[height=1.0em]{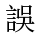}}\hskip0pt{} B \raisebox{-0.15em}{\includegraphics[height=1.0em]{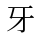}}\hskip0pt{} C \raisebox{-0.15em}{\includegraphics[height=1.0em]{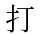}}\hskip0pt{} D \raisebox{-0.15em}{\includegraphics[height=1.0em]{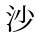}}\hskip0pt{} is ng6, ng\textbf{aa}4, d\textbf{aa}2, s\textbf{aa}1 in Jyutping. The last three characters share the same rhyme aa, so A is the odd one. 
    
    \item[3e.] \textbf{Couplet Reasoning}: One line of a couplet will be given. Characters in a couplet can be divided into two rhythmic categories, \raisebox{-0.15em}{\includegraphics[height=1.0em]{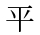}}\hskip0pt{} (ping4, \textbf{L}evel) and \raisebox{-0.15em}{\includegraphics[height=1.0em]{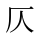}}\hskip0pt{} (zak1, \textbf{O}blique). This distinction is taught at school and is a longstanding poetry tradition. The task is to determine the category for each of the characters, and decide whether it is the upper (\raisebox{-0.15em}{\includegraphics[height=1.0em]{images/U+4E0A.png}}\hskip0pt{}\raisebox{-0.15em}{\includegraphics[height=1.0em]{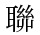}}\hskip0pt{}) or lower (\raisebox{-0.15em}{\includegraphics[height=1.0em]{images/U+4E0B.png}}\hskip0pt{}\raisebox{-0.15em}{\includegraphics[height=1.0em]{images/U+806F.png}}\hskip0pt{}) line. For example, the line \raisebox{-0.15em}{\includegraphics[height=1.0em]{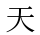}}\hskip0pt{}\raisebox{-0.15em}{\includegraphics[height=1.0em]{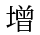}}\hskip0pt{}\raisebox{-0.15em}{\includegraphics[height=1.0em]{images/U+6B72.png}}\hskip0pt{}\raisebox{-0.15em}{\includegraphics[height=1.0em]{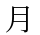}}\hskip0pt{}\raisebox{-0.15em}{\includegraphics[height=1.0em]{images/U+4EBA.png}}\hskip0pt{}\raisebox{-0.15em}{\includegraphics[height=1.0em]{images/U+589E.png}}\hskip0pt{}\raisebox{-0.15em}{\includegraphics[height=1.0em]{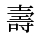}}\hskip0pt{} is tin\textbf{1} zang\textbf{1} seoi\underline{3} jyut\underline{6} jan\textbf{4} zang\textbf{1} sau\underline{6} in Jyutping, and its pattern is LLOOLLO, and as it ends with an oblique sound, it is the upper line. Therefore it is described as \raisebox{-0.15em}{\includegraphics[height=1.0em]{images/U+4E0A.png}}\hskip0pt{}\raisebox{-0.15em}{\includegraphics[height=1.0em]{images/U+806F.png}}\hskip0pt{}、\raisebox{-0.15em}{\includegraphics[height=1.0em]{images/U+5E73.png}}\hskip0pt{}\raisebox{-0.15em}{\includegraphics[height=1.0em]{images/U+5E73.png}}\hskip0pt{}\raisebox{-0.15em}{\includegraphics[height=1.0em]{images/U+4EC4.png}}\hskip0pt{}\raisebox{-0.15em}{\includegraphics[height=1.0em]{images/U+4EC4.png}}\hskip0pt{}\raisebox{-0.15em}{\includegraphics[height=1.0em]{images/U+5E73.png}}\hskip0pt{}\raisebox{-0.15em}{\includegraphics[height=1.0em]{images/U+5E73.png}}\hskip0pt{}\raisebox{-0.15em}{\includegraphics[height=1.0em]{images/U+4EC4.png}}\hskip0pt{}.
\end{enumerate}

The questions were evaluated using the following system prompt:
\begin{quote}
    You are a speaker of Cantonese from Hong Kong. Please answer these questions about the sounds of the language. Do not include any further explanation.
\end{quote}

Example questions for each task:
\begin{enumerate}
    \item Which of the following character is a homophone of \`{}\raisebox{-0.15em}{\includegraphics[height=1.0em]{images/U+4E00.png}}\hskip0pt{}\`{} in Cantonese? (A) \raisebox{-0.15em}{\includegraphics[height=1.0em]{images/U+4E8C.png}}\hskip0pt{} (B) \raisebox{-0.15em}{\includegraphics[height=1.0em]{images/U+9038.png}}\hskip0pt{} (C) \raisebox{-0.15em}{\includegraphics[height=1.0em]{images/U+91AB.png}}\hskip0pt{} (D) \raisebox{-0.15em}{\includegraphics[height=1.0em]{images/U+501A.png}}\hskip0pt{} \textbf{(E) None of the above}
    \item Which of the following character rhymes with the character \`{}\raisebox{-0.15em}{\includegraphics[height=1.0em]{images/U+6C11.png}}\hskip0pt{}\`{} in Cantonese? (A) \raisebox{-0.15em}{\includegraphics[height=1.0em]{images/U+68EE.png}}\hskip0pt{} (B) \raisebox{-0.15em}{\includegraphics[height=1.0em]{images/U+6562.png}}\hskip0pt{} \textbf{(C) \raisebox{-0.15em}{\includegraphics[height=1.0em]{images/U+7B28.png}}\hskip0pt{}} (D) \raisebox{-0.15em}{\includegraphics[height=1.0em]{images/U+4E3B.png}}\hskip0pt{} (E) None of the above
    \item[3a.] Which of the following character best represents the sound of \`{}\raisebox{-0.15em}{\includegraphics[height=1.0em]{images/U+6CB3.png}}\hskip0pt{}\`{} in the word \`{}\raisebox{-0.15em}{\includegraphics[height=1.0em]{images/U+904E.png}}\hskip0pt{}\raisebox{-0.15em}{\includegraphics[height=1.0em]{images/U+6CB3.png}}\hskip0pt{}\`{} in Cantonese? (A) \raisebox{-0.15em}{\includegraphics[height=1.0em]{images/U+53EF.png}}\hskip0pt{} (B) \raisebox{-0.15em}{\includegraphics[height=1.0em]{images/U+8CC0.png}}\hskip0pt{} (C) \raisebox{-0.15em}{\includegraphics[height=1.0em]{images/U+5475.png}}\hskip0pt{} (D) \raisebox{-0.15em}{\includegraphics[height=1.0em]{images/U+548C.png}}\hskip0pt{} \textbf{(E) None of the above}
    \item[3b.] Which of the following words have the nearest tone or pitch pattern with \`{}\raisebox{-0.15em}{\includegraphics[height=1.0em]{images/U+75F4.png}}\hskip0pt{}\raisebox{-0.15em}{\includegraphics[height=1.0em]{images/U+7DDA.png}}\hskip0pt{}\`{} in Cantonese? (A) \raisebox{-0.15em}{\includegraphics[height=1.0em]{images/U+4F4E.png}}\hskip0pt{}B \textbf{(B) \raisebox{-0.15em}{\includegraphics[height=1.0em]{images/U+591A.png}}\hskip0pt{}\raisebox{-0.15em}{\includegraphics[height=1.0em]{images/U+6B21.png}}\hskip0pt{}} (C) \raisebox{-0.15em}{\includegraphics[height=1.0em]{images/U+64D4.png}}\hskip0pt{}\raisebox{-0.15em}{\includegraphics[height=1.0em]{images/U+5FC3.png}}\hskip0pt{} (D) \raisebox{-0.15em}{\includegraphics[height=1.0em]{images/U+78C1.png}}\hskip0pt{}\raisebox{-0.15em}{\includegraphics[height=1.0em]{images/U+7DDA.png}}\hskip0pt{} (E) None of the above
    \item[3c.] In the following poem \\
\`{}\`{}\`{}\raisebox{-0.15em}{\includegraphics[height=1.0em]{images/U+767D.png}}\hskip0pt{}\raisebox{-0.15em}{\includegraphics[height=1.0em]{images/U+65E5.png}}\hskip0pt{}\raisebox{-0.15em}{\includegraphics[height=1.0em]{images/U+4F9D.png}}\hskip0pt{}\raisebox{-0.15em}{\includegraphics[height=1.0em]{images/U+5C71.png}}\hskip0pt{}\raisebox{-0.15em}{\includegraphics[height=1.0em]{images/U+76E1.png}}\hskip0pt{}，\\
\raisebox{-0.15em}{\includegraphics[height=1.0em]{images/U+9EC3.png}}\hskip0pt{}\raisebox{-0.15em}{\includegraphics[height=1.0em]{images/U+6CB3.png}}\hskip0pt{}\raisebox{-0.15em}{\includegraphics[height=1.0em]{images/U+5165.png}}\hskip0pt{}\raisebox{-0.15em}{\includegraphics[height=1.0em]{images/U+6D77.png}}\hskip0pt{}\raisebox{-0.15em}{\includegraphics[height=1.0em]{images/U+6D41.png}}\hskip0pt{}。\\
\raisebox{-0.15em}{\includegraphics[height=1.0em]{images/U+6B32.png}}\hskip0pt{}\raisebox{-0.15em}{\includegraphics[height=1.0em]{images/U+7AAE.png}}\hskip0pt{}\raisebox{-0.15em}{\includegraphics[height=1.0em]{images/U+5343.png}}\hskip0pt{}\raisebox{-0.15em}{\includegraphics[height=1.0em]{images/U+91CC.png}}\hskip0pt{}\raisebox{-0.15em}{\includegraphics[height=1.0em]{images/U+76EE.png}}\hskip0pt{}，\\
\raisebox{-0.15em}{\includegraphics[height=1.0em]{images/U+66F4.png}}\hskip0pt{}\raisebox{-0.15em}{\includegraphics[height=1.0em]{images/U+4E0A.png}}\hskip0pt{}\raisebox{-0.15em}{\includegraphics[height=1.0em]{images/U+4E00.png}}\hskip0pt{}\raisebox{-0.15em}{\includegraphics[height=1.0em]{images/U+5C64.png}}\hskip0pt{}\raisebox{-0.15em}{\includegraphics[height=1.0em]{images/U+6A13.png}}\hskip0pt{}。\`{}\`{}\`{} \\
What are the last words of the rhyming lines if recited in Cantonese? \`{}(A) \raisebox{-0.15em}{\includegraphics[height=1.0em]{images/U+76E1.png}}\hskip0pt{}\raisebox{-0.15em}{\includegraphics[height=1.0em]{images/U+6A13.png}}\hskip0pt{} (B) \raisebox{-0.15em}{\includegraphics[height=1.0em]{images/U+76E1.png}}\hskip0pt{}\raisebox{-0.15em}{\includegraphics[height=1.0em]{images/U+6D41.png}}\hskip0pt{}\raisebox{-0.15em}{\includegraphics[height=1.0em]{images/U+6A13.png}}\hskip0pt{} (C) \raisebox{-0.15em}{\includegraphics[height=1.0em]{images/U+6D41.png}}\hskip0pt{}\raisebox{-0.15em}{\includegraphics[height=1.0em]{images/U+76EE.png}}\hskip0pt{} \textbf{(D) \raisebox{-0.15em}{\includegraphics[height=1.0em]{images/U+6D41.png}}\hskip0pt{}\raisebox{-0.15em}{\includegraphics[height=1.0em]{images/U+6A13.png}}\hskip0pt{} }(E) None of the above\`{}

    \item[3d.] Three of the four characters below share a common phonological feature in Cantonese, and one does not. Which one is the odd one? \textbf{(A) \raisebox{-0.15em}{\includegraphics[height=1.0em]{images/U+8AA4.png}}\hskip0pt{}} (B) \raisebox{-0.15em}{\includegraphics[height=1.0em]{images/U+7259.png}}\hskip0pt{} (C) \raisebox{-0.15em}{\includegraphics[height=1.0em]{images/U+6253.png}}\hskip0pt{} (D) \raisebox{-0.15em}{\includegraphics[height=1.0em]{images/U+6C99.png}}\hskip0pt{} (E) None of the above

    \item[3e.] This is a line of a couplet (\raisebox{-0.15em}{\includegraphics[height=1.0em]{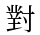}}\hskip0pt{}\raisebox{-0.15em}{\includegraphics[height=1.0em]{images/U+806F.png}}\hskip0pt{}): \`{}\raisebox{-0.15em}{\includegraphics[height=1.0em]{images/U+5929.png}}\hskip0pt{}\raisebox{-0.15em}{\includegraphics[height=1.0em]{images/U+589E.png}}\hskip0pt{}\raisebox{-0.15em}{\includegraphics[height=1.0em]{images/U+6B72.png}}\hskip0pt{}\raisebox{-0.15em}{\includegraphics[height=1.0em]{images/U+6708.png}}\hskip0pt{}\raisebox{-0.15em}{\includegraphics[height=1.0em]{images/U+4EBA.png}}\hskip0pt{}\raisebox{-0.15em}{\includegraphics[height=1.0em]{images/U+589E.png}}\hskip0pt{}\raisebox{-0.15em}{\includegraphics[height=1.0em]{images/U+58FD.png}}\hskip0pt{}\`{} \\
    Which of the following is a correct description of the line, according to its Cantonese pronunciation? \textbf{(A) \raisebox{-0.15em}{\includegraphics[height=1.0em]{images/U+4E0A.png}}\hskip0pt{}\raisebox{-0.15em}{\includegraphics[height=1.0em]{images/U+806F.png}}\hskip0pt{}、\raisebox{-0.15em}{\includegraphics[height=1.0em]{images/U+5E73.png}}\hskip0pt{}\raisebox{-0.15em}{\includegraphics[height=1.0em]{images/U+5E73.png}}\hskip0pt{}\raisebox{-0.15em}{\includegraphics[height=1.0em]{images/U+4EC4.png}}\hskip0pt{}\raisebox{-0.15em}{\includegraphics[height=1.0em]{images/U+4EC4.png}}\hskip0pt{}\raisebox{-0.15em}{\includegraphics[height=1.0em]{images/U+5E73.png}}\hskip0pt{}\raisebox{-0.15em}{\includegraphics[height=1.0em]{images/U+5E73.png}}\hskip0pt{}\raisebox{-0.15em}{\includegraphics[height=1.0em]{images/U+4EC4.png}}\hskip0pt{}} (B) \raisebox{-0.15em}{\includegraphics[height=1.0em]{images/U+4E0B.png}}\hskip0pt{}\raisebox{-0.15em}{\includegraphics[height=1.0em]{images/U+806F.png}}\hskip0pt{}、\raisebox{-0.15em}{\includegraphics[height=1.0em]{images/U+5E73.png}}\hskip0pt{}\raisebox{-0.15em}{\includegraphics[height=1.0em]{images/U+5E73.png}}\hskip0pt{}\raisebox{-0.15em}{\includegraphics[height=1.0em]{images/U+4EC4.png}}\hskip0pt{}\raisebox{-0.15em}{\includegraphics[height=1.0em]{images/U+4EC4.png}}\hskip0pt{}\raisebox{-0.15em}{\includegraphics[height=1.0em]{images/U+5E73.png}}\hskip0pt{}\raisebox{-0.15em}{\includegraphics[height=1.0em]{images/U+5E73.png}}\hskip0pt{}\raisebox{-0.15em}{\includegraphics[height=1.0em]{images/U+4EC4.png}}\hskip0pt{} (C) \raisebox{-0.15em}{\includegraphics[height=1.0em]{images/U+4E0A.png}}\hskip0pt{}\raisebox{-0.15em}{\includegraphics[height=1.0em]{images/U+806F.png}}\hskip0pt{}、\raisebox{-0.15em}{\includegraphics[height=1.0em]{images/U+5E73.png}}\hskip0pt{}\raisebox{-0.15em}{\includegraphics[height=1.0em]{images/U+4EC4.png}}\hskip0pt{}\raisebox{-0.15em}{\includegraphics[height=1.0em]{images/U+5E73.png}}\hskip0pt{}\raisebox{-0.15em}{\includegraphics[height=1.0em]{images/U+4EC4.png}}\hskip0pt{}\raisebox{-0.15em}{\includegraphics[height=1.0em]{images/U+4EC4.png}}\hskip0pt{}\raisebox{-0.15em}{\includegraphics[height=1.0em]{images/U+4EC4.png}}\hskip0pt{}\raisebox{-0.15em}{\includegraphics[height=1.0em]{images/U+5E73.png}}\hskip0pt{} (D) \raisebox{-0.15em}{\includegraphics[height=1.0em]{images/U+4E0B.png}}\hskip0pt{}\raisebox{-0.15em}{\includegraphics[height=1.0em]{images/U+806F.png}}\hskip0pt{}、\raisebox{-0.15em}{\includegraphics[height=1.0em]{images/U+5E73.png}}\hskip0pt{}\raisebox{-0.15em}{\includegraphics[height=1.0em]{images/U+4EC4.png}}\hskip0pt{}\raisebox{-0.15em}{\includegraphics[height=1.0em]{images/U+5E73.png}}\hskip0pt{}\raisebox{-0.15em}{\includegraphics[height=1.0em]{images/U+4EC4.png}}\hskip0pt{}\raisebox{-0.15em}{\includegraphics[height=1.0em]{images/U+4EC4.png}}\hskip0pt{}\raisebox{-0.15em}{\includegraphics[height=1.0em]{images/U+4EC4.png}}\hskip0pt{}\raisebox{-0.15em}{\includegraphics[height=1.0em]{images/U+5E73.png}}\hskip0pt{} (E) Cannot be determined.

\end{enumerate}

\subsection{Orthographic Knowledge Dataset}
\label{sec:appendix5_2}
The Orthographic Knowledge Dataset consists of three sub-tasks: Visual Similarity Judgment (25 questions), Cantonese Character Selection (26 questions), and a group of Orthographic Reasoning Tasks (Character Calculation, Radical Description, Character Structure, 54 questions in total).

\begin{enumerate}
    \item \textbf{Visual Similarity Judgement}: The task is to determine which character from a list is the most visually similar to the given character. For example, given the character \raisebox{-0.15em}{\includegraphics[height=1.0em]{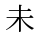}}\hskip0pt{}, and the list ``A \raisebox{-0.15em}{\includegraphics[height=1.0em]{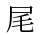}}\hskip0pt{} B \raisebox{-0.15em}{\includegraphics[height=1.0em]{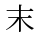}}\hskip0pt{} C \raisebox{-0.15em}{\includegraphics[height=1.0em]{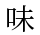}}\hskip0pt{} D \raisebox{-0.15em}{\includegraphics[height=1.0em]{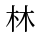}}\hskip0pt{}'', the answer will be B, as the character is only different from the given character in the length of the horizontal strokes.

    \item \textbf{Cantonese Characters Selection}: Cantonese words do not necessarily have a standardised form. Some very common characters can be written in multiple ways. The task provides a sentence frame with a missing character, which should be clear from the context what the syllable is. This is followed by a list of possible characters for insertion. For example, given the sentence ``\raisebox{-0.15em}{\includegraphics[height=1.0em]{images/U+6211.png}}\hskip0pt{}＿\raisebox{-0.15em}{\includegraphics[height=1.0em]{images/U+5F97.png}}\hskip0pt{}\raisebox{-0.15em}{\includegraphics[height=1.0em]{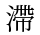}}\hskip0pt{}\raisebox{-0.15em}{\includegraphics[height=1.0em]{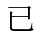}}\hskip0pt{}\raisebox{-0.15em}{\includegraphics[height=1.0em]{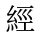}}\hskip0pt{}\raisebox{-0.15em}{\includegraphics[height=1.0em]{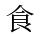}}\hskip0pt{}\raisebox{-0.15em}{\includegraphics[height=1.0em]{images/U+5514.png}}\hskip0pt{}\raisebox{-0.15em}{\includegraphics[height=1.0em]{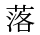}}\hskip0pt{}'' (I am too \textit{something} and I cannot finish eating this.), it should be filled with a word that means “full” in Cantonese. Given the list ``A \raisebox{-0.15em}{\includegraphics[height=1.0em]{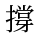}}\hskip0pt{} B \raisebox{-0.15em}{\includegraphics[height=1.0em]{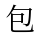}}\hskip0pt{} C \raisebox{-0.15em}{\includegraphics[height=1.0em]{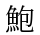}}\hskip0pt{} D \raisebox{-0.15em}{\includegraphics[height=1.0em]{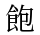}}\hskip0pt{}'', D should be chosen as it is the most accepted word for the Cantonese word \textit{baau2}. Answer A is marginally acceptable in Written Chinese, but it commonly means ``to support'' (caang3) in Cantonese, thus should not be selected.
    
    \item[3a.] \textbf{Character Calculation}: The task is to determine the end result of operations that involve adding, removing or changing glyph components of a character, and choose the closest character from a list. For example, given the instruction to remove  ``\raisebox{-0.15em}{\includegraphics[height=1.0em]{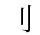}}\hskip0pt{}'' from ``\raisebox{-0.15em}{\includegraphics[height=1.0em]{images/U+5225.png}}\hskip0pt{}'', and a list of ``A \raisebox{-0.15em}{\includegraphics[height=1.0em]{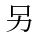}}\hskip0pt{} B \raisebox{-0.15em}{\includegraphics[height=1.0em]{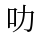}}\hskip0pt{} C \raisebox{-0.15em}{\includegraphics[height=1.0em]{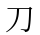}}\hskip0pt{} D \raisebox{-0.15em}{\includegraphics[height=1.0em]{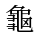}}\hskip0pt{}'', the answer will be A as this is the closest character that matches with the instruction.
    
    \item[3b.] \textbf{Radical Description}: A character will be given, followed by a list of Kangxi radicals. For example, the character ``\raisebox{-0.15em}{\includegraphics[height=1.0em]{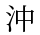}}\hskip0pt{}'' is listed under the ``\raisebox{-0.15em}{\includegraphics[height=1.0em]{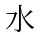}}\hskip0pt{}'' (water) radical, and the Cantonese name for the radical is ``\raisebox{-0.15em}{\includegraphics[height=1.0em]{images/U+4E09.png}}\hskip0pt{}\raisebox{-0.15em}{\includegraphics[height=1.0em]{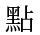}}\hskip0pt{}\raisebox{-0.15em}{\includegraphics[height=1.0em]{images/U+6C34.png}}\hskip0pt{}'' (three drops of water).
    
    \item[3c.] \textbf{Character Structure}: The task is to choose the best description of the structure of the given character. For example, the character ``\raisebox{-0.15em}{\includegraphics[height=1.0em]{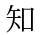}}\hskip0pt{}'' is formed by combining two characters in the left-right frame (``\raisebox{-0.15em}{\includegraphics[height=1.0em]{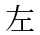}}\hskip0pt{}\raisebox{-0.15em}{\includegraphics[height=1.0em]{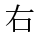}}\hskip0pt{}\raisebox{-0.15em}{\includegraphics[height=1.0em]{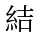}}\hskip0pt{}\raisebox{-0.15em}{\includegraphics[height=1.0em]{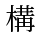}}\hskip0pt{}'').
\end{enumerate}

The questions were evaluated using the following system prompt:

\begin{quote}
    You are a speaker of Cantonese from Hong Kong. Please answer these questions about the properties of the language. Do not include any further explanation.
\end{quote}

Example questions for each task:
\begin{enumerate}
    \item Select the character that is visually similar to \`{}\raisebox{-0.15em}{\includegraphics[height=1.0em]{images/U+672A.png}}\hskip0pt{}\`{} in the Hong Kong context 
(A) \raisebox{-0.15em}{\includegraphics[height=1.0em]{images/U+5C3E.png}}\hskip0pt{} \textbf{(B) \raisebox{-0.15em}{\includegraphics[height=1.0em]{images/U+672B.png}}\hskip0pt{}} (C) \raisebox{-0.15em}{\includegraphics[height=1.0em]{images/U+5473.png}}\hskip0pt{} (D) \raisebox{-0.15em}{\includegraphics[height=1.0em]{images/U+6797.png}}\hskip0pt{} (E) None of the above
    \item Consider this Cantonese sentence \`{}\raisebox{-0.15em}{\includegraphics[height=1.0em]{images/U+6211.png}}\hskip0pt{}＿\raisebox{-0.15em}{\includegraphics[height=1.0em]{images/U+5F97.png}}\hskip0pt{}\raisebox{-0.15em}{\includegraphics[height=1.0em]{images/U+6EEF.png}}\hskip0pt{}\raisebox{-0.15em}{\includegraphics[height=1.0em]{images/U+5DF2.png}}\hskip0pt{}\raisebox{-0.15em}{\includegraphics[height=1.0em]{images/U+7D93.png}}\hskip0pt{}\raisebox{-0.15em}{\includegraphics[height=1.0em]{images/U+98DF.png}}\hskip0pt{}\raisebox{-0.15em}{\includegraphics[height=1.0em]{images/U+5514.png}}\hskip0pt{}\raisebox{-0.15em}{\includegraphics[height=1.0em]{images/U+843D.png}}\hskip0pt{}\`{}. Choose one character below that is the most widely-accepted way to represent the missing word. (A) \raisebox{-0.15em}{\includegraphics[height=1.0em]{images/U+6490.png}}\hskip0pt{} (B) \raisebox{-0.15em}{\includegraphics[height=1.0em]{images/U+5305.png}}\hskip0pt{} (C) \raisebox{-0.15em}{\includegraphics[height=1.0em]{images/U+9B91.png}}\hskip0pt{} \textbf{(D) \raisebox{-0.15em}{\includegraphics[height=1.0em]{images/U+98FD.png}}\hskip0pt{}} (E) None of the above
    \item[3a.] What character do you get by removing \`{}\raisebox{-0.15em}{\includegraphics[height=1.0em]{images/U+5202.png}}\hskip0pt{}\`{} from \`{}\raisebox{-0.15em}{\includegraphics[height=1.0em]{images/U+5225.png}}\hskip0pt{}\`{}? \textbf{(A) \raisebox{-0.15em}{\includegraphics[height=1.0em]{images/U+53E6.png}}\hskip0pt{}} (B) \raisebox{-0.15em}{\includegraphics[height=1.0em]{images/U+53FB.png}}\hskip0pt{} (C) \raisebox{-0.15em}{\includegraphics[height=1.0em]{images/U+5200.png}}\hskip0pt{} (D) \raisebox{-0.15em}{\includegraphics[height=1.0em]{images/U+9F9C.png}}\hskip0pt{} (E) None of the above
    \item[3b.] What is the radical of the character \`{}\raisebox{-0.15em}{\includegraphics[height=1.0em]{images/U+6C96.png}}\hskip0pt{}\`{} and how is it called in Cantonese? (A) \raisebox{-0.15em}{\includegraphics[height=1.0em]{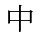}}\hskip0pt{}（\raisebox{-0.15em}{\includegraphics[height=1.0em]{images/U+6C34.png}}\hskip0pt{}\raisebox{-0.15em}{\includegraphics[height=1.0em]{images/U+4E2D.png}}\hskip0pt{}） (B) \raisebox{-0.15em}{\includegraphics[height=1.0em]{images/U+884C.png}}\hskip0pt{}（\raisebox{-0.15em}{\includegraphics[height=1.0em]{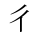}}\hskip0pt{}\raisebox{-0.15em}{\includegraphics[height=1.0em]{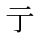}}\hskip0pt{}\raisebox{-0.15em}{\includegraphics[height=1.0em]{images/U+884C.png}}\hskip0pt{}） (C) \raisebox{-0.15em}{\includegraphics[height=1.0em]{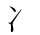}}\hskip0pt{}（\raisebox{-0.15em}{\includegraphics[height=1.0em]{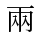}}\hskip0pt{}\raisebox{-0.15em}{\includegraphics[height=1.0em]{images/U+9EDE.png}}\hskip0pt{}\raisebox{-0.15em}{\includegraphics[height=1.0em]{images/U+6C34.png}}\hskip0pt{}） \textbf{(D) \raisebox{-0.15em}{\includegraphics[height=1.0em]{images/U+6C34.png}}\hskip0pt{}（\raisebox{-0.15em}{\includegraphics[height=1.0em]{images/U+4E09.png}}\hskip0pt{}\raisebox{-0.15em}{\includegraphics[height=1.0em]{images/U+9EDE.png}}\hskip0pt{}\raisebox{-0.15em}{\includegraphics[height=1.0em]{images/U+6C34.png}}\hskip0pt{}）} (E) None of the above
    \item[3c.] What is the best description of the character structure of \`{}\raisebox{-0.15em}{\includegraphics[height=1.0em]{images/U+77E5.png}}\hskip0pt{}\`{}? (A) \raisebox{-0.15em}{\includegraphics[height=1.0em]{images/U+4E0A.png}}\hskip0pt{}\raisebox{-0.15em}{\includegraphics[height=1.0em]{images/U+4E0B.png}}\hskip0pt{} (B) \raisebox{-0.15em}{\includegraphics[height=1.0em]{images/U+5305.png}}\hskip0pt{}\raisebox{-0.15em}{\includegraphics[height=1.0em]{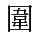}}\hskip0pt{} (C) \raisebox{-0.15em}{\includegraphics[height=1.0em]{images/U+5DE6.png}}\hskip0pt{}\raisebox{-0.15em}{\includegraphics[height=1.0em]{images/U+4E2D.png}}\hskip0pt{}\raisebox{-0.15em}{\includegraphics[height=1.0em]{images/U+53F3.png}}\hskip0pt{} (D) \raisebox{-0.15em}{\includegraphics[height=1.0em]{images/U+524D.png}}\hskip0pt{}\raisebox{-0.15em}{\includegraphics[height=1.0em]{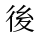}}\hskip0pt{} \textbf{(E) None of the above}

\end{enumerate}

\subsection{Grapheme-to-Phoneme (G2P) Conversion Dataset}
\label{sec:appendix5_3}
Cantonese transliteration is not trivial because the characters and pronunciation form a many-to-many relation. The same syllable can be represented by different characters, which is a crucial feature of an ideographic writing system, whereas the same character may have multiple pronunciations due to the overloading of certain characters or multiple layers of pronunciation norms. These judgements are often not well-documented. Characters with multiple pronunciations are often semantically or lexically determined, for example: ``\raisebox{-0.15em}{\includegraphics[height=1.0em]{images/U+884C.png}}\hskip0pt{}'' can be pronounced as \textit{hang4} (e.g. ``\raisebox{-0.15em}{\includegraphics[height=1.0em]{images/U+884C.png}}\hskip0pt{}\raisebox{-0.15em}{\includegraphics[height=1.0em]{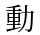}}\hskip0pt{}'' action, ``\raisebox{-0.15em}{\includegraphics[height=1.0em]{images/U+6D41.png}}\hskip0pt{}\raisebox{-0.15em}{\includegraphics[height=1.0em]{images/U+884C.png}}\hskip0pt{}'' trend), \textit{haang4} (e.g. ``\raisebox{-0.15em}{\includegraphics[height=1.0em]{images/U+884C.png}}\hskip0pt{}\raisebox{-0.15em}{\includegraphics[height=1.0em]{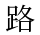}}\hskip0pt{}'' to walk, ``\raisebox{-0.15em}{\includegraphics[height=1.0em]{images/U+884C.png}}\hskip0pt{}\raisebox{-0.15em}{\includegraphics[height=1.0em]{images/U+8857.png}}\hskip0pt{}'' to go shopping), \textit{hong4} (e.g. ``\raisebox{-0.15em}{\includegraphics[height=1.0em]{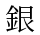}}\hskip0pt{}\raisebox{-0.15em}{\includegraphics[height=1.0em]{images/U+884C.png}}\hskip0pt{}'' bank, ``\raisebox{-0.15em}{\includegraphics[height=1.0em]{images/U+884C.png}}\hskip0pt{}\raisebox{-0.15em}{\includegraphics[height=1.0em]{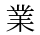}}\hskip0pt{}'' occupation), \textit{hong2} (e.g. ``\raisebox{-0.15em}{\includegraphics[height=1.0em]{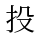}}\hskip0pt{}\raisebox{-0.15em}{\includegraphics[height=1.0em]{images/U+884C.png}}\hskip0pt{}'' investment bank). For a smaller subset of characters, there can be different pronunciations due to a literary-colloquial distinction, e.g. ``\raisebox{-0.15em}{\includegraphics[height=1.0em]{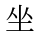}}\hskip0pt{}'' (to sit) can be \textit{zo6} or \textit{co5}; ``\raisebox{-0.15em}{\includegraphics[height=1.0em]{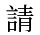}}\hskip0pt{}'' (to invite) can be \textit{cing2} or \textit{ceng2}.

The five-shot evaluation prompt used for the G2P dataset evaluation:
\begin{quote}
You are an expert in Cantonese linguistics. Please convert the given Cantonese sentence into Jyutping romanisation. You can ignore all punctuation marks, and normalise all numerals and English loanwords into Cantonese pronunciation. Do not include any further explanation.

\textbf{Example 1}\newline
\raisebox{-0.15em}{\includegraphics[height=1.0em]{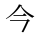}}\hskip0pt{}\raisebox{-0.15em}{\includegraphics[height=1.0em]{images/U+65E5.png}}\hskip0pt{}\raisebox{-0.15em}{\includegraphics[height=1.0em]{images/U+5929.png}}\hskip0pt{}\raisebox{-0.15em}{\includegraphics[height=1.0em]{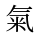}}\hskip0pt{}\raisebox{-0.15em}{\includegraphics[height=1.0em]{images/U+597D.png}}\hskip0pt{}\raisebox{-0.15em}{\includegraphics[height=1.0em]{images/U+597D.png}}\hskip0pt{}，\raisebox{-0.15em}{\includegraphics[height=1.0em]{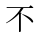}}\hskip0pt{}\raisebox{-0.15em}{\includegraphics[height=1.0em]{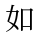}}\hskip0pt{}\raisebox{-0.15em}{\includegraphics[height=1.0em]{images/U+51FA.png}}\hskip0pt{}\raisebox{-0.15em}{\includegraphics[height=1.0em]{images/U+53BB.png}}\hskip0pt{}\raisebox{-0.15em}{\includegraphics[height=1.0em]{images/U+6563.png}}\hskip0pt{}\raisebox{-0.15em}{\includegraphics[height=1.0em]{images/U+6B65.png}}\hskip0pt{}？\newline
gam1 jat6 tin1 hei3 hou2 hou2 bat1 jyu4 ceot1 heoi3 saan3 bou6

\textbf{Example 2}\newline
\raisebox{-0.15em}{\includegraphics[height=1.0em]{images/U+6211.png}}\hskip0pt{}\raisebox{-0.15em}{\includegraphics[height=1.0em]{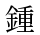}}\hskip0pt{}\raisebox{-0.15em}{\includegraphics[height=1.0em]{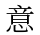}}\hskip0pt{}\raisebox{-0.15em}{\includegraphics[height=1.0em]{images/U+884C.png}}\hskip0pt{}\raisebox{-0.15em}{\includegraphics[height=1.0em]{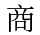}}\hskip0pt{}\raisebox{-0.15em}{\includegraphics[height=1.0em]{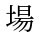}}\hskip0pt{}\raisebox{-0.15em}{\includegraphics[height=1.0em]{images/U+5514.png}}\hskip0pt{}\raisebox{-0.15em}{\includegraphics[height=1.0em]{images/U+937E.png}}\hskip0pt{}\raisebox{-0.15em}{\includegraphics[height=1.0em]{images/U+610F.png}}\hskip0pt{}\raisebox{-0.15em}{\includegraphics[height=1.0em]{images/U+884C.png}}\hskip0pt{}\raisebox{-0.15em}{\includegraphics[height=1.0em]{images/U+8857.png}}\hskip0pt{}\raisebox{-0.15em}{\includegraphics[height=1.0em]{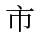}}\hskip0pt{}\newline
ngo5 zung1 ji3 haang4 soeng1 coeng4 m4 zung1 ji3 haang4 gaai1 si5

\textbf{Example 3}\newline
\raisebox{-0.15em}{\includegraphics[height=1.0em]{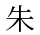}}\hskip0pt{}\raisebox{-0.15em}{\includegraphics[height=1.0em]{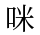}}\hskip0pt{}\raisebox{-0.15em}{\includegraphics[height=1.0em]{images/U+54AA.png}}\hskip0pt{}\raisebox{-0.15em}{\includegraphics[height=1.0em]{images/U+4FC2.png}}\hskip0pt{}\raisebox{-0.15em}{\includegraphics[height=1.0em]{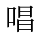}}\hskip0pt{}\raisebox{-0.15em}{\includegraphics[height=1.0em]{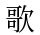}}\hskip0pt{}\raisebox{-0.15em}{\includegraphics[height=1.0em]{images/U+597D.png}}\hskip0pt{}\raisebox{-0.15em}{\includegraphics[height=1.0em]{images/U+53FB.png}}\hskip0pt{}\raisebox{-0.15em}{\includegraphics[height=1.0em]{images/U+5605.png}}\hskip0pt{}\raisebox{-0.15em}{\includegraphics[height=1.0em]{images/U+6B4C.png}}\hskip0pt{}\raisebox{-0.15em}{\includegraphics[height=1.0em]{images/U+624B.png}}\hskip0pt{}\newline
zyu1 mi1 mi4 hai6 coeng3 go1 hou2 lek1 ge3 go1 sau2

\textbf{Example 4}\newline
\raisebox{-0.15em}{\includegraphics[height=1.0em]{images/U+4ECA.png}}\hskip0pt{}\raisebox{-0.15em}{\includegraphics[height=1.0em]{images/U+5929.png}}\hskip0pt{}\raisebox{-0.15em}{\includegraphics[height=1.0em]{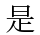}}\hskip0pt{}\raisebox{-0.15em}{\includegraphics[height=1.0em]{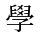}}\hskip0pt{}\raisebox{-0.15em}{\includegraphics[height=1.0em]{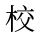}}\hskip0pt{}\raisebox{-0.15em}{\includegraphics[height=1.0em]{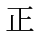}}\hskip0pt{}\raisebox{-0.15em}{\includegraphics[height=1.0em]{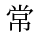}}\hskip0pt{}\raisebox{-0.15em}{\includegraphics[height=1.0em]{images/U+4E0A.png}}\hskip0pt{}\raisebox{-0.15em}{\includegraphics[height=1.0em]{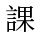}}\hskip0pt{}\raisebox{-0.15em}{\includegraphics[height=1.0em]{images/U+65E5.png}}\hskip0pt{}，\raisebox{-0.15em}{\includegraphics[height=1.0em]{images/U+8ACB.png}}\hskip0pt{}\raisebox{-0.15em}{\includegraphics[height=1.0em]{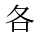}}\hskip0pt{}\raisebox{-0.15em}{\includegraphics[height=1.0em]{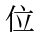}}\hskip0pt{}\raisebox{-0.15em}{\includegraphics[height=1.0em]{images/U+5BB6.png}}\hskip0pt{}\raisebox{-0.15em}{\includegraphics[height=1.0em]{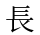}}\hskip0pt{}\raisebox{-0.15em}{\includegraphics[height=1.0em]{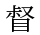}}\hskip0pt{}\raisebox{-0.15em}{\includegraphics[height=1.0em]{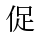}}\hskip0pt{}\raisebox{-0.15em}{\includegraphics[height=1.0em]{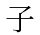}}\hskip0pt{}\raisebox{-0.15em}{\includegraphics[height=1.0em]{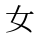}}\hskip0pt{}\raisebox{-0.15em}{\includegraphics[height=1.0em]{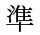}}\hskip0pt{}\raisebox{-0.15em}{\includegraphics[height=1.0em]{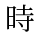}}\hskip0pt{}\raisebox{-0.15em}{\includegraphics[height=1.0em]{images/U+4E0A.png}}\hskip0pt{}\raisebox{-0.15em}{\includegraphics[height=1.0em]{images/U+5B78.png}}\hskip0pt{}。\newline
gam1 tin1 si6 hok6 haau6 zing3 soeng4 soeng5 fo3 jat6 ceng2 gok3 wai2 gaa1 zoeng2 duk1 cuk1 zi2 neoi5 zeon2 si4 soeng5 hok6

\textbf{Example 5}\newline
\raisebox{-0.15em}{\includegraphics[height=1.0em]{images/U+5B78.png}}\hskip0pt{}\raisebox{-0.15em}{\includegraphics[height=1.0em]{images/U+800C.png}}\hskip0pt{}\raisebox{-0.15em}{\includegraphics[height=1.0em]{images/U+6642.png}}\hskip0pt{}\raisebox{-0.15em}{\includegraphics[height=1.0em]{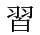}}\hskip0pt{}\raisebox{-0.15em}{\includegraphics[height=1.0em]{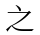}}\hskip0pt{}，\raisebox{-0.15em}{\includegraphics[height=1.0em]{images/U+4E0D.png}}\hskip0pt{}\raisebox{-0.15em}{\includegraphics[height=1.0em]{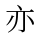}}\hskip0pt{}\raisebox{-0.15em}{\includegraphics[height=1.0em]{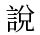}}\hskip0pt{}\raisebox{-0.15em}{\includegraphics[height=1.0em]{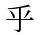}}\hskip0pt{}\newline
hok6 ji4 si4 zaap6 zi1 bat1 jik6 jyut6 fu4

\end{quote}

\textbf{Score calculation}: All acceptable variants have been listed as answers and all variants are considered equally good. This is to handle variant forms and ambiguous interpretations that may not be the standard, but native speakers of Cantonese accept. We will use the answer with the highest score for the subsequent calculation. Two metrics were used: character error rate (CER) and Levenshtein distance. A lower score means a better performance. The answers with the lowest Levenshtein distance (i.e. the best score) were used for the calculation.

\begin{table*}[]
\centering
\begin{tabular}{l|c|c|c|c}
Model             & \begin{tabular}[c]{@{}c@{}}Translation\\ (0-shot)\end{tabular} & \begin{tabular}[c]{@{}c@{}}Translation\\ (3-shot)\end{tabular} & Summarisation & Sentiment \\ \hline
Claude 3.5 Sonnet& 96.2\%& 91.9\%& \textbf{92.7\%}& 76.0\%\\
Doubao Pro& \textbf{97.5\%}& 97.4\%& 85.5\%& 67.7\%\\
Ernie 4.0& 86.2\%& 87.1\%& 83.3\%& 74.1\%\\
Gemini 1.5 Flash& 92.6\%& 95.5\%& 70.0\%& 74.9\%\\
Gemini 1.5 Pro& 95.3\%& 91.1\%& 90.0\%& 75.3\%\\
GPT4o& 96.7\%& \textbf{98.3\%}& 83.7\%& \textbf{79.7\%}\\
GPT4o-mini& 89.8\%& 95.2\%& 84.8\%& 74.7\%\\
SenseChat& 90.6\%& 93.8\%& 54.3\%& 76.5\%\\ \hline
Aya 23 8B         & 81.0\%                                                         & 79.4\%                                                         & 52.8\%        & 67.1\%    \\
CLLM v0.5 6B      & \textbf{97.5\%}                                                         & 94.7\%                                                         & 29.0\%        & 66.5\%    \\
CLLM v0.5 34B     & 86.6\%                                                         & 90.3\%                                                         & 43.2\%        & 73.1\%    \\
Yi 1.5 6B         & 62.9\%                                                         & 56.7\%                                                         & 42.5\%        & 64.3\%    \\
Yi 1.5 9B         & 69.5\%                                                         & 88.1\%                                                         & 62.2\%        & 69.1\%    \\
Yi 1.5 34B        & 87.3\%                                                         & 95.8\%                                                         & 70.3\%        & 78.2\%    \\
Gemma 2 2B        & 71.4\%                                                         & 64.4\%                                                         & 86.3\%        & 71.6\%    \\
Gemma 2 9B        & 87.8\%                                                         & 88.2\%                                                         & \textbf{91.3\%}        & 72.6\%    \\
Gemma 2 27B       & 89.3\%                                                         & 78.4\%                                                         & 89.0\%        & 76.1\%    \\
Llama 3.1 8B      & 76.3\%                                                         & 81.1\%                                                         & 15.3\%        & 68.6\%    \\
Llama 3.1 70B     & 92.1\%                                                         & 86.6\%                                                         & 81.7\%        & 77.5\%    \\
Llama 3.1 405B    & 77.5\%                                                         & 93.3\%                                                         & 8.0\%         & \textbf{78.8\%}    \\
Mistral Nemo 12B  & 77.0\%                                                         & 84.8\%                                                         & 40.7\%        & 72.7\%    \\
Qwen2 7B          & 84.5\%                                                         & 90.1\%                                                         & 14.8\%        & 77.9\%    \\
Qwen2 72B         & 97.1\%                                                         & \textbf{96.6\%}                                                         & 62.2\%        & 78.2\%    \\ \hline
Control           & 99.1\%                                                         & 99.1\%                                                         & -             & -        
\end{tabular}
\caption{\label{tab:nlp-perf-table}Model performance in various NLP tasks}
\end{table*}

\section{Translation Task}
\label{sec:appendix6}
The translation dataset consists of 20 Cantonese sentences. These sentences were designed to check the models' understanding of sentence nuances as they involve lexical or contextual ambiguities that require good linguistic reasoning to resolve. Each Cantonese sentence was translated into both English and written Chinese by a professional translator. This provides the source for four translation pairs per original Cantonese sentence: Cantonese-English, Cantonese-Written\_Chinese, English-Cantonese, Written\_Chinese-Cantonese. Here is an example sentence taken from the few shot examples:

Cantonese (avg. 28.6 characters):
\begin{quote}
    \raisebox{-0.15em}{\includegraphics[height=1.0em]{images/U+8FD4.png}}\hskip0pt{}\raisebox{-0.15em}{\includegraphics[height=1.0em]{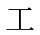}}\hskip0pt{}\raisebox{-0.15em}{\includegraphics[height=1.0em]{images/U+8FD4.png}}\hskip0pt{}\raisebox{-0.15em}{\includegraphics[height=1.0em]{images/U+5230.png}}\hskip0pt{}\raisebox{-0.15em}{\includegraphics[height=1.0em]{images/U+5F97.png}}\hskip0pt{}\raisebox{-0.15em}{\includegraphics[height=1.0em]{images/U+8FD4.png}}\hskip0pt{}\raisebox{-0.15em}{\includegraphics[height=1.0em]{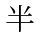}}\hskip0pt{}\raisebox{-0.15em}{\includegraphics[height=1.0em]{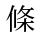}}\hskip0pt{}\raisebox{-0.15em}{\includegraphics[height=1.0em]{images/U+4EBA.png}}\hskip0pt{}\raisebox{-0.15em}{\includegraphics[height=1.0em]{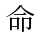}}\hskip0pt{}，\raisebox{-0.15em}{\includegraphics[height=1.0em]{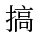}}\hskip0pt{}\raisebox{-0.15em}{\includegraphics[height=1.0em]{images/U+5230.png}}\hskip0pt{}\raisebox{-0.15em}{\includegraphics[height=1.0em]{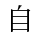}}\hskip0pt{}\raisebox{-0.15em}{\includegraphics[height=1.0em]{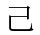}}\hskip0pt{}\raisebox{-0.15em}{\includegraphics[height=1.0em]{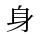}}\hskip0pt{}\raisebox{-0.15em}{\includegraphics[height=1.0em]{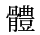}}\hskip0pt{}\raisebox{-0.15em}{\includegraphics[height=1.0em]{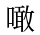}}\hskip0pt{}\raisebox{-0.15em}{\includegraphics[height=1.0em]{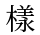}}\hskip0pt{}，\raisebox{-0.15em}{\includegraphics[height=1.0em]{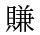}}\hskip0pt{}\raisebox{-0.15em}{\includegraphics[height=1.0em]{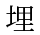}}\hskip0pt{}\raisebox{-0.15em}{\includegraphics[height=1.0em]{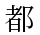}}\hskip0pt{}\raisebox{-0.15em}{\includegraphics[height=1.0em]{images/U+5514.png}}\hskip0pt{}\raisebox{-0.15em}{\includegraphics[height=1.0em]{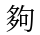}}\hskip0pt{}\raisebox{-0.15em}{\includegraphics[height=1.0em]{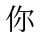}}\hskip0pt{}\raisebox{-0.15em}{\includegraphics[height=1.0em]{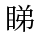}}\hskip0pt{}\raisebox{-0.15em}{\includegraphics[height=1.0em]{images/U+91AB.png}}\hskip0pt{}\raisebox{-0.15em}{\includegraphics[height=1.0em]{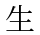}}\hskip0pt{}\raisebox{-0.15em}{\includegraphics[height=1.0em]{images/U+5566.png}}\hskip0pt{}。
\end{quote}

English (avg. 23.1 words):
\begin{quote}
    Your day job is so exhausting that you are like half-dead after work. If you keep pushing yourself like this, the money you earn won't even cover your medical bills.
\end{quote}

Written Chinese (avg. 29.6 characters):
\begin{quote}
    \raisebox{-0.15em}{\includegraphics[height=1.0em]{images/U+6253.png}}\hskip0pt{}\raisebox{-0.15em}{\includegraphics[height=1.0em]{images/U+5DE5.png}}\hskip0pt{}\raisebox{-0.15em}{\includegraphics[height=1.0em]{images/U+6253.png}}\hskip0pt{}\raisebox{-0.15em}{\includegraphics[height=1.0em]{images/U+5230.png}}\hskip0pt{}\raisebox{-0.15em}{\includegraphics[height=1.0em]{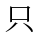}}\hskip0pt{}\raisebox{-0.15em}{\includegraphics[height=1.0em]{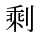}}\hskip0pt{}\raisebox{-0.15em}{\includegraphics[height=1.0em]{images/U+4E0B.png}}\hskip0pt{}\raisebox{-0.15em}{\includegraphics[height=1.0em]{images/U+534A.png}}\hskip0pt{}\raisebox{-0.15em}{\includegraphics[height=1.0em]{images/U+689D.png}}\hskip0pt{}\raisebox{-0.15em}{\includegraphics[height=1.0em]{images/U+547D.png}}\hskip0pt{}，\raisebox{-0.15em}{\includegraphics[height=1.0em]{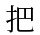}}\hskip0pt{}\raisebox{-0.15em}{\includegraphics[height=1.0em]{images/U+81EA.png}}\hskip0pt{}\raisebox{-0.15em}{\includegraphics[height=1.0em]{images/U+5DF1.png}}\hskip0pt{}\raisebox{-0.15em}{\includegraphics[height=1.0em]{images/U+8EAB.png}}\hskip0pt{}\raisebox{-0.15em}{\includegraphics[height=1.0em]{images/U+9AD4.png}}\hskip0pt{}\raisebox{-0.15em}{\includegraphics[height=1.0em]{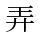}}\hskip0pt{}\raisebox{-0.15em}{\includegraphics[height=1.0em]{images/U+6210.png}}\hskip0pt{}\raisebox{-0.15em}{\includegraphics[height=1.0em]{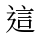}}\hskip0pt{}\raisebox{-0.15em}{\includegraphics[height=1.0em]{images/U+6A23.png}}\hskip0pt{}，\raisebox{-0.15em}{\includegraphics[height=1.0em]{images/U+8CFA.png}}\hskip0pt{}\raisebox{-0.15em}{\includegraphics[height=1.0em]{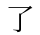}}\hskip0pt{}\raisebox{-0.15em}{\includegraphics[height=1.0em]{images/U+7684.png}}\hskip0pt{}\raisebox{-0.15em}{\includegraphics[height=1.0em]{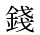}}\hskip0pt{}\raisebox{-0.15em}{\includegraphics[height=1.0em]{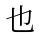}}\hskip0pt{}\raisebox{-0.15em}{\includegraphics[height=1.0em]{images/U+4E0D.png}}\hskip0pt{}\raisebox{-0.15em}{\includegraphics[height=1.0em]{images/U+5920.png}}\hskip0pt{}\raisebox{-0.15em}{\includegraphics[height=1.0em]{images/U+4F60.png}}\hskip0pt{}\raisebox{-0.15em}{\includegraphics[height=1.0em]{images/U+53BB.png}}\hskip0pt{}\raisebox{-0.15em}{\includegraphics[height=1.0em]{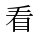}}\hskip0pt{}\raisebox{-0.15em}{\includegraphics[height=1.0em]{images/U+91AB.png}}\hskip0pt{}\raisebox{-0.15em}{\includegraphics[height=1.0em]{images/U+751F.png}}\hskip0pt{}。
\end{quote}

Evaluation prompt for the three-shot translation task (from Written Chinese to Cantonese) (Note: the prompt uses the word ``Traditional Chinese'' to force the model to return the results in the Traditional script.):

\begin{quote}
    \raisebox{-0.15em}{\includegraphics[height=1.0em]{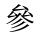}}\hskip0pt{}\raisebox{-0.15em}{\includegraphics[height=1.0em]{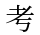}}\hskip0pt{}\raisebox{-0.15em}{\includegraphics[height=1.0em]{images/U+4EE5.png}}\hskip0pt{}\raisebox{-0.15em}{\includegraphics[height=1.0em]{images/U+4E0B.png}}\hskip0pt{}\raisebox{-0.15em}{\includegraphics[height=1.0em]{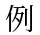}}\hskip0pt{}\raisebox{-0.15em}{\includegraphics[height=1.0em]{images/U+5B50.png}}\hskip0pt{}，\raisebox{-0.15em}{\includegraphics[height=1.0em]{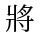}}\hskip0pt{}\raisebox{-0.15em}{\includegraphics[height=1.0em]{images/U+4EE5.png}}\hskip0pt{}\raisebox{-0.15em}{\includegraphics[height=1.0em]{images/U+4E0B.png}}\hskip0pt{}\raisebox{-0.15em}{\includegraphics[height=1.0em]{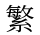}}\hskip0pt{}\raisebox{-0.15em}{\includegraphics[height=1.0em]{images/U+9AD4.png}}\hskip0pt{}\raisebox{-0.15em}{\includegraphics[height=1.0em]{images/U+4E2D.png}}\hskip0pt{}\raisebox{-0.15em}{\includegraphics[height=1.0em]{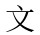}}\hskip0pt{}\raisebox{-0.15em}{\includegraphics[height=1.0em]{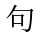}}\hskip0pt{}\raisebox{-0.15em}{\includegraphics[height=1.0em]{images/U+5B50.png}}\hskip0pt{}\raisebox{-0.15em}{\includegraphics[height=1.0em]{images/U+7FFB.png}}\hskip0pt{}\raisebox{-0.15em}{\includegraphics[height=1.0em]{images/U+8B6F.png}}\hskip0pt{}\raisebox{-0.15em}{\includegraphics[height=1.0em]{images/U+6210.png}}\hskip0pt{}\raisebox{-0.15em}{\includegraphics[height=1.0em]{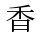}}\hskip0pt{}\raisebox{-0.15em}{\includegraphics[height=1.0em]{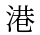}}\hskip0pt{}\raisebox{-0.15em}{\includegraphics[height=1.0em]{images/U+7684.png}}\hskip0pt{}\raisebox{-0.15em}{\includegraphics[height=1.0em]{images/U+5EE3.png}}\hskip0pt{}\raisebox{-0.15em}{\includegraphics[height=1.0em]{images/U+6771.png}}\hskip0pt{}\raisebox{-0.15em}{\includegraphics[height=1.0em]{images/U+8A71.png}}\hskip0pt{}：\\

\raisebox{-0.15em}{\includegraphics[height=1.0em]{images/U+4F8B.png}}\hskip0pt{}\raisebox{-0.15em}{\includegraphics[height=1.0em]{images/U+5B50.png}}\hskip0pt{}1：\raisebox{-0.15em}{\includegraphics[height=1.0em]{images/U+4E2D.png}}\hskip0pt{}\raisebox{-0.15em}{\includegraphics[height=1.0em]{images/U+6587.png}}\hskip0pt{}\raisebox{-0.15em}{\includegraphics[height=1.0em]{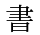}}\hskip0pt{}\raisebox{-0.15em}{\includegraphics[height=1.0em]{images/U+9762.png}}\hskip0pt{}\raisebox{-0.15em}{\includegraphics[height=1.0em]{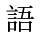}}\hskip0pt{}: (e.g. 1) \raisebox{-0.15em}{\includegraphics[height=1.0em]{images/U+5EE3.png}}\hskip0pt{}\raisebox{-0.15em}{\includegraphics[height=1.0em]{images/U+6771.png}}\hskip0pt{}\raisebox{-0.15em}{\includegraphics[height=1.0em]{images/U+8A71.png}}\hskip0pt{}: (e.g. 1)\\
\raisebox{-0.15em}{\includegraphics[height=1.0em]{images/U+4F8B.png}}\hskip0pt{}\raisebox{-0.15em}{\includegraphics[height=1.0em]{images/U+5B50.png}}\hskip0pt{}2：\raisebox{-0.15em}{\includegraphics[height=1.0em]{images/U+4E2D.png}}\hskip0pt{}\raisebox{-0.15em}{\includegraphics[height=1.0em]{images/U+6587.png}}\hskip0pt{}\raisebox{-0.15em}{\includegraphics[height=1.0em]{images/U+66F8.png}}\hskip0pt{}\raisebox{-0.15em}{\includegraphics[height=1.0em]{images/U+9762.png}}\hskip0pt{}\raisebox{-0.15em}{\includegraphics[height=1.0em]{images/U+8A9E.png}}\hskip0pt{}: (e.g. 2) \raisebox{-0.15em}{\includegraphics[height=1.0em]{images/U+5EE3.png}}\hskip0pt{}\raisebox{-0.15em}{\includegraphics[height=1.0em]{images/U+6771.png}}\hskip0pt{}\raisebox{-0.15em}{\includegraphics[height=1.0em]{images/U+8A71.png}}\hskip0pt{}: (e.g. 2)\\
\raisebox{-0.15em}{\includegraphics[height=1.0em]{images/U+4F8B.png}}\hskip0pt{}\raisebox{-0.15em}{\includegraphics[height=1.0em]{images/U+5B50.png}}\hskip0pt{}3：\raisebox{-0.15em}{\includegraphics[height=1.0em]{images/U+4E2D.png}}\hskip0pt{}\raisebox{-0.15em}{\includegraphics[height=1.0em]{images/U+6587.png}}\hskip0pt{}\raisebox{-0.15em}{\includegraphics[height=1.0em]{images/U+66F8.png}}\hskip0pt{}\raisebox{-0.15em}{\includegraphics[height=1.0em]{images/U+9762.png}}\hskip0pt{}\raisebox{-0.15em}{\includegraphics[height=1.0em]{images/U+8A9E.png}}\hskip0pt{}: (e.g. 3) \raisebox{-0.15em}{\includegraphics[height=1.0em]{images/U+5EE3.png}}\hskip0pt{}\raisebox{-0.15em}{\includegraphics[height=1.0em]{images/U+6771.png}}\hskip0pt{}\raisebox{-0.15em}{\includegraphics[height=1.0em]{images/U+8A71.png}}\hskip0pt{}: (e.g. 3)\\

\raisebox{-0.15em}{\includegraphics[height=1.0em]{images/U+8ACB.png}}\hskip0pt{}\raisebox{-0.15em}{\includegraphics[height=1.0em]{images/U+7FFB.png}}\hskip0pt{}\raisebox{-0.15em}{\includegraphics[height=1.0em]{images/U+8B6F.png}}\hskip0pt{}\raisebox{-0.15em}{\includegraphics[height=1.0em]{images/U+4EE5.png}}\hskip0pt{}\raisebox{-0.15em}{\includegraphics[height=1.0em]{images/U+4E0B.png}}\hskip0pt{}\raisebox{-0.15em}{\includegraphics[height=1.0em]{images/U+6587.png}}\hskip0pt{}\raisebox{-0.15em}{\includegraphics[height=1.0em]{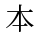}}\hskip0pt{}：\\
(Text to translate)\\

\raisebox{-0.15em}{\includegraphics[height=1.0em]{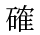}}\hskip0pt{}\raisebox{-0.15em}{\includegraphics[height=1.0em]{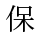}}\hskip0pt{}\raisebox{-0.15em}{\includegraphics[height=1.0em]{images/U+7FFB.png}}\hskip0pt{}\raisebox{-0.15em}{\includegraphics[height=1.0em]{images/U+8B6F.png}}\hskip0pt{}\raisebox{-0.15em}{\includegraphics[height=1.0em]{images/U+6E96.png}}\hskip0pt{}\raisebox{-0.15em}{\includegraphics[height=1.0em]{images/U+78BA.png}}\hskip0pt{}\raisebox{-0.15em}{\includegraphics[height=1.0em]{images/U+81EA.png}}\hskip0pt{}\raisebox{-0.15em}{\includegraphics[height=1.0em]{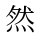}}\hskip0pt{}，\raisebox{-0.15em}{\includegraphics[height=1.0em]{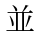}}\hskip0pt{}\raisebox{-0.15em}{\includegraphics[height=1.0em]{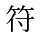}}\hskip0pt{}\raisebox{-0.15em}{\includegraphics[height=1.0em]{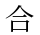}}\hskip0pt{}\raisebox{-0.15em}{\includegraphics[height=1.0em]{images/U+9999.png}}\hskip0pt{}\raisebox{-0.15em}{\includegraphics[height=1.0em]{images/U+6E2F.png}}\hskip0pt{}\raisebox{-0.15em}{\includegraphics[height=1.0em]{images/U+7684.png}}\hskip0pt{}\raisebox{-0.15em}{\includegraphics[height=1.0em]{images/U+5EE3.png}}\hskip0pt{}\raisebox{-0.15em}{\includegraphics[height=1.0em]{images/U+6771.png}}\hskip0pt{}\raisebox{-0.15em}{\includegraphics[height=1.0em]{images/U+8A71.png}}\hskip0pt{}\raisebox{-0.15em}{\includegraphics[height=1.0em]{images/U+8A9E.png}}\hskip0pt{}\raisebox{-0.15em}{\includegraphics[height=1.0em]{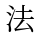}}\hskip0pt{}\raisebox{-0.15em}{\includegraphics[height=1.0em]{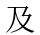}}\hskip0pt{}\raisebox{-0.15em}{\includegraphics[height=1.0em]{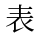}}\hskip0pt{}\raisebox{-0.15em}{\includegraphics[height=1.0em]{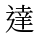}}\hskip0pt{}\raisebox{-0.15em}{\includegraphics[height=1.0em]{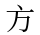}}\hskip0pt{}\raisebox{-0.15em}{\includegraphics[height=1.0em]{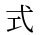}}\hskip0pt{}\\

\raisebox{-0.15em}{\includegraphics[height=1.0em]{images/U+53EA.png}}\hskip0pt{}\raisebox{-0.15em}{\includegraphics[height=1.0em]{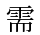}}\hskip0pt{}\raisebox{-0.15em}{\includegraphics[height=1.0em]{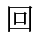}}\hskip0pt{}\raisebox{-0.15em}{\includegraphics[height=1.0em]{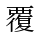}}\hskip0pt{}\raisebox{-0.15em}{\includegraphics[height=1.0em]{images/U+7FFB.png}}\hskip0pt{}\raisebox{-0.15em}{\includegraphics[height=1.0em]{images/U+8B6F.png}}\hskip0pt{}\raisebox{-0.15em}{\includegraphics[height=1.0em]{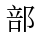}}\hskip0pt{}\raisebox{-0.15em}{\includegraphics[height=1.0em]{images/U+4EFD.png}}\hskip0pt{}，\raisebox{-0.15em}{\includegraphics[height=1.0em]{images/U+4E0D.png}}\hskip0pt{}\raisebox{-0.15em}{\includegraphics[height=1.0em]{images/U+9700.png}}\hskip0pt{}\raisebox{-0.15em}{\includegraphics[height=1.0em]{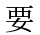}}\hskip0pt{}\raisebox{-0.15em}{\includegraphics[height=1.0em]{images/U+89E3.png}}\hskip0pt{}\raisebox{-0.15em}{\includegraphics[height=1.0em]{images/U+91CB.png}}\hskip0pt{}\\
\end{quote}

English translation of the prompt:

\begin{quote}
    Translate the following Traditional Chinese sentence to the Cantonese used in Hong Kong, referring to the examples below:\\

Example 1: Written Chinese: (e.g. 1) Cantonese: (e.g. 1)\\
Example 2: Written Chinese: (e.g. 2) Cantonese: (e.g. 2)\\
Example 3: Written Chinese: (e.g. 3) Cantonese: (e.g. 3)\\

Text to Translate:\\
(Text to translate)\\

Ensure the translation is accurate and natural, preserving the original meaning and using concise and fluent English expression.\\
Only return the translation. Do not explain.
\end{quote}

The other 2 translation pairs used a similar format, but an English prompt is used when translating from Cantonese to English:

\begin{quote}
    Translate the following Cantonese text into English, referring to the examples below:\\

Example 1: Cantonese: (e.g. 1) English: (e.g. 1)\\
Example 2: Cantonese: (e.g. 2) English: (e.g. 2)\\
Example 3: Cantonese: (e.g. 3) English: (e.g. 3)\\

Text to Translate:\\
(Text to translate)\\

Ensure the translation is accurate and natural, preserving the original meaning and using concise and fluent English expression.\\
ONLY RETURN THE TRANSLATION. DO NOT EXPLAIN.
\end{quote}

For zero-shot evaluation, similar prompts were used but without the few-shot examples and references to them.

\subsection{Translation Results Evaluation}
\label{sec:appendix6_1}

We used manual evaluation for the translation task. The BLEU score was not used in this work to evaluate the translation task because it relied on exact matches between the answer and the gold standard, which is often not ideal for pairs that involve Written Chinese. Translation between closely related varieties in a diglossic situation is similar to stylistic change, with a wide range of acceptable answers. One kind of translation strategy would be favoured by using a BLEU score, and models that could have been rated excellent by Hong Kong users would be penalised. Adding to this is the lack of existing libraries that handle orthographic variants well enough to conduct a fair string comparison with the gold standard. This is why manual annotation was opted for by us. This is to address the problems of an earlier benchmark \citep{jiang2024how}, outlined in Appendix \ref{sec:appendix1}.

A graphical user interface based on Label Studio was configured such that annotators (paid undergraduate students and teaching assistants from Hong Kong) can highlight mistakes in the translated text for the following categories:

\begin{enumerate}
    \item Additional words
    \item Collocation
    \item Inconsistent Terminology
    \item Literal Translation 
    \item Mistranslation
    \item Mixed up Cantonese-Mandarin 
    \item Orthography
    \item Omission (Labelled in the source text)
    \item Register Issues
    \item Ungrammatical
    \item Unintelligibility
    \item Unnatural Code-mixing
\end{enumerate}

The annotators were required to label the five most relevant issues in the translated text, and every sentence was evaluated by four annotators. It should be noted that translation performance can be highly subjective, especially for languages like Cantonese that do not have a well-defined standard norm for its written form. 

Some models exhibited a tendency to repeat the last sentence excessively, resulting in extra words beyond the expected translation.  To account for this, if the "Additional words" labelled constituted more than half of the generated output, the labelled additional words were removed and not counted as erroneous output. The accuracy of the translation was then calculated using the following formula:

\begin{equation}
    accuracy = \frac{len_{target} - len_{labelled~error_{target}}}{len_{target} + len_{labelled~error_{source}}}
\end{equation}

The average score across all translation pairs of all models can be found in Table \ref{tab:nlp-perf-table}.

\section{Summarisation}
\label{sec:appendix7}
The following prompt was used in the summarisation task evaluation:

\begin{quote}
    \raisebox{-0.15em}{\includegraphics[height=1.0em]{images/U+6211.png}}\hskip0pt{}\raisebox{-0.15em}{\includegraphics[height=1.0em]{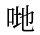}}\hskip0pt{}\raisebox{-0.15em}{\includegraphics[height=1.0em]{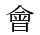}}\hskip0pt{}\raisebox{-0.15em}{\includegraphics[height=1.0em]{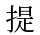}}\hskip0pt{}\raisebox{-0.15em}{\includegraphics[height=1.0em]{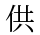}}\hskip0pt{}\raisebox{-0.15em}{\includegraphics[height=1.0em]{images/U+4E00.png}}\hskip0pt{}\raisebox{-0.15em}{\includegraphics[height=1.0em]{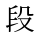}}\hskip0pt{}\raisebox{-0.15em}{\includegraphics[height=1.0em]{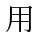}}\hskip0pt{}\raisebox{-0.15em}{\includegraphics[height=1.0em]{images/U+5EE3.png}}\hskip0pt{}\raisebox{-0.15em}{\includegraphics[height=1.0em]{images/U+6771.png}}\hskip0pt{}\raisebox{-0.15em}{\includegraphics[height=1.0em]{images/U+8A71.png}}\hskip0pt{}\raisebox{-0.15em}{\includegraphics[height=1.0em]{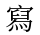}}\hskip0pt{}\raisebox{-0.15em}{\includegraphics[height=1.0em]{images/U+6210.png}}\hskip0pt{}\raisebox{-0.15em}{\includegraphics[height=1.0em]{images/U+5605.png}}\hskip0pt{}\raisebox{-0.15em}{\includegraphics[height=1.0em]{images/U+6587.png}}\hskip0pt{}\raisebox{-0.15em}{\includegraphics[height=1.0em]{images/U+672C.png}}\hskip0pt{}。\raisebox{-0.15em}{\includegraphics[height=1.0em]{images/U+8ACB.png}}\hskip0pt{}\raisebox{-0.15em}{\includegraphics[height=1.0em]{images/U+4F60.png}}\hskip0pt{}\raisebox{-0.15em}{\includegraphics[height=1.0em]{images/U+5C07.png}}\hskip0pt{}\raisebox{-0.15em}{\includegraphics[height=1.0em]{images/U+6587.png}}\hskip0pt{}\raisebox{-0.15em}{\includegraphics[height=1.0em]{images/U+672C.png}}\hskip0pt{}\raisebox{-0.15em}{\includegraphics[height=1.0em]{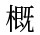}}\hskip0pt{}\raisebox{-0.15em}{\includegraphics[height=1.0em]{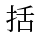}}\hskip0pt{}\raisebox{-0.15em}{\includegraphics[height=1.0em]{images/U+6210.png}}\hskip0pt{}200\raisebox{-0.15em}{\includegraphics[height=1.0em]{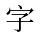}}\hskip0pt{}\raisebox{-0.15em}{\includegraphics[height=1.0em]{images/U+5605.png}}\hskip0pt{}\raisebox{-0.15em}{\includegraphics[height=1.0em]{images/U+5EE3.png}}\hskip0pt{}\raisebox{-0.15em}{\includegraphics[height=1.0em]{images/U+6771.png}}\hskip0pt{}\raisebox{-0.15em}{\includegraphics[height=1.0em]{images/U+8A71.png}}\hskip0pt{}，\raisebox{-0.15em}{\includegraphics[height=1.0em]{images/U+4FDD.png}}\hskip0pt{}\raisebox{-0.15em}{\includegraphics[height=1.0em]{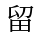}}\hskip0pt{}\raisebox{-0.15em}{\includegraphics[height=1.0em]{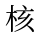}}\hskip0pt{}\raisebox{-0.15em}{\includegraphics[height=1.0em]{images/U+5FC3.png}}\hskip0pt{}\raisebox{-0.15em}{\includegraphics[height=1.0em]{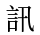}}\hskip0pt{}\raisebox{-0.15em}{\includegraphics[height=1.0em]{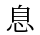}}\hskip0pt{}\raisebox{-0.15em}{\includegraphics[height=1.0em]{images/U+540C.png}}\hskip0pt{}\raisebox{-0.15em}{\includegraphics[height=1.0em]{images/U+4E3B.png}}\hskip0pt{}\raisebox{-0.15em}{\includegraphics[height=1.0em]{images/U+984C.png}}\hskip0pt{}，\raisebox{-0.15em}{\includegraphics[height=1.0em]{images/U+78BA.png}}\hskip0pt{}\raisebox{-0.15em}{\includegraphics[height=1.0em]{images/U+4FDD.png}}\hskip0pt{}\raisebox{-0.15em}{\includegraphics[height=1.0em]{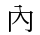}}\hskip0pt{}\raisebox{-0.15em}{\includegraphics[height=1.0em]{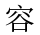}}\hskip0pt{}\raisebox{-0.15em}{\includegraphics[height=1.0em]{images/U+6E96.png}}\hskip0pt{}\raisebox{-0.15em}{\includegraphics[height=1.0em]{images/U+78BA.png}}\hskip0pt{}、\raisebox{-0.15em}{\includegraphics[height=1.0em]{images/U+884C.png}}\hskip0pt{}\raisebox{-0.15em}{\includegraphics[height=1.0em]{images/U+6587.png}}\hskip0pt{}\raisebox{-0.15em}{\includegraphics[height=1.0em]{images/U+6D41.png}}\hskip0pt{}\raisebox{-0.15em}{\includegraphics[height=1.0em]{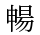}}\hskip0pt{}\raisebox{-0.15em}{\includegraphics[height=1.0em]{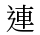}}\hskip0pt{}\raisebox{-0.15em}{\includegraphics[height=1.0em]{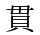}}\hskip0pt{}，\raisebox{-0.15em}{\includegraphics[height=1.0em]{images/U+4E26.png}}\hskip0pt{}\raisebox{-0.15em}{\includegraphics[height=1.0em]{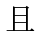}}\hskip0pt{}\raisebox{-0.15em}{\includegraphics[height=1.0em]{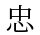}}\hskip0pt{}\raisebox{-0.15em}{\includegraphics[height=1.0em]{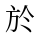}}\hskip0pt{}\raisebox{-0.15em}{\includegraphics[height=1.0em]{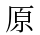}}\hskip0pt{}\raisebox{-0.15em}{\includegraphics[height=1.0em]{images/U+610F.png}}\hskip0pt{}。

\#\# \raisebox{-0.15em}{\includegraphics[height=1.0em]{images/U+6587.png}}\hskip0pt{}\raisebox{-0.15em}{\includegraphics[height=1.0em]{images/U+672C.png}}\hskip0pt{}（\raisebox{-0.15em}{\includegraphics[height=1.0em]{images/U+539F.png}}\hskip0pt{}\raisebox{-0.15em}{\includegraphics[height=1.0em]{images/U+6587.png}}\hskip0pt{}）：
\textit{(Original Text)}

\#\# \raisebox{-0.15em}{\includegraphics[height=1.0em]{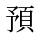}}\hskip0pt{}\raisebox{-0.15em}{\includegraphics[height=1.0em]{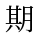}}\hskip0pt{}\raisebox{-0.15em}{\includegraphics[height=1.0em]{images/U+8F38.png}}\hskip0pt{}\raisebox{-0.15em}{\includegraphics[height=1.0em]{images/U+51FA.png}}\hskip0pt{}：
\end{quote}

English translation of the prompt:
\begin{quote}
    We will provide a text written in Cantonese. Please summarise the text into a 200 words Cantonese text while preserving the core message and theme. Ensure the content is accurate, the writing flows smoothly and coherently, and it remains faithful to the original meaning.

\#\# Text（Original text）：\textit{(Original Text)}

\#\# Expected Output：
\end{quote}

\subsection{Summarisation Results Evaluation}
\label{sec:appendix7_1}
For the summarisation outputs, annotators (paid undergraduate students and research assistants from Hong Kong) were given the following rubric and tasked to give a score for each output. Under each category, the annotators rated whether the summary passes (5 points) or fails (3 points for minor violation, 1 point for complete failure).

Task instructions given to annotators: 

\begin{quote}
\raisebox{-0.15em}{\includegraphics[height=1.0em]{images/U+4F60.png}}\hskip0pt{}\raisebox{-0.15em}{\includegraphics[height=1.0em]{images/U+6703.png}}\hskip0pt{}\raisebox{-0.15em}{\includegraphics[height=1.0em]{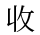}}\hskip0pt{}\raisebox{-0.15em}{\includegraphics[height=1.0em]{images/U+5230.png}}\hskip0pt{}\raisebox{-0.15em}{\includegraphics[height=1.0em]{images/U+591A.png}}\hskip0pt{}\raisebox{-0.15em}{\includegraphics[height=1.0em]{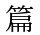}}\hskip0pt{} TED \raisebox{-0.15em}{\includegraphics[height=1.0em]{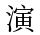}}\hskip0pt{}\raisebox{-0.15em}{\includegraphics[height=1.0em]{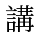}}\hskip0pt{}\raisebox{-0.15em}{\includegraphics[height=1.0em]{images/U+5605.png}}\hskip0pt{}\raisebox{-0.15em}{\includegraphics[height=1.0em]{images/U+5EE3.png}}\hskip0pt{}\raisebox{-0.15em}{\includegraphics[height=1.0em]{images/U+6771.png}}\hskip0pt{}\raisebox{-0.15em}{\includegraphics[height=1.0em]{images/U+8A71.png}}\hskip0pt{}\raisebox{-0.15em}{\includegraphics[height=1.0em]{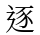}}\hskip0pt{}\raisebox{-0.15em}{\includegraphics[height=1.0em]{images/U+5B57.png}}\hskip0pt{}\raisebox{-0.15em}{\includegraphics[height=1.0em]{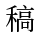}}\hskip0pt{}\raisebox{-0.15em}{\includegraphics[height=1.0em]{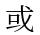}}\hskip0pt{}\raisebox{-0.15em}{\includegraphics[height=1.0em]{images/U+5EE3.png}}\hskip0pt{}\raisebox{-0.15em}{\includegraphics[height=1.0em]{images/U+6771.png}}\hskip0pt{}\raisebox{-0.15em}{\includegraphics[height=1.0em]{images/U+8A71.png}}\hskip0pt{}\raisebox{-0.15em}{\includegraphics[height=1.0em]{images/U+6587.png}}\hskip0pt{}\raisebox{-0.15em}{\includegraphics[height=1.0em]{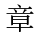}}\hskip0pt{}（\raisebox{-0.15em}{\includegraphics[height=1.0em]{images/U+539F.png}}\hskip0pt{}\raisebox{-0.15em}{\includegraphics[height=1.0em]{images/U+6587.png}}\hskip0pt{}）
\raisebox{-0.15em}{\includegraphics[height=1.0em]{images/U+4F60.png}}\hskip0pt{}\raisebox{-0.15em}{\includegraphics[height=1.0em]{images/U+5605.png}}\hskip0pt{}\raisebox{-0.15em}{\includegraphics[height=1.0em]{images/U+5DE5.png}}\hskip0pt{}\raisebox{-0.15em}{\includegraphics[height=1.0em]{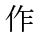}}\hskip0pt{}\raisebox{-0.15em}{\includegraphics[height=1.0em]{images/U+4FC2.png}}\hskip0pt{}\raisebox{-0.15em}{\includegraphics[height=1.0em]{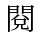}}\hskip0pt{}\raisebox{-0.15em}{\includegraphics[height=1.0em]{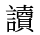}}\hskip0pt{}\raisebox{-0.15em}{\includegraphics[height=1.0em]{images/U+539F.png}}\hskip0pt{}\raisebox{-0.15em}{\includegraphics[height=1.0em]{images/U+6587.png}}\hskip0pt{}，\raisebox{-0.15em}{\includegraphics[height=1.0em]{images/U+7136.png}}\hskip0pt{}\raisebox{-0.15em}{\includegraphics[height=1.0em]{images/U+5F8C.png}}\hskip0pt{}\raisebox{-0.15em}{\includegraphics[height=1.0em]{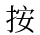}}\hskip0pt{}\raisebox{-0.15em}{\includegraphics[height=1.0em]{images/U+7167.png}}\hskip0pt{}\raisebox{-0.15em}{\includegraphics[height=1.0em]{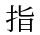}}\hskip0pt{}\raisebox{-0.15em}{\includegraphics[height=1.0em]{images/U+5B9A.png}}\hskip0pt{}\raisebox{-0.15em}{\includegraphics[height=1.0em]{images/U+6E96.png}}\hskip0pt{}\raisebox{-0.15em}{\includegraphics[height=1.0em]{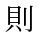}}\hskip0pt{}\raisebox{-0.15em}{\includegraphics[height=1.0em]{images/U+70BA.png}}\hskip0pt{}\raisebox{-0.15em}{\includegraphics[height=1.0em]{images/U+4E0D.png}}\hskip0pt{}\raisebox{-0.15em}{\includegraphics[height=1.0em]{images/U+540C.png}}\hskip0pt{}\raisebox{-0.15em}{\includegraphics[height=1.0em]{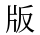}}\hskip0pt{}\raisebox{-0.15em}{\includegraphics[height=1.0em]{images/U+672C.png}}\hskip0pt{}\raisebox{-0.15em}{\includegraphics[height=1.0em]{images/U+5605.png}}\hskip0pt{}\raisebox{-0.15em}{\includegraphics[height=1.0em]{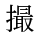}}\hskip0pt{}\raisebox{-0.15em}{\includegraphics[height=1.0em]{images/U+5BEB.png}}\hskip0pt{}\raisebox{-0.15em}{\includegraphics[height=1.0em]{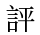}}\hskip0pt{}\raisebox{-0.15em}{\includegraphics[height=1.0em]{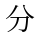}}\hskip0pt{}
\raisebox{-0.15em}{\includegraphics[height=1.0em]{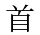}}\hskip0pt{}\raisebox{-0.15em}{\includegraphics[height=1.0em]{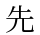}}\hskip0pt{}\raisebox{-0.15em}{\includegraphics[height=1.0em]{images/U+4F60.png}}\hskip0pt{}\raisebox{-0.15em}{\includegraphics[height=1.0em]{images/U+9700.png}}\hskip0pt{}\raisebox{-0.15em}{\includegraphics[height=1.0em]{images/U+8981.png}}\hskip0pt{}\raisebox{-0.15em}{\includegraphics[height=1.0em]{images/U+95B1.png}}\hskip0pt{}\raisebox{-0.15em}{\includegraphics[height=1.0em]{images/U+8B80.png}}\hskip0pt{}\raisebox{-0.15em}{\includegraphics[height=1.0em]{images/U+539F.png}}\hskip0pt{}\raisebox{-0.15em}{\includegraphics[height=1.0em]{images/U+6587.png}}\hskip0pt{}，\raisebox{-0.15em}{\includegraphics[height=1.0em]{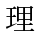}}\hskip0pt{}\raisebox{-0.15em}{\includegraphics[height=1.0em]{images/U+89E3.png}}\hskip0pt{}\raisebox{-0.15em}{\includegraphics[height=1.0em]{images/U+6587.png}}\hskip0pt{}\raisebox{-0.15em}{\includegraphics[height=1.0em]{images/U+7AE0.png}}\hskip0pt{}\raisebox{-0.15em}{\includegraphics[height=1.0em]{images/U+5927.png}}\hskip0pt{}\raisebox{-0.15em}{\includegraphics[height=1.0em]{images/U+610F.png}}\hskip0pt{}
\raisebox{-0.15em}{\includegraphics[height=1.0em]{images/U+53EF.png}}\hskip0pt{}\raisebox{-0.15em}{\includegraphics[height=1.0em]{images/U+4EE5.png}}\hskip0pt{}\raisebox{-0.15em}{\includegraphics[height=1.0em]{images/U+8B80.png}}\hskip0pt{}\raisebox{-0.15em}{\includegraphics[height=1.0em]{images/U+591A.png}}\hskip0pt{}\raisebox{-0.15em}{\includegraphics[height=1.0em]{images/U+904E.png}}\hskip0pt{}\raisebox{-0.15em}{\includegraphics[height=1.0em]{images/U+4E00.png}}\hskip0pt{}\raisebox{-0.15em}{\includegraphics[height=1.0em]{images/U+6B21.png}}\hskip0pt{}
\raisebox{-0.15em}{\includegraphics[height=1.0em]{images/U+7136.png}}\hskip0pt{}\raisebox{-0.15em}{\includegraphics[height=1.0em]{images/U+5F8C.png}}\hskip0pt{}\raisebox{-0.15em}{\includegraphics[height=1.0em]{images/U+958B.png}}\hskip0pt{}\raisebox{-0.15em}{\includegraphics[height=1.0em]{images/U+59CB.png}}\hskip0pt{}\raisebox{-0.15em}{\includegraphics[height=1.0em]{images/U+9010.png}}\hskip0pt{}\raisebox{-0.15em}{\includegraphics[height=1.0em]{images/U+7BC7.png}}\hskip0pt{}\raisebox{-0.15em}{\includegraphics[height=1.0em]{images/U+64AE.png}}\hskip0pt{}\raisebox{-0.15em}{\includegraphics[height=1.0em]{images/U+5BEB.png}}\hskip0pt{}\raisebox{-0.15em}{\includegraphics[height=1.0em]{images/U+95B1.png}}\hskip0pt{}\raisebox{-0.15em}{\includegraphics[height=1.0em]{images/U+8B80.png}}\hskip0pt{}
\raisebox{-0.15em}{\includegraphics[height=1.0em]{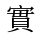}}\hskip0pt{}\raisebox{-0.15em}{\includegraphics[height=1.0em]{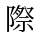}}\hskip0pt{}\raisebox{-0.15em}{\includegraphics[height=1.0em]{images/U+8A55.png}}\hskip0pt{}\raisebox{-0.15em}{\includegraphics[height=1.0em]{images/U+5206.png}}\hskip0pt{}\raisebox{-0.15em}{\includegraphics[height=1.0em]{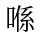}}\hskip0pt{} Google Form \raisebox{-0.15em}{\includegraphics[height=1.0em]{images/U+4E0A.png}}\hskip0pt{}\raisebox{-0.15em}{\includegraphics[height=1.0em]{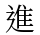}}\hskip0pt{}\raisebox{-0.15em}{\includegraphics[height=1.0em]{images/U+884C.png}}\hskip0pt{}
\end{quote}

English Translation: 
\begin{quote}
You will receive multiple Cantonese transcripts of TED talks or passages (\textit{Original Text}). Your task is to read the original text and then score different versions of the summaries according to the specified guidelines. 
First, you need to read the original text to understand the main idea of the article. You may read it more than once. 
Then, begin reading the summaries one by one.
The actual scoring will be conducted on a Google Form.
\end{quote}

The annotation rubrics were explained in detail in face-to-face meetings and in online working chat. A manually-crafted marking scheme was prepared to ease the annotation task.

Annotation Rubrics (English Translation):
\begin{itemize}
    \item \textbf{Category: Relevance} \\
    The summary successfully retains the key points of the original text, rather than extracting overly verbose content, judged by whether the summary contains all the key points listed in the reference marking scheme provided. Omitted content and problematic sentences should be added as a comment if a fail (3 or 1) rating is given.
    
    \item \textbf{Category: Accuracy} \\
    The summary correctly extracted information from the original text, judging by comparing sentence by sentence with the original text, to identify any completely opposite or fabricated content. Sentences that contain incorrect information should be added as a comment if a fail rating is given.
    
    \item \textbf{Category: Fluency}\\
    The words and sentence structures used in the summary are smooth and conforming to Cantonese usage. This score does not require referencing the original text. The text should be acceptable for reading on a news programme. If words that are exclusive to Written Chinese (e.g. \raisebox{-0.15em}{\includegraphics[height=1.0em]{images/U+9019.png}}\hskip0pt{}, \raisebox{-0.15em}{\includegraphics[height=1.0em]{images/U+7684.png}}\hskip0pt{}, \raisebox{-0.15em}{\includegraphics[height=1.0em]{images/U+4E86.png}}\hskip0pt{}) or grammatical issues were found, a fail rating should be given.

    \item \textbf{Category: Coherence} \\
    The sentences and paragraphs in the summary are logically coherent, with appropriate structure and adequate sentential connection. This score does not require referencing the original text.
\end{itemize}

\textit{Score Calculation: }Scores from all raters on the four categories were aggregated and normalised to 100\%, then a length penalty is applied, based on its length: if the summary exceeds 500 characters, a penalty factor is calculated by reducing the score proportionally to the excess length, ensuring the penalty factor remains between 0 and 1. The resultant score can be found in Table \ref{tab:nlp-perf-table}.

\section{Sentiment Analysis}
\label{sec:appendix8}
As detailed in Section 3.5, the sentiment analysis dataset consists of an existing dataset known as the OpenRice dataset \citep{openrice} and a newly created dataset. The newly created datasets were sourced from Facebook comments and filtered by CantoneseDetect \citep{lau2024extraction}. Paid interns (undergraduate students from Hong Kong) then labelled the comments with the following guidelines:

\begin{quote}
\raisebox{-0.15em}{\includegraphics[height=1.0em]{images/U+5462.png}}\hskip0pt{}\raisebox{-0.15em}{\includegraphics[height=1.0em]{images/U+500B.png}}\hskip0pt{}\raisebox{-0.15em}{\includegraphics[height=1.0em]{images/U+5DE5.png}}\hskip0pt{}\raisebox{-0.15em}{\includegraphics[height=1.0em]{images/U+4F5C.png}}\hskip0pt{}\raisebox{-0.15em}{\includegraphics[height=1.0em]{images/U+5605.png}}\hskip0pt{}\raisebox{-0.15em}{\includegraphics[height=1.0em]{images/U+76EE.png}}\hskip0pt{}\raisebox{-0.15em}{\includegraphics[height=1.0em]{images/U+6A19.png}}\hskip0pt{}\raisebox{-0.15em}{\includegraphics[height=1.0em]{images/U+4FC2.png}}\hskip0pt{}\raisebox{-0.15em}{\includegraphics[height=1.0em]{images/U+8981.png}}\hskip0pt{}\raisebox{-0.15em}{\includegraphics[height=1.0em]{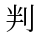}}\hskip0pt{}\raisebox{-0.15em}{\includegraphics[height=1.0em]{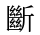}}\hskip0pt{}\raisebox{-0.15em}{\includegraphics[height=1.0em]{images/U+4E00.png}}\hskip0pt{}\raisebox{-0.15em}{\includegraphics[height=1.0em]{images/U+6BB5.png}}\hskip0pt{}\raisebox{-0.15em}{\includegraphics[height=1.0em]{images/U+6587.png}}\hskip0pt{}\raisebox{-0.15em}{\includegraphics[height=1.0em]{images/U+5B57.png}}\hskip0pt{}\raisebox{-0.15em}{\includegraphics[height=1.0em]{images/U+8868.png}}\hskip0pt{}\raisebox{-0.15em}{\includegraphics[height=1.0em]{images/U+9054.png}}\hskip0pt{}\raisebox{-0.15em}{\includegraphics[height=1.0em]{images/U+5605.png}}\hskip0pt{}\raisebox{-0.15em}{\includegraphics[height=1.0em]{images/U+60C5.png}}\hskip0pt{}\raisebox{-0.15em}{\includegraphics[height=1.0em]{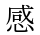}}\hskip0pt{}\raisebox{-0.15em}{\includegraphics[height=1.0em]{images/U+4FC2.png}}\hskip0pt{}\raisebox{-0.15em}{\includegraphics[height=1.0em]{images/U+6B63.png}}\hskip0pt{}\raisebox{-0.15em}{\includegraphics[height=1.0em]{images/U+9762.png}}\hskip0pt{}(positive)、\raisebox{-0.15em}{\includegraphics[height=1.0em]{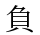}}\hskip0pt{}\raisebox{-0.15em}{\includegraphics[height=1.0em]{images/U+9762.png}}\hskip0pt{}(negative)，\raisebox{-0.15em}{\includegraphics[height=1.0em]{images/U+5B9A.png}}\hskip0pt{}\raisebox{-0.15em}{\includegraphics[height=1.0em]{images/U+4FC2.png}}\hskip0pt{}\raisebox{-0.15em}{\includegraphics[height=1.0em]{images/U+4E2D.png}}\hskip0pt{}\raisebox{-0.15em}{\includegraphics[height=1.0em]{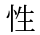}}\hskip0pt{}(neutral)。\raisebox{-0.15em}{\includegraphics[height=1.0em]{images/U+6211.png}}\hskip0pt{}\raisebox{-0.15em}{\includegraphics[height=1.0em]{images/U+54CB.png}}\hskip0pt{}\raisebox{-0.15em}{\includegraphics[height=1.0em]{images/U+6703.png}}\hskip0pt{}\raisebox{-0.15em}{\includegraphics[height=1.0em]{images/U+7528.png}}\hskip0pt{}\raisebox{-0.15em}{\includegraphics[height=1.0em]{images/U+4E09.png}}\hskip0pt{}\raisebox{-0.15em}{\includegraphics[height=1.0em]{images/U+500B.png}}\hskip0pt{}\raisebox{-0.15em}{\includegraphics[height=1.0em]{images/U+6A19.png}}\hskip0pt{}\raisebox{-0.15em}{\includegraphics[height=1.0em]{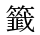}}\hskip0pt{}：\raisebox{-0.15em}{\includegraphics[height=1.0em]{images/U+6B63.png}}\hskip0pt{}\raisebox{-0.15em}{\includegraphics[height=1.0em]{images/U+9762.png}}\hskip0pt{}(positive), \raisebox{-0.15em}{\includegraphics[height=1.0em]{images/U+8CA0.png}}\hskip0pt{}\raisebox{-0.15em}{\includegraphics[height=1.0em]{images/U+9762.png}}\hskip0pt{}(negative), \raisebox{-0.15em}{\includegraphics[height=1.0em]{images/U+540C.png}}\hskip0pt{}\raisebox{-0.15em}{\includegraphics[height=1.0em]{images/U+57CB.png}}\hskip0pt{}\raisebox{-0.15em}{\includegraphics[height=1.0em]{images/U+4E2D.png}}\hskip0pt{}\raisebox{-0.15em}{\includegraphics[height=1.0em]{images/U+6027.png}}\hskip0pt{}(neutral)。\raisebox{-0.15em}{\includegraphics[height=1.0em]{images/U+8ACB.png}}\hskip0pt{}\raisebox{-0.15em}{\includegraphics[height=1.0em]{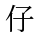}}\hskip0pt{}\raisebox{-0.15em}{\includegraphics[height=1.0em]{images/U+7D30.png}}\hskip0pt{}\raisebox{-0.15em}{\includegraphics[height=1.0em]{images/U+95B1.png}}\hskip0pt{}\raisebox{-0.15em}{\includegraphics[height=1.0em]{images/U+8B80.png}}\hskip0pt{}\raisebox{-0.15em}{\includegraphics[height=1.0em]{images/U+4EE5.png}}\hskip0pt{}\raisebox{-0.15em}{\includegraphics[height=1.0em]{images/U+4E0B.png}}\hskip0pt{}\raisebox{-0.15em}{\includegraphics[height=1.0em]{images/U+5605.png}}\hskip0pt{}\raisebox{-0.15em}{\includegraphics[height=1.0em]{images/U+6307.png}}\hskip0pt{}\raisebox{-0.15em}{\includegraphics[height=1.0em]{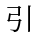}}\hskip0pt{}，\raisebox{-0.15em}{\includegraphics[height=1.0em]{images/U+78BA.png}}\hskip0pt{}\raisebox{-0.15em}{\includegraphics[height=1.0em]{images/U+4FDD.png}}\hskip0pt{}\raisebox{-0.15em}{\includegraphics[height=1.0em]{images/U+6A19.png}}\hskip0pt{}\raisebox{-0.15em}{\includegraphics[height=1.0em]{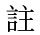}}\hskip0pt{}\raisebox{-0.15em}{\includegraphics[height=1.0em]{images/U+5605.png}}\hskip0pt{}\raisebox{-0.15em}{\includegraphics[height=1.0em]{images/U+4E00.png}}\hskip0pt{}\raisebox{-0.15em}{\includegraphics[height=1.0em]{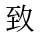}}\hskip0pt{}\raisebox{-0.15em}{\includegraphics[height=1.0em]{images/U+6027.png}}\hskip0pt{}\raisebox{-0.15em}{\includegraphics[height=1.0em]{images/U+540C.png}}\hskip0pt{}\raisebox{-0.15em}{\includegraphics[height=1.0em]{images/U+57CB.png}}\hskip0pt{}\raisebox{-0.15em}{\includegraphics[height=1.0em]{images/U+6E96.png}}\hskip0pt{}\raisebox{-0.15em}{\includegraphics[height=1.0em]{images/U+78BA.png}}\hskip0pt{}\raisebox{-0.15em}{\includegraphics[height=1.0em]{images/U+6027.png}}\hskip0pt{}。

\textbf{\raisebox{-0.15em}{\includegraphics[height=1.0em]{images/U+6B63.png}}\hskip0pt{}\raisebox{-0.15em}{\includegraphics[height=1.0em]{images/U+9762.png}}\hskip0pt{}(positive)}\\
\begin{enumerate}
    \item \raisebox{-0.15em}{\includegraphics[height=1.0em]{images/U+610F.png}}\hskip0pt{}\raisebox{-0.15em}{\includegraphics[height=1.0em]{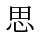}}\hskip0pt{}: \raisebox{-0.15em}{\includegraphics[height=1.0em]{images/U+8868.png}}\hskip0pt{}\raisebox{-0.15em}{\includegraphics[height=1.0em]{images/U+9054.png}}\hskip0pt{}\raisebox{-0.15em}{\includegraphics[height=1.0em]{images/U+958B.png}}\hskip0pt{}\raisebox{-0.15em}{\includegraphics[height=1.0em]{images/U+5FC3.png}}\hskip0pt{}、\raisebox{-0.15em}{\includegraphics[height=1.0em]{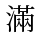}}\hskip0pt{}\raisebox{-0.15em}{\includegraphics[height=1.0em]{images/U+610F.png}}\hskip0pt{}、\raisebox{-0.15em}{\includegraphics[height=1.0em]{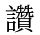}}\hskip0pt{}\raisebox{-0.15em}{\includegraphics[height=1.0em]{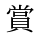}}\hskip0pt{}、\raisebox{-0.15em}{\includegraphics[height=1.0em]{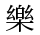}}\hskip0pt{}\raisebox{-0.15em}{\includegraphics[height=1.0em]{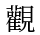}}\hskip0pt{}、\raisebox{-0.15em}{\includegraphics[height=1.0em]{images/U+6216.png}}\hskip0pt{}\raisebox{-0.15em}{\includegraphics[height=1.0em]{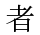}}\hskip0pt{}\raisebox{-0.15em}{\includegraphics[height=1.0em]{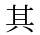}}\hskip0pt{}\raisebox{-0.15em}{\includegraphics[height=1.0em]{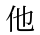}}\hskip0pt{}\raisebox{-0.15em}{\includegraphics[height=1.0em]{images/U+6B63.png}}\hskip0pt{}\raisebox{-0.15em}{\includegraphics[height=1.0em]{images/U+9762.png}}\hskip0pt{}\raisebox{-0.15em}{\includegraphics[height=1.0em]{images/U+5605.png}}\hskip0pt{}\raisebox{-0.15em}{\includegraphics[height=1.0em]{images/U+60C5.png}}\hskip0pt{}\raisebox{-0.15em}{\includegraphics[height=1.0em]{images/U+611F.png}}\hskip0pt{}。
    \item \raisebox{-0.15em}{\includegraphics[height=1.0em]{images/U+89E3.png}}\hskip0pt{}\raisebox{-0.15em}{\includegraphics[height=1.0em]{images/U+91CB.png}}\hskip0pt{}: \raisebox{-0.15em}{\includegraphics[height=1.0em]{images/U+6587.png}}\hskip0pt{}\raisebox{-0.15em}{\includegraphics[height=1.0em]{images/U+5B57.png}}\hskip0pt{}\raisebox{-0.15em}{\includegraphics[height=1.0em]{images/U+5165.png}}\hskip0pt{}\raisebox{-0.15em}{\includegraphics[height=1.0em]{images/U+9762.png}}\hskip0pt{}\raisebox{-0.15em}{\includegraphics[height=1.0em]{images/U+5605.png}}\hskip0pt{}\raisebox{-0.15em}{\includegraphics[height=1.0em]{images/U+5167.png}}\hskip0pt{}\raisebox{-0.15em}{\includegraphics[height=1.0em]{images/U+5BB9.png}}\hskip0pt{}\raisebox{-0.15em}{\includegraphics[height=1.0em]{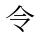}}\hskip0pt{}\raisebox{-0.15em}{\includegraphics[height=1.0em]{images/U+4EBA.png}}\hskip0pt{}\raisebox{-0.15em}{\includegraphics[height=1.0em]{images/U+611F.png}}\hskip0pt{}\raisebox{-0.15em}{\includegraphics[height=1.0em]{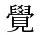}}\hskip0pt{}\raisebox{-0.15em}{\includegraphics[height=1.0em]{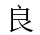}}\hskip0pt{}\raisebox{-0.15em}{\includegraphics[height=1.0em]{images/U+597D.png}}\hskip0pt{}、\raisebox{-0.15em}{\includegraphics[height=1.0em]{images/U+958B.png}}\hskip0pt{}\raisebox{-0.15em}{\includegraphics[height=1.0em]{images/U+5FC3.png}}\hskip0pt{}、\raisebox{-0.15em}{\includegraphics[height=1.0em]{images/U+6216.png}}\hskip0pt{}\raisebox{-0.15em}{\includegraphics[height=1.0em]{images/U+8005.png}}\hskip0pt{}\raisebox{-0.15em}{\includegraphics[height=1.0em]{images/U+5C0D.png}}\hskip0pt{}\raisebox{-0.15em}{\includegraphics[height=1.0em]{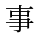}}\hskip0pt{}\raisebox{-0.15em}{\includegraphics[height=1.0em]{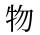}}\hskip0pt{}\raisebox{-0.15em}{\includegraphics[height=1.0em]{images/U+6709.png}}\hskip0pt{}\raisebox{-0.15em}{\includegraphics[height=1.0em]{images/U+6B63.png}}\hskip0pt{}\raisebox{-0.15em}{\includegraphics[height=1.0em]{images/U+9762.png}}\hskip0pt{}\raisebox{-0.15em}{\includegraphics[height=1.0em]{images/U+8A55.png}}\hskip0pt{}\raisebox{-0.15em}{\includegraphics[height=1.0em]{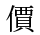}}\hskip0pt{}。
    \item \raisebox{-0.15em}{\includegraphics[height=1.0em]{images/U+4F8B.png}}\hskip0pt{}\raisebox{-0.15em}{\includegraphics[height=1.0em]{images/U+5B50.png}}\hskip0pt{}: ``\raisebox{-0.15em}{\includegraphics[height=1.0em]{images/U+5462.png}}\hskip0pt{}\raisebox{-0.15em}{\includegraphics[height=1.0em]{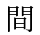}}\hskip0pt{}\raisebox{-0.15em}{\includegraphics[height=1.0em]{images/U+9910.png}}\hskip0pt{}\raisebox{-0.15em}{\includegraphics[height=1.0em]{images/U+5EF3.png}}\hskip0pt{}\raisebox{-0.15em}{\includegraphics[height=1.0em]{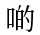}}\hskip0pt{}\raisebox{-0.15em}{\includegraphics[height=1.0em]{images/U+5622.png}}\hskip0pt{}\raisebox{-0.15em}{\includegraphics[height=1.0em]{images/U+98DF.png}}\hskip0pt{}\raisebox{-0.15em}{\includegraphics[height=1.0em]{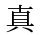}}\hskip0pt{}\raisebox{-0.15em}{\includegraphics[height=1.0em]{images/U+4FC2.png}}\hskip0pt{}\raisebox{-0.15em}{\includegraphics[height=1.0em]{images/U+597D.png}}\hskip0pt{}\raisebox{-0.15em}{\includegraphics[height=1.0em]{images/U+597D.png}}\hskip0pt{}\raisebox{-0.15em}{\includegraphics[height=1.0em]{images/U+5473.png}}\hskip0pt{}！\raisebox{-0.15em}{\includegraphics[height=1.0em]{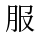}}\hskip0pt{}\raisebox{-0.15em}{\includegraphics[height=1.0em]{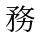}}\hskip0pt{}\raisebox{-0.15em}{\includegraphics[height=1.0em]{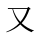}}\hskip0pt{}\raisebox{-0.15em}{\includegraphics[height=1.0em]{images/U+597D.png}}\hskip0pt{}！''
    \item \raisebox{-0.15em}{\includegraphics[height=1.0em]{images/U+4F8B.png}}\hskip0pt{}\raisebox{-0.15em}{\includegraphics[height=1.0em]{images/U+5B50.png}}\hskip0pt{}: ``\raisebox{-0.15em}{\includegraphics[height=1.0em]{images/U+4ECA.png}}\hskip0pt{}\raisebox{-0.15em}{\includegraphics[height=1.0em]{images/U+65E5.png}}\hskip0pt{}\raisebox{-0.15em}{\includegraphics[height=1.0em]{images/U+5929.png}}\hskip0pt{}\raisebox{-0.15em}{\includegraphics[height=1.0em]{images/U+6C23.png}}\hskip0pt{}\raisebox{-0.15em}{\includegraphics[height=1.0em]{images/U+597D.png}}\hskip0pt{}\raisebox{-0.15em}{\includegraphics[height=1.0em]{images/U+597D.png}}\hskip0pt{}，\raisebox{-0.15em}{\includegraphics[height=1.0em]{images/U+5FC3.png}}\hskip0pt{}\raisebox{-0.15em}{\includegraphics[height=1.0em]{images/U+60C5.png}}\hskip0pt{}\raisebox{-0.15em}{\includegraphics[height=1.0em]{images/U+90FD.png}}\hskip0pt{}\raisebox{-0.15em}{\includegraphics[height=1.0em]{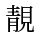}}\hskip0pt{}\raisebox{-0.15em}{\includegraphics[height=1.0em]{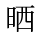}}\hskip0pt{}！''
\end{enumerate}

\textbf{\raisebox{-0.15em}{\includegraphics[height=1.0em]{images/U+8CA0.png}}\hskip0pt{}\raisebox{-0.15em}{\includegraphics[height=1.0em]{images/U+9762.png}}\hskip0pt{}(negative)}\\
    \begin{enumerate}
        \item \raisebox{-0.15em}{\includegraphics[height=1.0em]{images/U+610F.png}}\hskip0pt{}\raisebox{-0.15em}{\includegraphics[height=1.0em]{images/U+601D.png}}\hskip0pt{}: \raisebox{-0.15em}{\includegraphics[height=1.0em]{images/U+8868.png}}\hskip0pt{}\raisebox{-0.15em}{\includegraphics[height=1.0em]{images/U+9054.png}}\hskip0pt{}\raisebox{-0.15em}{\includegraphics[height=1.0em]{images/U+5514.png}}\hskip0pt{}\raisebox{-0.15em}{\includegraphics[height=1.0em]{images/U+958B.png}}\hskip0pt{}\raisebox{-0.15em}{\includegraphics[height=1.0em]{images/U+5FC3.png}}\hskip0pt{}、\raisebox{-0.15em}{\includegraphics[height=1.0em]{images/U+5514.png}}\hskip0pt{}\raisebox{-0.15em}{\includegraphics[height=1.0em]{images/U+6EFF.png}}\hskip0pt{}\raisebox{-0.15em}{\includegraphics[height=1.0em]{images/U+610F.png}}\hskip0pt{}、\raisebox{-0.15em}{\includegraphics[height=1.0em]{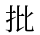}}\hskip0pt{}\raisebox{-0.15em}{\includegraphics[height=1.0em]{images/U+8A55.png}}\hskip0pt{}、\raisebox{-0.15em}{\includegraphics[height=1.0em]{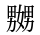}}\hskip0pt{}\raisebox{-0.15em}{\includegraphics[height=1.0em]{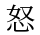}}\hskip0pt{}、\raisebox{-0.15em}{\includegraphics[height=1.0em]{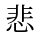}}\hskip0pt{}\raisebox{-0.15em}{\includegraphics[height=1.0em]{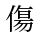}}\hskip0pt{}、\raisebox{-0.15em}{\includegraphics[height=1.0em]{images/U+6216.png}}\hskip0pt{}\raisebox{-0.15em}{\includegraphics[height=1.0em]{images/U+8005.png}}\hskip0pt{}\raisebox{-0.15em}{\includegraphics[height=1.0em]{images/U+5176.png}}\hskip0pt{}\raisebox{-0.15em}{\includegraphics[height=1.0em]{images/U+4ED6.png}}\hskip0pt{}\raisebox{-0.15em}{\includegraphics[height=1.0em]{images/U+8CA0.png}}\hskip0pt{}\raisebox{-0.15em}{\includegraphics[height=1.0em]{images/U+9762.png}}\hskip0pt{}\raisebox{-0.15em}{\includegraphics[height=1.0em]{images/U+5605.png}}\hskip0pt{}\raisebox{-0.15em}{\includegraphics[height=1.0em]{images/U+60C5.png}}\hskip0pt{}\raisebox{-0.15em}{\includegraphics[height=1.0em]{images/U+611F.png}}\hskip0pt{}。
        \item \raisebox{-0.15em}{\includegraphics[height=1.0em]{images/U+89E3.png}}\hskip0pt{}\raisebox{-0.15em}{\includegraphics[height=1.0em]{images/U+91CB.png}}\hskip0pt{}: \raisebox{-0.15em}{\includegraphics[height=1.0em]{images/U+6587.png}}\hskip0pt{}\raisebox{-0.15em}{\includegraphics[height=1.0em]{images/U+5B57.png}}\hskip0pt{}\raisebox{-0.15em}{\includegraphics[height=1.0em]{images/U+5165.png}}\hskip0pt{}\raisebox{-0.15em}{\includegraphics[height=1.0em]{images/U+9762.png}}\hskip0pt{}\raisebox{-0.15em}{\includegraphics[height=1.0em]{images/U+5605.png}}\hskip0pt{}\raisebox{-0.15em}{\includegraphics[height=1.0em]{images/U+5167.png}}\hskip0pt{}\raisebox{-0.15em}{\includegraphics[height=1.0em]{images/U+5BB9.png}}\hskip0pt{}\raisebox{-0.15em}{\includegraphics[height=1.0em]{images/U+4EE4.png}}\hskip0pt{}\raisebox{-0.15em}{\includegraphics[height=1.0em]{images/U+4EBA.png}}\hskip0pt{}\raisebox{-0.15em}{\includegraphics[height=1.0em]{images/U+611F.png}}\hskip0pt{}\raisebox{-0.15em}{\includegraphics[height=1.0em]{images/U+89BA.png}}\hskip0pt{}\raisebox{-0.15em}{\includegraphics[height=1.0em]{images/U+5514.png}}\hskip0pt{}\raisebox{-0.15em}{\includegraphics[height=1.0em]{images/U+597D.png}}\hskip0pt{}、\raisebox{-0.15em}{\includegraphics[height=1.0em]{images/U+5514.png}}\hskip0pt{}\raisebox{-0.15em}{\includegraphics[height=1.0em]{images/U+958B.png}}\hskip0pt{}\raisebox{-0.15em}{\includegraphics[height=1.0em]{images/U+5FC3.png}}\hskip0pt{}、\raisebox{-0.15em}{\includegraphics[height=1.0em]{images/U+6216.png}}\hskip0pt{}\raisebox{-0.15em}{\includegraphics[height=1.0em]{images/U+8005.png}}\hskip0pt{}\raisebox{-0.15em}{\includegraphics[height=1.0em]{images/U+5C0D.png}}\hskip0pt{}\raisebox{-0.15em}{\includegraphics[height=1.0em]{images/U+4E8B.png}}\hskip0pt{}\raisebox{-0.15em}{\includegraphics[height=1.0em]{images/U+7269.png}}\hskip0pt{}\raisebox{-0.15em}{\includegraphics[height=1.0em]{images/U+6709.png}}\hskip0pt{}\raisebox{-0.15em}{\includegraphics[height=1.0em]{images/U+8CA0.png}}\hskip0pt{}\raisebox{-0.15em}{\includegraphics[height=1.0em]{images/U+9762.png}}\hskip0pt{}\raisebox{-0.15em}{\includegraphics[height=1.0em]{images/U+8A55.png}}\hskip0pt{}\raisebox{-0.15em}{\includegraphics[height=1.0em]{images/U+50F9.png}}\hskip0pt{}。
        \item \raisebox{-0.15em}{\includegraphics[height=1.0em]{images/U+4F8B.png}}\hskip0pt{}\raisebox{-0.15em}{\includegraphics[height=1.0em]{images/U+5B50.png}}\hskip0pt{}: ``\raisebox{-0.15em}{\includegraphics[height=1.0em]{images/U+9593.png}}\hskip0pt{}\raisebox{-0.15em}{\includegraphics[height=1.0em]{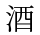}}\hskip0pt{}\raisebox{-0.15em}{\includegraphics[height=1.0em]{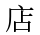}}\hskip0pt{}\raisebox{-0.15em}{\includegraphics[height=1.0em]{images/U+5605.png}}\hskip0pt{}\raisebox{-0.15em}{\includegraphics[height=1.0em]{images/U+670D.png}}\hskip0pt{}\raisebox{-0.15em}{\includegraphics[height=1.0em]{images/U+52D9.png}}\hskip0pt{}\raisebox{-0.15em}{\includegraphics[height=1.0em]{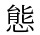}}\hskip0pt{}\raisebox{-0.15em}{\includegraphics[height=1.0em]{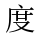}}\hskip0pt{}\raisebox{-0.15em}{\includegraphics[height=1.0em]{images/U+771F.png}}\hskip0pt{}\raisebox{-0.15em}{\includegraphics[height=1.0em]{images/U+4FC2.png}}\hskip0pt{}\raisebox{-0.15em}{\includegraphics[height=1.0em]{images/U+597D.png}}\hskip0pt{}\raisebox{-0.15em}{\includegraphics[height=1.0em]{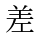}}\hskip0pt{}，\raisebox{-0.15em}{\includegraphics[height=1.0em]{images/U+4EE5.png}}\hskip0pt{}\raisebox{-0.15em}{\includegraphics[height=1.0em]{images/U+5F8C.png}}\hskip0pt{}\raisebox{-0.15em}{\includegraphics[height=1.0em]{images/U+90FD.png}}\hskip0pt{}\raisebox{-0.15em}{\includegraphics[height=1.0em]{images/U+5514.png}}\hskip0pt{}\raisebox{-0.15em}{\includegraphics[height=1.0em]{images/U+6703.png}}\hskip0pt{}\raisebox{-0.15em}{\includegraphics[height=1.0em]{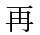}}\hskip0pt{}\raisebox{-0.15em}{\includegraphics[height=1.0em]{images/U+569F.png}}\hskip0pt{}！''
        \item \raisebox{-0.15em}{\includegraphics[height=1.0em]{images/U+4F8B.png}}\hskip0pt{}\raisebox{-0.15em}{\includegraphics[height=1.0em]{images/U+5B50.png}}\hskip0pt{}: ``\raisebox{-0.15em}{\includegraphics[height=1.0em]{images/U+4EFD.png}}\hskip0pt{}\raisebox{-0.15em}{\includegraphics[height=1.0em]{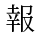}}\hskip0pt{}\raisebox{-0.15em}{\includegraphics[height=1.0em]{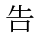}}\hskip0pt{}\raisebox{-0.15em}{\includegraphics[height=1.0em]{images/U+5BEB.png}}\hskip0pt{}\raisebox{-0.15em}{\includegraphics[height=1.0em]{images/U+5F97.png}}\hskip0pt{}\raisebox{-0.15em}{\includegraphics[height=1.0em]{images/U+4E00.png}}\hskip0pt{}\raisebox{-0.15em}{\includegraphics[height=1.0em]{images/U+5572.png}}\hskip0pt{}\raisebox{-0.15em}{\includegraphics[height=1.0em]{images/U+90FD.png}}\hskip0pt{}\raisebox{-0.15em}{\includegraphics[height=1.0em]{images/U+5514.png}}\hskip0pt{}\raisebox{-0.15em}{\includegraphics[height=1.0em]{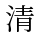}}\hskip0pt{}\raisebox{-0.15em}{\includegraphics[height=1.0em]{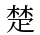}}\hskip0pt{}，\raisebox{-0.15em}{\includegraphics[height=1.0em]{images/U+7747.png}}\hskip0pt{}\raisebox{-0.15em}{\includegraphics[height=1.0em]{images/U+5230.png}}\hskip0pt{}\raisebox{-0.15em}{\includegraphics[height=1.0em]{images/U+6211.png}}\hskip0pt{}\raisebox{-0.15em}{\includegraphics[height=1.0em]{images/U+597D.png}}\hskip0pt{}\raisebox{-0.15em}{\includegraphics[height=1.0em]{images/U+5B32.png}}\hskip0pt{}！''
    \end{enumerate}
\textbf{\raisebox{-0.15em}{\includegraphics[height=1.0em]{images/U+4E2D.png}}\hskip0pt{}\raisebox{-0.15em}{\includegraphics[height=1.0em]{images/U+6027.png}}\hskip0pt{}(neutral)}
\begin{enumerate}
    \item \raisebox{-0.15em}{\includegraphics[height=1.0em]{images/U+610F.png}}\hskip0pt{}\raisebox{-0.15em}{\includegraphics[height=1.0em]{images/U+601D.png}}\hskip0pt{}: \raisebox{-0.15em}{\includegraphics[height=1.0em]{images/U+8868.png}}\hskip0pt{}\raisebox{-0.15em}{\includegraphics[height=1.0em]{images/U+9054.png}}\hskip0pt{}\raisebox{-0.15em}{\includegraphics[height=1.0em]{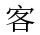}}\hskip0pt{}\raisebox{-0.15em}{\includegraphics[height=1.0em]{images/U+89C0.png}}\hskip0pt{}\raisebox{-0.15em}{\includegraphics[height=1.0em]{images/U+4E8B.png}}\hskip0pt{}\raisebox{-0.15em}{\includegraphics[height=1.0em]{images/U+5BE6.png}}\hskip0pt{}、\raisebox{-0.15em}{\includegraphics[height=1.0em]{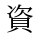}}\hskip0pt{}\raisebox{-0.15em}{\includegraphics[height=1.0em]{images/U+8A0A.png}}\hskip0pt{}、\raisebox{-0.15em}{\includegraphics[height=1.0em]{images/U+6216.png}}\hskip0pt{}\raisebox{-0.15em}{\includegraphics[height=1.0em]{images/U+8005.png}}\hskip0pt{}\raisebox{-0.15em}{\includegraphics[height=1.0em]{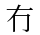}}\hskip0pt{}\raisebox{-0.15em}{\includegraphics[height=1.0em]{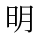}}\hskip0pt{}\raisebox{-0.15em}{\includegraphics[height=1.0em]{images/U+986F.png}}\hskip0pt{}\raisebox{-0.15em}{\includegraphics[height=1.0em]{images/U+60C5.png}}\hskip0pt{}\raisebox{-0.15em}{\includegraphics[height=1.0em]{images/U+611F.png}}\hskip0pt{}\raisebox{-0.15em}{\includegraphics[height=1.0em]{images/U+5605.png}}\hskip0pt{}\raisebox{-0.15em}{\includegraphics[height=1.0em]{images/U+5167.png}}\hskip0pt{}\raisebox{-0.15em}{\includegraphics[height=1.0em]{images/U+5BB9.png}}\hskip0pt{}。
    \item \raisebox{-0.15em}{\includegraphics[height=1.0em]{images/U+89E3.png}}\hskip0pt{}\raisebox{-0.15em}{\includegraphics[height=1.0em]{images/U+91CB.png}}\hskip0pt{}: \raisebox{-0.15em}{\includegraphics[height=1.0em]{images/U+6587.png}}\hskip0pt{}\raisebox{-0.15em}{\includegraphics[height=1.0em]{images/U+5B57.png}}\hskip0pt{}\raisebox{-0.15em}{\includegraphics[height=1.0em]{images/U+4E3B.png}}\hskip0pt{}\raisebox{-0.15em}{\includegraphics[height=1.0em]{images/U+8981.png}}\hskip0pt{}\raisebox{-0.15em}{\includegraphics[height=1.0em]{images/U+4FC2.png}}\hskip0pt{}\raisebox{-0.15em}{\includegraphics[height=1.0em]{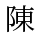}}\hskip0pt{}\raisebox{-0.15em}{\includegraphics[height=1.0em]{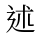}}\hskip0pt{}\raisebox{-0.15em}{\includegraphics[height=1.0em]{images/U+4E8B.png}}\hskip0pt{}\raisebox{-0.15em}{\includegraphics[height=1.0em]{images/U+5BE6.png}}\hskip0pt{}、\raisebox{-0.15em}{\includegraphics[height=1.0em]{images/U+63D0.png}}\hskip0pt{}\raisebox{-0.15em}{\includegraphics[height=1.0em]{images/U+4F9B.png}}\hskip0pt{}\raisebox{-0.15em}{\includegraphics[height=1.0em]{images/U+8CC7.png}}\hskip0pt{}\raisebox{-0.15em}{\includegraphics[height=1.0em]{images/U+8A0A.png}}\hskip0pt{}，\raisebox{-0.15em}{\includegraphics[height=1.0em]{images/U+5187.png}}\hskip0pt{}\raisebox{-0.15em}{\includegraphics[height=1.0em]{images/U+660E.png}}\hskip0pt{}\raisebox{-0.15em}{\includegraphics[height=1.0em]{images/U+986F.png}}\hskip0pt{}\raisebox{-0.15em}{\includegraphics[height=1.0em]{images/U+5605.png}}\hskip0pt{}\raisebox{-0.15em}{\includegraphics[height=1.0em]{images/U+6B63.png}}\hskip0pt{}\raisebox{-0.15em}{\includegraphics[height=1.0em]{images/U+9762.png}}\hskip0pt{}\raisebox{-0.15em}{\includegraphics[height=1.0em]{images/U+6216.png}}\hskip0pt{}\raisebox{-0.15em}{\includegraphics[height=1.0em]{images/U+8CA0.png}}\hskip0pt{}\raisebox{-0.15em}{\includegraphics[height=1.0em]{images/U+9762.png}}\hskip0pt{}\raisebox{-0.15em}{\includegraphics[height=1.0em]{images/U+60C5.png}}\hskip0pt{}\raisebox{-0.15em}{\includegraphics[height=1.0em]{images/U+611F.png}}\hskip0pt{}\raisebox{-0.15em}{\includegraphics[height=1.0em]{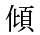}}\hskip0pt{}\raisebox{-0.15em}{\includegraphics[height=1.0em]{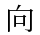}}\hskip0pt{}。
    \item \raisebox{-0.15em}{\includegraphics[height=1.0em]{images/U+4F8B.png}}\hskip0pt{}\raisebox{-0.15em}{\includegraphics[height=1.0em]{images/U+5B50.png}}\hskip0pt{}: ``\raisebox{-0.15em}{\includegraphics[height=1.0em]{images/U+4ECA.png}}\hskip0pt{}\raisebox{-0.15em}{\includegraphics[height=1.0em]{images/U+65E5.png}}\hskip0pt{}\raisebox{-0.15em}{\includegraphics[height=1.0em]{images/U+9999.png}}\hskip0pt{}\raisebox{-0.15em}{\includegraphics[height=1.0em]{images/U+6E2F.png}}\hskip0pt{}\raisebox{-0.15em}{\includegraphics[height=1.0em]{images/U+5605.png}}\hskip0pt{}\raisebox{-0.15em}{\includegraphics[height=1.0em]{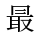}}\hskip0pt{}\raisebox{-0.15em}{\includegraphics[height=1.0em]{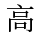}}\hskip0pt{}\raisebox{-0.15em}{\includegraphics[height=1.0em]{images/U+6C23.png}}\hskip0pt{}\raisebox{-0.15em}{\includegraphics[height=1.0em]{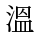}}\hskip0pt{}\raisebox{-0.15em}{\includegraphics[height=1.0em]{images/U+4FC2.png}}\hskip0pt{}\raisebox{-0.15em}{\includegraphics[height=1.0em]{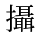}}\hskip0pt{}\raisebox{-0.15em}{\includegraphics[height=1.0em]{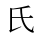}}\hskip0pt{}32\raisebox{-0.15em}{\includegraphics[height=1.0em]{images/U+5EA6.png}}\hskip0pt{}。''
    \item \raisebox{-0.15em}{\includegraphics[height=1.0em]{images/U+4F8B.png}}\hskip0pt{}\raisebox{-0.15em}{\includegraphics[height=1.0em]{images/U+5B50.png}}\hskip0pt{}: ``\raisebox{-0.15em}{\includegraphics[height=1.0em]{images/U+5462.png}}\hskip0pt{}\raisebox{-0.15em}{\includegraphics[height=1.0em]{images/U+500B.png}}\hskip0pt{}\raisebox{-0.15em}{\includegraphics[height=1.0em]{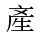}}\hskip0pt{}\raisebox{-0.15em}{\includegraphics[height=1.0em]{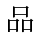}}\hskip0pt{}\raisebox{-0.15em}{\includegraphics[height=1.0em]{images/U+5605.png}}\hskip0pt{}\raisebox{-0.15em}{\includegraphics[height=1.0em]{images/U+4E3B.png}}\hskip0pt{}\raisebox{-0.15em}{\includegraphics[height=1.0em]{images/U+8981.png}}\hskip0pt{}\raisebox{-0.15em}{\includegraphics[height=1.0em]{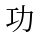}}\hskip0pt{}\raisebox{-0.15em}{\includegraphics[height=1.0em]{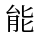}}\hskip0pt{}\raisebox{-0.15em}{\includegraphics[height=1.0em]{images/U+4FC2.png}}\hskip0pt{}\raisebox{-0.15em}{\includegraphics[height=1.0em]{images/U+5E6B.png}}\hskip0pt{}\raisebox{-0.15em}{\includegraphics[height=1.0em]{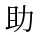}}\hskip0pt{}\raisebox{-0.15em}{\includegraphics[height=1.0em]{images/U+7528.png}}\hskip0pt{}\raisebox{-0.15em}{\includegraphics[height=1.0em]{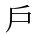}}\hskip0pt{}\raisebox{-0.15em}{\includegraphics[height=1.0em]{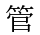}}\hskip0pt{}\raisebox{-0.15em}{\includegraphics[height=1.0em]{images/U+7406.png}}\hskip0pt{}\raisebox{-0.15em}{\includegraphics[height=1.0em]{images/U+6642.png}}\hskip0pt{}\raisebox{-0.15em}{\includegraphics[height=1.0em]{images/U+9593.png}}\hskip0pt{}。''
\end{enumerate}
\end{quote}

Translation in English:
\begin{quote}
This task aims to determine whether the sentiment expressed in a given text is positive, negative, or neutral. We will use three labels: positive, negative, and neutral. Please carefully read the following guidelines to ensure consistency and accuracy in labelling.

\textbf{Positive}\\
\begin{enumerate}
    \item Meaning: Expresses happiness, satisfaction, appreciation, optimism, or other positive emotions.
    \item Explanation: The content of the text makes people feel good and happy, or has a positive evaluation of things.
    \item Examples: 
        \begin{enumerate}
            \item ``The food in this restaurant is really delicious! The service is also great!''
            \item ``The weather is so nice today, it makes me feel good!''
        \end{enumerate}
\end{enumerate}

\textbf{Negative}\\
\begin{enumerate}
    \item Meaning: Expresses unhappiness, dissatisfaction, criticism, anger, sadness, or other negative emotions.
    \item Explanation: The content of the text makes people feel bad, unhappy, or has a negative evaluation of things.
    \item Examples:
        \begin{enumerate}
            \item ``The service attitude of this hotel is really bad, I will never come again!''
            \item ``This report is not clear at all, it makes me very angry!''
        \end{enumerate}
\end{enumerate}

\textbf{Neutral}\\
\begin{enumerate}
    \item Meaning: Expresses objective facts, information, or content without obvious emotions.
    \item Explanation: The text mainly states facts, provides information, and has no obvious positive or negative emotional tendency.
    \item Examples:
        \begin{enumerate}
            \item ``The highest temperature in Hong Kong today is 32 degrees Celsius.''
            \item ``The main function of this product is to help users manage their time.''
        \end{enumerate}
\end{enumerate}
\end{quote}

The average score across the two datasets of each model can be found in Table \ref{tab:nlp-perf-table}.
\end{CJK*}


\end{document}